\documentclass[10pt,twocolumn,letterpaper]{article}
\pdfoutput=1

\usepackage{subcaption}
\usepackage{pgfplots}
\usepackage{pgfplotstable}
\usepgfplotslibrary{groupplots}
\pgfplotsset{compat=newest}

\usepackage{cvpr}
\usepackage{times}
\usepackage{epsfig}
\usepackage{graphicx}
\usepackage{amsmath}
\usepackage{amssymb}
\usepackage{color}
\usepackage{multirow}
\usepackage[T1]{fontenc}
\usepackage{rotating}
\usepackage{tabularx}
\newcommand{\rotatedlabel}[1]{\begin{sideways}#1\end{sideways}}
\usepackage[percent]{overpic}
\usepackage[super]{nth}
\usepackage{xcolor}

\usepackage{algorithm}
\usepackage{algpseudocode}

\algnewcommand\algorithmicparameters{\textbf{Parameters:}}
\algnewcommand\Parameters{\item[\algorithmicparameters]}
\usepackage{ctable}
\setupctable{captionskip = -0.5ex}
\setupctable{botcap}
\newcolumntype{A}{@{}l}
\newcolumntype{B}{@{}X}
\newcolumntype{C}{c@{}}
\newlength{\csTableSkip}
\newcolumntype{U}{X@{\hskip 3\csTableSkip}}
\newcolumntype{V}{@{\hskip \csTableSkip}c@{\hskip \csTableSkip}}
\newcolumntype{W}{@{\hskip 3\csTableSkip}c}
\newcolumntype{N}{@{\hskip 3\csTableSkip}c@{\hskip \csTableSkip}}
\newcolumntype{M}{@{\hskip \csTableSkip}c}
\usepackage{nicefrac}
\usepackage{siunitx,units}
\DeclareSIUnit\inches{in}
\usepackage{cite}
\usepackage{comment}
\graphicspath{{./figures/}}
\graphicspath{{./supplementary_material/}}
\usepackage{placeins}
\usepackage[pagebackref=true,breaklinks=true,colorlinks,bookmarks=false]{hyperref}
\hypersetup{pdftitle={The Cityscapes Dataset for Semantic Urban Scene Understanding},
pdfsubject={The Cityscapes Dataset},
pdfauthor={Marius Cordts, Mohamed Omran, Sebastian Ramos, Timo Rehfeld, Markus Enzweiler, Rodrigo Benenson, Uwe Franke, Stefan Roth, and Bernt Schiele}}
\usepackage[capitalize]{cleveref}

\crefname{section}{Sec.}{Sections}
\Crefname{section}{Section}{Sections}
\crefname{table}{Tab.}{Tables}
\Crefname{table}{Table}{Tables}

\newcommand{\myparagraph}[1]{\vspace{0.5em}\noindent\textbf{#1}}

\cvprfinalcopy % *** Uncomment this line for the final submission

\def\cvprPaperID{1595} % *** Enter the CVPR Paper ID here

\usepackage{amsmath, amssymb, amsfonts, bm, mathtools}

\makeatletter

\newcommand{\minimize}{\@ifstar{\@minimizes}{\@minimize}}
\newcommand{\@minimizes}[1]{\ensuremath{ \operatorname{minimize } } }
\newcommand{\@minimize }[1]{\ensuremath{&\operatorname{minimize} &&}}
\newcommand{\maximize}{\@ifstar{\@maximizes}{\@maximize}}
\newcommand{\@maximizes}[1]{\ensuremath{ \operatorname{maximize } } }
\newcommand{\@maximize }[1]{\ensuremath{&\operatorname{maximize} &&}}
\newcommand{\minimizex}{\@ifstar{\@minimizexs}{\@minimizex}}
\newcommand{\@minimizexs}[1]{\ensuremath{ \underset{#1}{\operatorname{minimize}}\ }}
\newcommand{\@minimizex }[1]{\ensuremath{&\underset{#1}{\operatorname{minimize}}&&}}
\newcommand{\maximizex}{\@ifstar{\@maximizexs}{\@maximizex}}
\newcommand{\@maximizexs}[1]{\ensuremath{ \underset{#1}{\operatorname{maximize}}\ }}
\newcommand{\@maximizex }[1]{\ensuremath{&\underset{#1}{\operatorname{maximize}}&&}}
\newcommand{\argmin}{\@ifstar{\@argmins}{\@argmin}}
\newcommand{\@argmins}[1]{\ensuremath{ \operatorname{argmin } } }
\newcommand{\@argmin }[1]{\ensuremath{&\operatorname{argmin} &&}}
\newcommand{\argmax}{\@ifstar{\@argmaxs}{\@argmax}}
\newcommand{\@argmaxs}[1]{\ensuremath{ \operatorname{argmax } } }
\newcommand{\@argmax }[1]{\ensuremath{&\operatorname{argmax} &&}}
\newcommand{\argminx}{\@ifstar{\@argminxs}{\@argminx}}
\newcommand{\@argminxs}[1]{\ensuremath{ \underset{#1}{\operatorname{argmin}}\ }}
\newcommand{\@argminx }[1]{\ensuremath{&\underset{#1}{\operatorname{argmin}}&&}}
\newcommand{\argmaxx}{\@ifstar{\@argmaxxs}{\@argmaxx}}
\newcommand{\@argmaxxs}[1]{\ensuremath{ \underset{#1}{\operatorname{argmax}}\ }}
\newcommand{\@argmaxx }[1]{\ensuremath{&\underset{#1}{\operatorname{argmax}}&&}}
\newcommand{\minx}{\@ifstar{\@minxs}{\@minx}}
\newcommand{\@minxs}[1]{\ensuremath{ \underset{#1}{\operatorname{min}}\ }}
\newcommand{\@minx }[1]{\ensuremath{&\underset{#1}{\operatorname{min}}&&}}
\newcommand{\maxx}{\@ifstar{\@maxxs}{\@maxx}}
\newcommand{\@maxxs}[1]{\ensuremath{ \underset{#1}{\operatorname{max}}\ }}
\newcommand{\@maxx }[1]{\ensuremath{&\underset{#1}{\operatorname{max}}&&}}

\makeatother

\newcommand{\coarselevel}{\text{category}}
\newcommand{\finelevel}{\text{class}}

\newcommand{\tp}{\text{TP}}
\newcommand{\fp}{\text{FP}}
\newcommand{\fn}{\text{FN}}
\newcommand{\iou}{\text{IoU}}
\newcommand{\miou}{\text{IoU}}
\newcommand{\iiou}{\text{iIoU}}
\newcommand{\miiou}{\text{iIoU}}
\newcommand{\itp}{\text{iTP}}
\newcommand{\ifn}{\text{iFN}}
\newcommand{\apr}{\text{AP}}
\newcommand{\mapr}{\text{AP}}
\newcommand{\bst}[1]{$\mathbf{#1}$} % best number

\definecolor{cityscapes}{RGB}{ 62,135,207}
\definecolor{dus}       {RGB}{220, 20, 60}
\definecolor{camvid}    {RGB}{128, 64,128}
\definecolor{kitti}     {RGB}{ 49,161,137}
\DeclareSIUnit\inch{inch}

\begin{document}
\makeatletter
\renewcommand{\paragraph}{%
\@startsection{paragraph}{4}%
{\z@}{1.0ex \@plus 1ex \@minus .2ex}{-2em}%
{\normalfont \normalsize \bfseries}%
}
\makeatother

%%%%%%%% TITLE
\title{The Cityscapes Dataset for Semantic Urban Scene Understanding}

% Marius Cordts, Mohamed Omran, Sebastian Ramos, Timo Rehfeld, Markus Enzweiler, Rodrigo Benenson, Uwe Franke, Stefan Roth, Bernt Schiele

\author{Marius Cordts$^{1,2}$\\
\and Mohamed Omran$^{3}$\\
\and Sebastian Ramos$^{1,4}$\\
\and Timo Rehfeld$^{1,2}$\\
\and Markus Enzweiler$^{1}$\\
\and Rodrigo Benenson$^{3}$\\
\and Uwe Franke$^{1}$\\
\and Stefan Roth$^{2}$\\
\and Bernt Schiele$^{3}$\\\vspace{-4mm}
\and $^{1}$Daimler AG R\&D, $^{2}$TU Darmstadt, $^{3}$MPI Informatics, $^{4}$TU Dresden\\
\and \url{www.cityscapes-dataset.net}\vspace{-1mm}
}

\twocolumn[{%
    \renewcommand\twocolumn[1][]{#1}%
    \maketitle
    \begin{center}
      \vspace*{-15pt}
      \begin{overpic}[width=0.33\textwidth]{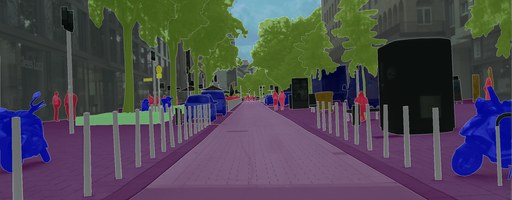}
        \put (6,3) {\textcolor{white}{\textsf{\footnotesize train/val -- fine annotation -- \num{3475} images}}}
      \end{overpic}
      \begin{overpic}[width=0.33\textwidth]{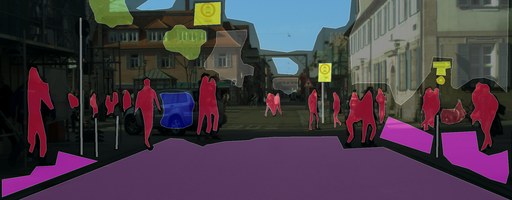}
        \put (4.5,3) {\textcolor{white}{\textsf{\footnotesize train -- coarse annotation -- \num{20000} images}}}
      \end{overpic}
      \begin{overpic}[width=0.33\textwidth]{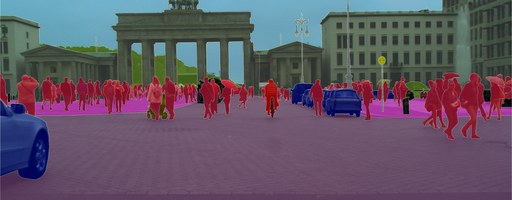}
        \put (10,3) {\textcolor{white}{\textsf{\footnotesize test -- fine annotation -- \num{1525} images}}}
      \end{overpic}
      \vspace*{-5pt}
    \end{center}%
}]

%%%%%%%%% ABSTRACT
\begin{abstract}

Visual understanding of complex urban street scenes is an enabling
factor for a wide range of applications.
Object detection has benefited enormously from large-scale
datasets, especially in the context of deep learning.
For semantic urban scene understanding, however, no current dataset
adequately captures the complexity of real-world urban
scenes.
To address this, we introduce \emph{Cityscapes}, a
benchmark suite and large-scale dataset to train and test approaches
for pixel-level and instance-level semantic labeling.
Cityscapes is comprised of a large, diverse set of stereo video
sequences recorded in streets from \num{50} different cities.
\num{5000} of these images have high quality pixel-level annotations;
\num{20000} additional images have coarse annotations to
enable methods that leverage large volumes of weakly-labeled data.
Crucially, our effort exceeds previous attempts in terms of dataset
size, annotation richness, scene variability, and complexity.
Our accompanying empirical study provides an in-depth analysis of the
dataset characteristics, as well as a performance evaluation of several
state-of-the-art approaches based on our benchmark.

\end{abstract}

\section{Introduction}
\label{sec:introduction}

Visual scene understanding has moved from an elusive goal to a focus
of much recent research in computer vision~\cite{Hoiem2015}.
Semantic reasoning about the contents of a scene is thereby done on
several levels of abstraction.
Scene recognition aims to determine the overall scene category by
putting emphasis on understanding its global properties,
\eg~\cite{Zhou2014, Oliva2001}.
Scene labeling methods, on the other hand, seek to identify the
individual constituent parts of a whole scene as well as their
interrelations on a more local pixel- and instance-level,
\eg~\cite{Long2015,Tighe2015}.
Specialized object-centric methods fall somewhere in between by
focusing on detecting a certain subset of (mostly dynamic) scene
constituents,
\eg~\cite{Felzenszwalb2010, Dollar2012PAMI,Enzweiler2012, Benenson2012}.
Despite significant advances, visual scene understanding remains
challenging, particularly when taking human performance as a
reference.

The resurrection of deep learning~\cite{lecun2015nature} has had a
major impact on the current state-of-the-art in machine learning and
computer vision.
Many top-performing methods in a variety of applications are
nowadays built around deep neural
networks~\cite{Krizhevsky2012,Long2015,Sermanet2014}.
A major contributing factor to their success is the availability of
large-scale, publicly available datasets such as
\textit{ImageNet}~\cite{Russakovsky2014}, \textit{PASCAL
  VOC}~\cite{Everingham2015},
\textit{PASCAL-Context}~\cite{Mottaghi2014}, and \textit{Microsoft
  COCO}~\cite{Lin2014} that allow deep neural networks to develop
their full potential.

% this figure has nothing to do here, but Latex fought me..
% ..and lost
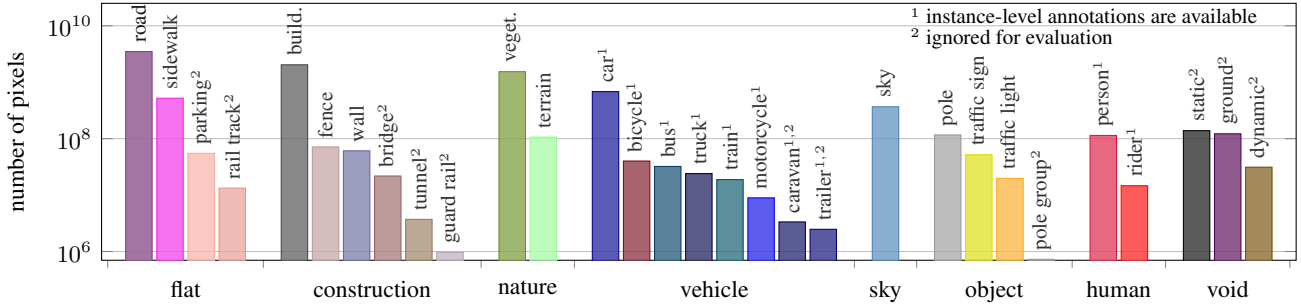
\begin{figure*}[tb]
    \centering
    \definecolor{unlabeled}           {RGB}{  0,  0,  0}
\definecolor{ego vehicle}         {RGB}{  0,  0,  0}
\definecolor{rectification border}{RGB}{  0,  0,  0}
\definecolor{out of roi}          {RGB}{  0,  0,  0}
\definecolor{static}              {RGB}{  0,  0,  0}
\definecolor{dynamic}             {RGB}{111, 74,  0}
\definecolor{ground}              {RGB}{ 81,  0, 81}
\definecolor{road}                {RGB}{128, 64,128}
\definecolor{sidewalk}            {RGB}{244, 35,232}
\definecolor{parking}             {RGB}{250,170,160}
\definecolor{rail track}          {RGB}{230,150,140}
\definecolor{building}            {RGB}{ 70, 70, 70}
\definecolor{wall}                {RGB}{102,102,156}
\definecolor{fence}               {RGB}{190,153,153}
\definecolor{guard rail}          {RGB}{180,165,180}
\definecolor{bridge}              {RGB}{150,100,100}
\definecolor{tunnel}              {RGB}{150,120, 90}
\definecolor{pole}                {RGB}{153,153,153}
\definecolor{pole group}          {RGB}{153,153,153}
\definecolor{traffic light}       {RGB}{250,170, 30}
\definecolor{traffic sign}        {RGB}{220,220,  0}
\definecolor{vegetation}          {RGB}{107,142, 35}
\definecolor{terrain}             {RGB}{152,251,152}
\definecolor{sky}                 {RGB}{ 70,130,180}
\definecolor{person}              {RGB}{220, 20, 60}
\definecolor{rider}               {RGB}{255,  0,  0}
\definecolor{car}                 {RGB}{  0,  0,142}
\definecolor{truck}               {RGB}{  0,  0, 70}
\definecolor{bus}                 {RGB}{  0, 60,100}
\definecolor{caravan}             {RGB}{  0,  0, 90}
\definecolor{trailer}             {RGB}{  0,  0,110}
\definecolor{train}               {RGB}{  0, 80,100}
\definecolor{motorcycle}          {RGB}{  0,  0,230}
\definecolor{bicycle}             {RGB}{119, 11, 32}
\definecolor{license plate}       {RGB}{  0,  0,142}

\newcommand{\fnn}[1]{$\scriptstyle^{#1}$}
\begin{tikzpicture}
\tikzstyle{every node}=[font=\small]
\begin{axis}[
        ybar,
        ymode=log,
        width=\textwidth,
        height=5cm,
        xmin=-0.2,
        xmax=38.2,
        ymin=7e5,
        ymax=2.5e10,
        ylabel={number of pixels},
        xtick={2.5,8.5,13.5,19.5,25,28.5,32.5,36},
        minor xtick={0,5,12,15,24,26,31,34,38},
        xticklabels = {
            flat,
            construction,
            nature,
            vehicle,
            sky,
            object,
            human,
            void,
        },
        major x tick style = {opacity=0},
        minor x tick num = 1,
        xtick pos=left,
        ymajorgrids=true,
        every node near coord/.append style={
                anchor=west,
                rotate=90,
                font=\footnotesize,
        }
        ]

% flat
\addplot[bar shift=0pt,draw=road,          fill opacity=0.9,fill=road!80!white           , nodes near coords=road                 ] plot coordinates{ ( 1,     3488951404  ) };
\addplot[bar shift=0pt,draw=sidewalk,      fill opacity=0.8,fill=sidewalk!80!white       , nodes near coords=sidewalk             ] plot coordinates{ ( 2,     520629811   ) };
\addplot[bar shift=0pt,draw=parking,       fill opacity=0.8,fill=parking!80!white        , nodes near coords=parking\fnn{2}       ] plot coordinates{ ( 3,     54936950    ) };
\addplot[bar shift=0pt,draw=rail track,    fill opacity=0.8,fill=rail track!80!white     , nodes near coords=rail track\fnn{2}    ] plot coordinates{ ( 4,     13301541    ) };
% construction                                                                                                                                                                %
\addplot[bar shift=0pt,draw=building,      fill opacity=0.8,fill=building!80!white       , nodes near coords=build.               ] plot coordinates{ ( 6,     2035086662  ) };
\addplot[bar shift=0pt,draw=fence,         fill opacity=0.8,fill=fence!80!white          , nodes near coords=fence                ] plot coordinates{ ( 7,     71401075    ) };
\addplot[bar shift=0pt,draw=wall,          fill opacity=0.8,fill=wall!80!white           , nodes near coords=wall                 ] plot coordinates{ ( 8,     60820433    ) };
\addplot[bar shift=0pt,draw=bridge,        fill opacity=0.8,fill=bridge!80!white         , nodes near coords=bridge\fnn{2}        ] plot coordinates{ ( 9,     21789831    ) };
\addplot[bar shift=0pt,draw=tunnel,        fill opacity=0.8,fill=tunnel!80!white         , nodes near coords=tunnel\fnn{2}        ] plot coordinates{ ( 10,    3709525     ) };
\addplot[bar shift=0pt,draw=guard rail,    fill opacity=0.8,fill=guard rail!80!white     , nodes near coords=guard rail\fnn{2}    ] plot coordinates{ ( 11,    962778      ) };

% nature                                                                                                                                                                      %
\addplot[bar shift=0pt,draw=vegetation,    fill opacity=0.8,fill=vegetation!80!white     , nodes near coords=veget.               ] plot coordinates{ ( 13,    1538899715  ) };
\addplot[bar shift=0pt,draw=terrain,       fill opacity=0.8,fill=terrain!80!white        , nodes near coords=terrain              ] plot coordinates{ ( 14,    106932536   ) };

% vehicle                                                                                                                                                                     %
\addplot[bar shift=0pt,draw=car,           fill opacity=0.8,fill=car!80!white            , nodes near coords=car\fnn{1}           ] plot coordinates{ ( 16,    682420739   ) };
\addplot[bar shift=0pt,draw=bicycle,       fill opacity=0.8,fill=bicycle!80!white        , nodes near coords=bicycle\fnn{1}       ] plot coordinates{ ( 17,    40121126    ) };
\addplot[bar shift=0pt,draw=bus,           fill opacity=0.8,fill=bus!80!white            , nodes near coords=bus\fnn{1}           ] plot coordinates{ ( 18,    32155223    ) };
\addplot[bar shift=0pt,draw=truck,         fill opacity=0.8,fill=truck!80!white          , nodes near coords=truck\fnn{1}         ] plot coordinates{ ( 19,    24072451    ) };
\addplot[bar shift=0pt,draw=train,         fill opacity=0.8,fill=train!80!white          , nodes near coords=train\fnn{1}         ] plot coordinates{ ( 20,    18813992    ) };
\addplot[bar shift=0pt,draw=motorcycle,    fill opacity=0.8,fill=motorcycle!80!white     , nodes near coords=motorcycle\fnn{1}    ] plot coordinates{ ( 21,    8933989     ) };
\addplot[bar shift=0pt,draw=caravan,       fill opacity=0.8,fill=caravan!80!white        , nodes near coords=caravan\fnn{1,2}     ] plot coordinates{ ( 22,    3346870     ) };
\addplot[bar shift=0pt,draw=trailer,       fill opacity=0.8,fill=trailer!80!white        , nodes near coords=trailer\fnn{1,2}     ] plot coordinates{ ( 23,    2476788     ) };

% sky                                                                                                                                                                         %
\addplot[bar shift=0pt,draw=sky,           fill opacity=0.8,fill=sky!80!white            , nodes near coords=sky                  ] plot coordinates{ ( 25,    367249493   ) };

% object                                                                                                                                                                      %
\addplot[bar shift=0pt,draw=pole,          fill opacity=0.8,fill=pole!80!white           , nodes near coords=pole                 ] plot coordinates{ ( 27,    116436623   ) };
\addplot[bar shift=0pt,draw=traffic sign,  fill opacity=0.8,fill=traffic sign!80!white   , nodes near coords=traffic sign         ] plot coordinates{ ( 28,    51867269    ) };
\addplot[bar shift=0pt,draw=traffic light, fill opacity=0.8,fill=traffic light!80!white  , nodes near coords=traffic light        ] plot coordinates{ ( 29,    19737442    ) };
\addplot[bar shift=0pt,draw=pole group,    fill opacity=0.8,fill=pole group!80!white     , nodes near coords=pole group\fnn{2}    ] plot coordinates{ ( 30,    727994      ) };

% human                                                                                                                                                                       %
\addplot[bar shift=0pt,draw=person,        fill opacity=0.8,fill=person!80!white         , nodes near coords=person\fnn{1}        ] plot coordinates{ ( 32,    113919774   ) };
\addplot[bar shift=0pt,draw=rider,         fill opacity=0.8,fill=rider!80!white          , nodes near coords=rider\fnn{1}         ] plot coordinates{ ( 33,    14651945    ) };

% Void                                                                                                                                                                        %
\addplot[bar shift=0pt,draw=static,        fill opacity=0.8,fill=static!80!white         , nodes near coords=static\fnn{2}        ] plot coordinates{ ( 35,    138567142   ) };
\addplot[bar shift=0pt,draw=ground,        fill opacity=0.8,fill=ground!80!white         , nodes near coords=ground\fnn{2}        ] plot coordinates{ ( 36,    121870792   ) };
\addplot[bar shift=0pt,draw=dynamic,       fill opacity=0.8,fill=dynamic!80!white        , nodes near coords=dynamic\fnn{2}       ] plot coordinates{ ( 37,    31219463    ) };

\node at (axis cs:25.5,3e10) [draw=none,anchor=north west,font=\footnotesize] {\fnn{1} instance-level annotations are available};
\node at (axis cs:25.5,1.3e10) [draw=none,anchor=north west,font=\footnotesize] {\fnn{2} ignored for evaluation};

\end{axis}
\end{tikzpicture}
    \caption{Number of finely annotated pixels (y-axis) per class and their associated categories (x-axis).}
    \label{fig:pixeldistr}
    \vspace{-1em}
\end{figure*}

Despite the existing gap to human performance, scene understanding
approaches have started to become essential components of advanced
real-world systems.
A particularly popular and challenging application involves
self-driving cars, which make extreme demands on system performance
and reliability.
Consequently, significant research efforts have gone into new vision
technologies for understanding complex traffic scenes and
driving scenarios~\cite{Franke2013, Furgale2013,Geiger2014, Scharwachter2014a,
  Ros2015, Badrinarayanan2015}.
Also in this area, research progress can be heavily linked to the existence of datasets such as the \textit{KITTI Vision Benchmark Suite} \cite{Geiger2013a}, \textit{CamVid} \cite{Brostow2009}, \textit{Leuven} \cite{Leibe2007}, and \textit{Daimler Urban Segmentation} \cite{Scharwachter2013} datasets.
These urban scene datasets are often much smaller than datasets
addressing more general settings.
Moreover, we argue that they do not fully capture the variability and
complexity of real-world inner-city traffic scenes.
Both shortcomings currently inhibit further progress in visual understanding of street
scenes.
To this end, we propose the \textit{Cityscapes} benchmark suite
and a corresponding dataset, specifically tailored for autonomous driving
in an urban environment and involving a much wider range of highly
complex inner-city street scenes that were recorded in 50 different
cities.
Cityscapes significantly exceeds previous efforts in terms of size,
annotation richness, and, more importantly, regarding scene complexity and
variability.
We go beyond pixel-level semantic labeling by also considering
instance-level semantic labeling in both our annotations and
evaluation metrics.
To facilitate research on 3D scene understanding, we also provide
depth information through stereo vision.

Very recently, \cite{Xie2015} announced a new semantic scene labeling dataset
for suburban traffic scenes. It provides temporally consistent 3D semantic
instance annotations with 2D annotations obtained through back-projection. We
consider our efforts to be complementary given the differences in the way that
semantic annotations are obtained, and in the type of scenes considered,~\ie
suburban \vs inner-city traffic. To maximize synergies between both datasets,
a common label definition that allows for cross-dataset evaluation has been
mutually agreed upon and implemented.

\section{Dataset}
\label{sec:dataset}

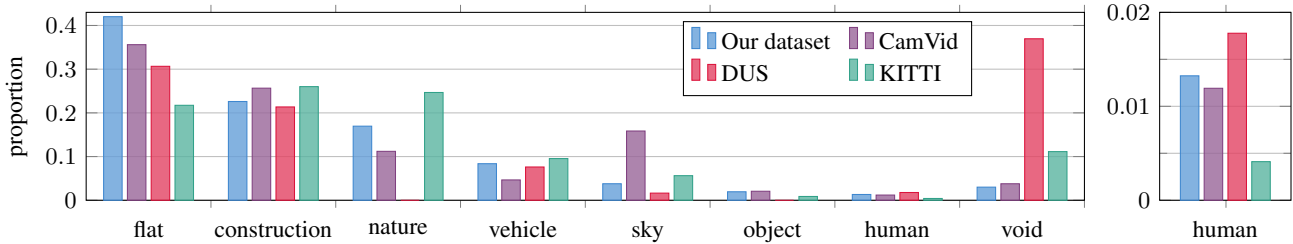
\begin{figure*}
    \centering
    %cityscapes   dus        camvid    kitti                 cityscapes   dus        camvid    kitti
% absolute nb of pixels                                normalized to 1 for each dataset
\pgfplotstableread{
0 2821937623  69061694  230039802   55862948         0.4201334151   0.3065631533    0.3560758347    0.2173418518
1 1581729233  48112037  165728053   66848856         0.2260218146   0.2135681435    0.2565284542    0.2600839138
2 1109008384         0  72346344    63396024         0.1695683414   0               0.1119840332    0.246650235
3 533828644   17156656  30069991    24532369         0.08369464516  0.07615797207   0.04654497636   0.09544627876
4 252167570    3686590  102398569   14469768         0.03783732357  0.01636468192   0.1585015098    0.05629645918
5 131867997          0  13341972    2239957          0.01944870253  0               0.02065187753   0.008714835499
6 88508045     4004878  7701092     1056569          0.01324660708  0.01777755449   0.01192042742   0.004110715084
7 217225139   83255362  24415777    28621539         0.03004915054  0.3695684948    0.03779288671   0.1113557109
}\dataset

 %        flat  0
 %construction  1
 %      nature  2
 %     vehicle  3
 %         sky  4
 %      object  5
 %       human  6
 %        void  7
\pgfplotstableread{
6 88508045     4004878  7701092     1056569          0.01324660708  0.01777755449   0.01192042742   0.004110715084
}\human

\begin{tikzpicture}
\tikzstyle{every node}=[font=\small]
\begin{groupplot}[
        group style = {group size = 2 by 1},
        ybar,
        scale only axis,
        height=2.5cm,
        ymajorgrids=true,
        xtick=data,
        major x tick style = {opacity=0},
        minor x tick num = 1,
        legend columns=2,
        legend style={
            /tikz/column 2/.style={
                column sep=5pt,
            },
        },
        legend cell align=left,
        legend style={at={(axis cs:6.6,0.41)},anchor=north east}
        ]
\nextgroupplot[
        width=0.76\linewidth,
        ymin=0,
        ymax=0.43,
        xmin=-0.5,
        xmax=7.5,
        bar width=7pt,
        xticklabels = {
                   flat,
           construction,
                 nature,
                vehicle,
                    sky,
                 object,
                  human,
                   void
        },
        ylabel={proportion},
]
\addplot[fill opacity=0.8,draw=cityscapes,fill=cityscapes!80!white ] table[x index=0,y index=5] \dataset;
\addplot[fill opacity=0.8,draw=camvid    ,fill=camvid!80!white     ] table[x index=0,y index=7] \dataset;
\addplot[fill opacity=0.8,draw=dus       ,fill=dus!80!white        ] table[x index=0,y index=6] \dataset;
\addplot[fill opacity=0.8,draw=kitti     ,fill=kitti!80!white      ] table[x index=0,y index=8] \dataset;
\legend{Our dataset,CamVid,DUS,KITTI}
\nextgroupplot[
        width=0.1\linewidth,
        xmin=5.5,
        xmax=6.5,
        ymin=0,
        ymax=0.02,
        scaled y ticks = false,
        ytick={0,0.01,0.02},
        minor y tick num = 1,
        yminorgrids=true,
        y tick label style={/pgf/number format/fixed},
        xticklabels = {
                  human
        },
        bar width=7pt,
]
\addplot[fill opacity=0.8,draw=cityscapes,fill=cityscapes!80!white ] table[x index=0,y index=5] \human;
\addplot[fill opacity=0.8,draw=camvid    ,fill=camvid!80!white     ] table[x index=0,y index=7] \human;
\addplot[fill opacity=0.8,draw=dus       ,fill=dus!80!white        ] table[x index=0,y index=6] \human;
\addplot[fill opacity=0.8,draw=kitti     ,fill=kitti!80!white      ] table[x index=0,y index=8] \human;
\end{groupplot}
\end{tikzpicture}
    \caption{Proportion of annotated pixels (y-axis) per category
      (x-axis) for Cityscapes, CamVid \cite{Brostow2009}, DUS
      \cite{Scharwachter2013}, and KITTI \cite{Geiger2013a}.}
    \label{fig:pixeldistrCat}
    \vspace{-0.7em}
\end{figure*}

Designing a large-scale dataset requires a multitude of decisions, \eg
on the modalities of data recording, data preparation, and the
annotation protocol.
Our choices were guided by the ultimate goal of enabling significant
progress in the field of semantic urban scene understanding.

\subsection{Data specifications}
\label{sec:dataset_acquisition}

Our data recording and annotation methodology was carefully designed
to capture the high variability of outdoor street scenes.
Several hundreds of thousands of frames were acquired from a moving
vehicle during the span of several months, covering spring, summer,
and fall in \num{50} cities, primarily in Germany but also in
neighboring countries.
We deliberately did not record in adverse weather conditions, such as
heavy rain or snow, as we believe such conditions to require
specialized techniques and datasets~\cite{Pfeiffer2013}.

Our camera system and post-processing reflect the current state-of-the-art in
the automotive domain. Images were recorded with an automotive-grade
\SI{22}{\centi\meter} baseline stereo camera using \SI{1 / 3}{\inches} CMOS
\SI{2}{\mega P} sensors (OnSemi AR0331) with rolling shutters at a frame-rate
of \SI{17}{\hertz}. The sensors were mounted behind the windshield and yield
high dynamic-range (HDR) images with \SI{16}{bits} linear color depth. Each
\SI{16}{bit} stereo image pair was subsequently debayered and rectified. We
relied on \cite{Krueger2004} for extrinsic and intrinsic calibration. To
ensure calibration accuracy we re-calibrated on-site before each recording
session.

For comparability and compatibility with existing datasets we also provide
low dynamic-range (LDR) \SI{8}{bit} RGB images that are obtained by applying a
logarithmic compression curve. Such tone mappings are common in automotive
vision, since they can be computed efficiently and independently for each
pixel. To facilitate highest annotation quality, we applied a separate tone
mapping to each image. The resulting images are less realistic, but visually
more pleasing and proved easier to annotate. \num{5000} images were manually
selected from \num{27} cities for dense pixel-level annotation, aiming for
high diversity of foreground objects, background, and overall scene layout.
The annotations (see \cref{sec:dataset_annotations}) were done on the \nth{20}
frame of a \num{30}-frame video snippet, which we provide in full to supply
context information. For the remaining \num{23} cities, a single image every
\SI{20}{\second} or \SI{20}{\meter} driving distance (whatever comes first)
was selected for coarse annotation, yielding \num{20000} images in total.

In addition to the rectified \SI{16}{bit} HDR and \SI{8}{bit} LDR stereo image
pairs and corresponding annotations, our dataset includes vehicle odometry
obtained from in-vehicle sensors, outside temperature, and GPS tracks.

\subsection{Classes and annotations}
\label{sec:dataset_annotations}

We provide coarse and fine annotations at pixel level including instance-level
labels for humans and vehicles.

Our \num{5000} fine pixel-level annotations consist of layered polygons (\`{a}
la LabelMe \cite{Russell2008IJCV}) and were realized in-house to guarantee
highest quality levels. Annotation and quality control required more than
\SI{1.5}{\hour} on average for a single image. Annotators were asked to label
the image from back to front such that no object boundary was marked more than
once. Each annotation thus implicitly provides a depth ordering of the objects
in the scene. Given our label scheme, annotations can be easily extended to
cover additional or more fine-grained classes.

For our \num{20000} coarse pixel-level annotations, accuracy on object
boundaries was traded off for annotation speed. We aimed to correctly annotate
as many pixels as possible within a given span of less than \SI{7}{\minute} of
annotation time per image. This was achieved by labeling coarse polygons under
the sole constraint that each polygon must only include pixels belonging to a
single object class.

In two experiments we assessed the quality of our labeling. First, \num{30}
images were finely annotated twice by different annotators and passed the same
quality control. It turned out that \SI{96}{\percent} of all pixels were
assigned to the same label. Since our annotators were instructed to choose a
\textit{void} label if unclear (such that the region is ignored in training
and evaluation), we exclude pixels having at least one \textit{void} label and
recount, yielding \SI{98}{\percent} agreement. Second, all our fine
annotations were additionally coarsely annotated such that we can enable
research on densifying coarse labels. We found that \SI{97}{\percent} of all
labeled pixels in the coarse annotations were assigned the same class as in
the fine annotations.

We defined \num{30} visual classes for annotation, which are grouped into
eight categories: flat, construction, nature, vehicle, sky, object, human, and
void. Classes were selected based on their frequency, relevance from an
application standpoint, practical considerations regarding the annotation
effort, as well as to facilitate compatibility with existing datasets,
\eg~\cite{Geiger2013a,Brostow2009,Xie2015}. Classes that are too rare are
excluded from our benchmark, leaving \num{19} classes for evaluation, see
\cref{fig:pixeldistr} for details. We plan to release our annotation tool upon
publication of the dataset.

\subsection{Dataset splits}
\label{sec:splits}

We split our densely annotated images into separate training,
validation, and test sets.
The coarsely annotated images serve as additional training data only.
We chose not to split the data randomly, but rather in a way that
ensures each split to be representative of the variability of
different street scene scenarios.
The underlying split criteria involve a balanced distribution of
geographic location and population size of the individual cities, as
well as regarding the time of year when recordings took place.
Specifically, each of the three split sets is comprised of data
recorded with the following properties in equal shares:
(i) in large, medium, and small cities;
(ii) in the geographic west, center, and east;
(iii) in the geographic north, center, and south;
(iv) at the beginning, middle, and end of the year.
Note that the data is split at the city level, \ie a city
is completely within a single split.
Following this scheme, we arrive at a unique split consisting of
\num{2975} training and \num{500} validation images with publicly
available annotations, as well as \num{1525} test images with
annotations withheld for benchmarking purposes.

In order to assess how uniform (representative) the splits are
regarding the four split characteristics, we trained a fully
convolutional network~\cite{Long2015} on the \num{500} images in our
validation set.
This model was then evaluated on the whole test set, as well as eight
subsets thereof that reflect the extreme values of the four
characteristics.
With the exception of the time of year, the performance is very
homogeneous, varying less than \SI{1.5}{\percent} points (often much
less).
Interestingly, the performance on the \textit{end of the year} subset is
\SI{3.8}{\percent} points better than on the whole test set.
We hypothesize that this is due to softer lighting conditions in the
frequently cloudy fall.
To verify this hypothesis, we additionally tested on images taken in
low- or high-temperature conditions, finding a \SI{4.5}{\percent}
point increase in low temperatures (cloudy) and a \SI{0.9}{\percent}
point decrease in warm (sunny) weather.
Moreover, specifically training for either condition
leads to an improvement on the respective test set, but not on the
balanced set.
These findings support our hypothesis and
underline the importance of a dataset covering a wide range of
conditions encountered in the real world in a balanced way.

\subsection{Statistical analysis}
\label{sec:stat_analysis}

\ctable[
% star,
caption = {Absolute number and density of annotated pixels for Cityscapes, DUS, KITTI, and CamVid (upscaled to $1280 \times 720$ pixels to maintain the original aspect ratio).\vspace{-1.5em}},
label = table:annodensity,
pos = tb,
width=0.9\columnwidth,
% captionskip = 0.3ex,
doinside=\small,
]{Xcc}{
}{
\FL
 & \#pixels [$10^9$]
 & annot.~density [\si{\percent}]
\ML
Ours (fine)   & $9.43$ & \bst{97.1}\NN
Ours (coarse) & \bst{26.0} & $67.5$\NN
CamVid        & $0.62$ & $96.2$\NN
DUS           & $0.14$ & $63.0$\NN
KITTI         & $0.23$ & $88.9$\LL
}

\newcolumntype{O}{>{\centering\arraybackslash}X}
\ctable[
% star,
caption = {Absolute and average number of instances for Cityscapes,
  KITTI, and Caltech ($^1$ via interpolation) on the respective training and validation datasets.\vspace{-1.5em}},
label = table:instanceStats,
pos = tb,
width=\columnwidth,
% captionskip = 0.3ex,
doinside=\small,
]{lOOOO}{
}{
\FL
 & \#humans [$10^3$]
 & \#vehicles [$10^3$]
 & \#h/image
 & \#v/image
\ML
Ours (fine) & $24.4$        & \bst{41.0} & \bst{7.0} & \bst{11.8} \NN
KITTI       & $ 6.1$        & $30.3$     & $0.8$     & $4.1$      \NN
Caltech     & \bst{192}$^1$ & -          & $1.5$     & -          \LL
}

We compare Cityscapes to other datasets in terms of
(i) annotation volume and density, (ii) the distribution of visual classes,
and (iii) scene complexity.
Regarding the first two aspects, we compare Cityscapes to other datasets with semantic pixel-wise
annotations,~\ie CamVid \cite{Brostow2009}, DUS \cite{Scharwachter2014a},
and KITTI \cite{Geiger2013a}. Note that there are many other datasets with dense semantic
annotations, \eg \cite{Ardeshir2015,Song2015,Sengupta2012automatic,Riemenschneider2014,Tighe2012}.
However, we restrict this part of the analysis to those with a focus on autonomous driving.

CamVid consists of ten minutes of video footage with pixel-wise annotations
for over \num{700} frames. DUS consists of a video sequence of \num{5000}
images from which \num{500} have been annotated. KITTI
addresses several different tasks including semantic labeling and object
detection. As no official pixel-wise annotations exist for KITTI,
several independent groups have annotated approximately \num{700} frames~
\cite{Xu2013,Ladicky2014,Guney2015,Sengupta2013,Ros2015,Kundu2014,Hu2013,Zhang2015}.
We map those labels to our high-level categories and analyze this
consolidated set. In comparison, Cityscapes provides significantly more
annotated images, \ie \num{5000} fine and \num{20000} coarse annotations.
Moreover, the annotation quality and richness is notably better. As Cityscapes
provides recordings from \num{50} different cities, it also covers a
significantly larger area than previous datasets that contain images from a
single city only, \eg Cambridge (CamVid), Heidelberg (DUS), and Karlsruhe (KITTI).
In terms of absolute and relative numbers of semantically annotated pixels
(training, validation, and test data), Cityscapes compares favorably to
CamVid, DUS, and KITTI with up to two orders of magnitude more annotated pixels,
\cf \cref{table:annodensity}. The majority of all annotated pixels
in Cityscapes belong to the coarse annotations, providing many
individual (but correlated) training samples, but missing information close to
object boundaries.

\Cref{fig:pixeldistr,fig:pixeldistrCat} compare the distribution of
annotations across individual classes and their associated higher-level
categories. Notable differences stem from the inherently different
configurations of the datasets. Cityscapes involves dense inner-city traffic
with wide roads and large intersections, whereas KITTI is
composed of less busy suburban traffic scenes. As a result, KITTI exhibits
significantly fewer \textit{flat} ground structures, fewer \textit{humans},
and more \textit{nature}. In terms of overall composition, DUS and CamVid seem
more aligned with Cityscapes. Exceptions are an abundance of \textit{sky}
pixels in CamVid due to cameras with a comparably large vertical field-of-view
and the absence of certain categories in DUS, \ie \textit{nature} and
\textit{object}.

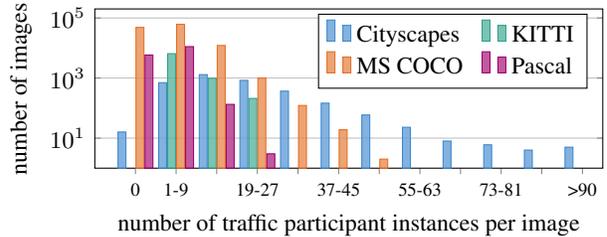
\begin{figure}[t]
    \centering
    % Cityscapes COCO KITTI Pascal
% with explicit zero
% Bins: 0,   5,  14,  23,  32,  41,  50,  59,  68,  77,  86,  95, 104,  113, 122
\pgfplotstableread{
0    16   48503  nan    5849
1    702  61487  6479   11139
2    1302 12155  982    133
3    837  999    210    3
4    367  122    nan    nan
5    146  19     nan    nan
6    59   2      nan    nan
7    23   nan    nan    nan
8    8    nan    nan    nan
9    6    nan    nan    nan
10   4    nan    nan    nan
11   5    nan    nan    nan
}\dataset

\definecolor{cityscapes}{RGB}{ 62,135,207}
\definecolor{dus}       {RGB}{220, 20, 60}
\definecolor{camvid}    {RGB}{128, 64,128}
\definecolor{kitti}     {RGB}{ 49,161,137}
\definecolor{coco}      {RGB}{227,111, 38}
\definecolor{pascal}    {RGB}{158,  0,107}

\begin{tikzpicture}
\tikzstyle{every node}=[font=\small]
\begin{axis}[
        ybar=\pgflinewidth,
        ymode=log,
        width=\columnwidth,
        height=3.7cm,
        xmin=-1,
        xmax=11.5,
        ymin=1,
        ymax=200000,
        ymajorgrids=true,
        bar width=3pt,
        ylabel={number of images},
        xlabel={number of traffic participant instances per image},
        xtick=data,
        xticklabels = {
        0,   1-9,  ,  19-27,  ,  37-45,  ,  55-63,  ,  73-81,  ,  \textgreater 90
        },
        xtick pos=left,
        x tick label style={font=\scriptsize},
        major tick length=0.5ex,
        legend columns=2,
        legend style={
            /tikz/column 2/.style={
                column sep=5pt,
            },
        },
        legend cell align=left,
        legend pos=north east
        ]
\addplot[fill opacity=0.8,draw=cityscapes ,fill=cityscapes!80!white , unbounded coords=jump ]  coordinates {
    ( 0 ,   16  )
    ( 1 ,   702 )
    ( 2 ,   1302)
    ( 3 ,   837 )
    ( 4 ,   367 )
    ( 5 ,   146 )
    ( 6 ,   59  )
    ( 7 ,   23  )
    ( 8 ,   8   )
    ( 9 ,   6   )
    ( 10,   4   )
    ( 11,   5   )
};
\addplot[fill opacity=0.8,draw=kitti      ,fill=kitti!80!white      , unbounded coords=jump ] coordinates {
    ( 0 ,   nan )
    ( 1 ,   6479)
    ( 2 ,   982 )
    ( 3 ,   210 )
    ( 4 ,   nan )
    ( 5 ,   nan )
    ( 6 ,   nan )
    ( 7 ,   nan )
    ( 8 ,   nan )
    ( 9 ,   nan )
    ( 10,   nan )
    ( 11,   nan )
};
\addplot[fill opacity=0.8,draw=coco       ,fill=coco!80!white       , unbounded coords=jump ] coordinates {
    ( 0 ,   48503)
    ( 1 ,   61487)
    ( 2 ,   12155)
    ( 3 ,   999  )
    ( 4 ,   122  )
    ( 5 ,   19   )
    ( 6 ,   2    )
    ( 7 ,   nan  )
    ( 8 ,   nan  )
    ( 9 ,   nan  )
    ( 10,   nan  )
    ( 11,   nan  )
};
\addplot[fill opacity=0.8,draw=pascal     ,fill=pascal!80!white     , unbounded coords=jump ] coordinates {
    ( 0 ,   5849)
    ( 1 ,   11139)
    ( 2 ,   133)
    ( 3 ,   3)
    ( 4 ,   nan)
    ( 5 ,   nan)
    ( 6 ,   nan)
    ( 7 ,   nan)
    ( 8 ,   nan)
    ( 9 ,   nan)
    ( 10,   nan)
    ( 11,   nan)
};

\legend{Cityscapes,KITTI,MS COCO,Pascal}

\end{axis}
\end{tikzpicture}
    \caption{Dataset statistics regarding scene complexity. Only MS COCO and Cityscapes provide instance segmentation masks.}
    \label{fig:instPerImg}
    \vspace{-0.5em}
\end{figure}

\begin{figure}[tb]
    \vspace{-0.1em}
    \centering
    % Cityscapes KITTI
\pgfplotstableread{
0      0    129
1      0    255
2      0    431
3      2    369
4    266    976
5    644   1296
6    647   1377
7   1089   2117
8   1809   2946
9   2380   3875
10  3039   4473
11  3777   4901
12  3685   4062
13  4679   3649
14  4238   1885
15  3935    447
16  3287     17
17  1886     11
18  1355      0
19   943      0
20   365      0
}\dataset

\begin{tikzpicture}
\tikzstyle{every node}=[font=\small]
\begin{axis}[
        ybar=\pgflinewidth,
        % ymode=log,
        width=\columnwidth,
        height=4cm,
        xmin=-0.3,
        xmax=20.3,
        ymin=0,
        ymax=5000,
        ymajorgrids=true,
        bar width=3pt,
        ylabel={number of vehicles},
        xtick=data,
        xticklabels = {
            1 ,
            2 ,
            3 ,
            4 ,
            5 ,
            7 ,
            9 ,
           11 ,
           14 ,
           18 ,
           23 ,
           29 ,
           37 ,
           46 ,
           59 ,
           75 ,
           95 ,
          121 ,
          154 ,
          196 ,
          249 ,
        },
        ytick={0,1000,2000,3000,4000},
        yticklabels = {
            0,
            1k,
            2k,
            3k,
            4k
        },
        xtick pos=left,
        x tick label style={rotate=45, anchor=north east, inner sep=0mm, font=\scriptsize},
        major tick length=0.5ex,
        legend cell align=left,
        legend pos=north west
        ]
\addplot[fill opacity=0.8,draw=cityscapes ,fill=cityscapes!80!white  ] table[x index=0,y index=1] \dataset;
\addplot[fill opacity=0.8,draw=kitti      ,fill=kitti!80!white       ] table[x index=0,y index=2] \dataset;

\legend{Our dataset,KITTI}

\end{axis}
\end{tikzpicture}
    \caption{Histogram of object distances in meters for class \textit{vehicle}.}
    \label{fig:disthist}
    \vspace{-1.5em}
\end{figure}
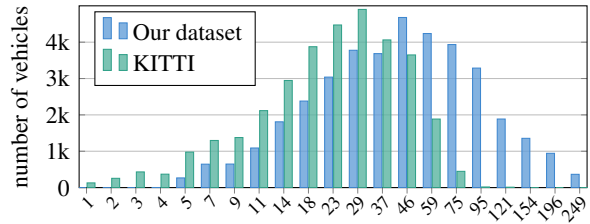

Finally, we assess scene complexity, where density and scale of
traffic participants (humans and vehicles) serve as proxy measures.
Out of the previously discussed datasets,
only Cityscapes and KITTI provide instance-level annotations for humans
and vehicles. We additionally compare to the Caltech Pedestrian
Dataset~\cite{Dollar2012PAMI}, which only contains annotations for humans, but
none for vehicles. Furthermore, KITTI and Caltech only provide instance-level
annotations in terms of axis-aligned bounding boxes.
We use the respective training and validation splits for our analysis,
since test set annotations are not publicly available for all datasets.
In absolute terms, Cityscapes contains significantly more object
instance annotations than KITTI, see \cref{table:instanceStats}. Being
a specialized benchmark, the Caltech dataset provides the most
annotations for humans by a margin. The major share of those labels
was obtained, however, by interpolation between a sparse set of manual
annotations resulting in significantly degraded label quality. The
relative statistics emphasize the much higher complexity of
Cityscapes, as the average numbers of object instances per image
notably exceed those of KITTI and Caltech.
We extend our analysis to MS COCO~\cite{Lin2014} and PASCAL
VOC~\cite{Everingham2015} that also contain street scenes while not
being specific for them.  We analyze the
frequency of scenes with a certain number of traffic
participant instances,
see \cref{fig:instPerImg}. We find our dataset to cover a greater variety of scene
complexity and to have a higher portion of highly complex scenes than previous
datasets.
Using stereo data, we analyze the distribution
of vehicle distances to the camera.
From \cref{fig:disthist} we observe, that in comparison to KITTI, Cityscapes covers
a larger distance range. We attribute this to both our higher-resolution imagery
and the careful annotation procedure. As a consequence, algorithms need to take
a larger range of scales and object sizes into account to score well in our benchmark.

\section{Semantic Labeling} \label{sec:semantic-labeling}

The first Cityscapes task involves predicting a per-pixel
\emph{semantic labeling} of the image without considering higher-level
object instance or boundary information.

\subsection{Tasks and metrics}\label{sec:semantic-metrics}

To assess labeling performance, we rely on a standard and a novel metric.
The first is the standard Jaccard Index, commonly known as the PASCAL VOC intersection-over-union metric $\iou =
\frac{\tp}{\tp+\fp+\fn}$ \cite{Everingham2015}, where $\tp$, $\fp$, and $\fn$ are the numbers of
true positive, false positive, and false negative pixels, respectively, determined over the whole
test set. Owing to the two semantic granularities, \ie classes and categories, we report two separate
mean performance scores: $\miou_{\coarselevel}$ and $\miou_{\finelevel}$. In either case, pixels
labeled as void do not contribute to the score.

The global $\iou$ measure is biased toward object instances that cover a large
image area. In street scenes with their strong scale variation this can be problematic.
Specifically for traffic participants, which are the key classes in our scenario, we aim to evaluate
how well the individual instances in the scene are represented in the labeling. To address this, we
\emph{additionally} evaluate the semantic labeling using an instance-level intersection-over-union
metric $\iiou = \frac{\itp}{\itp+\fp+\ifn}$. Here, $\itp$, and $\ifn$ denote weighted counts of
true positive and false negative pixels, respectively. In contrast to the
standard $\iou$ measure, the contribution of each pixel is weighted by the ratio of the class' average
instance size to the size of the respective ground truth instance.
As before, $\fp$ is the number of false positive pixels.
It is important to note here that unlike the instance-level task in
\cref{sec:instance-level}, we assume that the methods only yield a standard per-pixel semantic class
labeling as output. Therefore, the false positive pixels are not associated with any instance and
thus do not require normalization. The final scores, $\miiou_{\coarselevel}$ and
$\miiou_{\finelevel}$, are obtained as the means for the two semantic granularities, while only
classes with instance annotations are included.

\subsection{Control experiments}\label{sec:semantic-oracles}

We conduct several
control experiments to put our baseline results below into perspective.
First, we count the relative frequency of every class label at each
pixel location of the fine (coarse) training annotations.
Using the most frequent label at each pixel as a constant prediction
irrespective of the test image (called \textit{static fine}, SF, and
\textit{static coarse}, SC) results in roughly \SI{10}{\percent}
$\miou_{\finelevel}$, as shown in \cref{tab:controlResults}.
These low scores emphasize the high diversity of our data. SC and SF
having similar performance indicates the value of our additional
coarse annotations.
Even if the ground truth (GT) segments are re-classified using the
most frequent training label (SF or SC) within each segment mask, the
performance does not notably increase.

Secondly, we re-classify each ground truth segment using FCN-8s~\cite{Long2015},
\cf \cref{sec:semantic-baselines}. We compute the average scores
within each segment and assign the maximizing label.
The performance is significantly better than the static
predictors but still far from \SI{100}{\percent}.
We conclude that it is necessary to optimize both classification and
segmentation quality at the same time.

Thirdly, we evaluate the performance of subsampled ground truth
annotations as predictors.
Subsampling was done by majority voting of neighboring pixels,
followed by resampling back to full resolution.
This yields an upper bound on the performance at a fixed output
resolution and is particularly relevant for deep learning approaches
that often apply downscaling due to constraints on time, memory, or
the network architecture itself.
Downsampling factors \num{2} and \num{4} correspond to the most common setting of
our \nth{3}-party baselines (\cref{sec:semantic-baselines}).
Note that while subsampling by a factor of $2$ hardly affects the
$\iou$ score, it clearly decreases the $\iiou$ score given its
comparatively large impact on small, but nevertheless important
objects.
This underlines the importance of the separate instance-normalized
evaluation.
The downsampling factors of \num{8}, \num{16}, and \num{32} are motivated by the
corresponding strides of the FCN model.
The performance of a GT downsampling by a factor of \num{64} is comparable
to the current state of the art, while downsampling by a factor of \num{128}
is the smallest (power of \num{2}) downsampling for which all images have a
distinct labeling.

Lastly, we employ \num{128}-times subsampled annotations and retrieve the
nearest training annotation in terms of the Hamming distance.
The full resolution version of this training annotation is then used
as prediction, resulting in \SI{21}{\percent} $\miou_{\finelevel}$.
While outperforming the static predictions, the poor result demonstrates
the high variability of our dataset and its demand for approaches that
generalize well.

\subsection{State of the art}
\label{sec:semantic-sota}

Drawing on the success of deep learning algorithms, a number of
semantic labeling approaches have shown very promising
results and significantly advanced the state of the art.
These new approaches take enormous advantage from recently introduced
large-scale datasets, \eg \textit{PASCAL-Context}~\cite{Mottaghi2014}
and \textit{Microsoft COCO}~\cite{Lin2014}.
Cityscapes aims to complement these, particularly in the context of
understanding complex urban scenarios, in order to enable further
research in this area.

The popular work of Long \etal~\cite{Long2015} showed how a
top-performing Convolutional Neural Network (CNN) for image
classification can be successfully adapted for the task of semantic
labeling.
Following this line, \cite{Chen2015, Zheng2015, Schwing2015, Lin2015,
  Liu2015} propose different approaches that combine the strengths of
CNNs and Conditional Random Fields (CRFs).

Other work takes advantage of deep learning for explicitly integrating
global scene context in the prediction of pixel-wise semantic labels,
in particular through CNNs \cite{LiuW2015, Badrinarayanan2015,
  Mostajabi2015, Sharma2015} or Recurrent Neural Networks
(RNNs) \cite{Pinheiro2014, Byeon2015}. Furthermore, a novel
CNN architecture explicitly designed for dense
prediction has been proposed recently by~\cite{Yu2016}.

Last but not least, several studies \cite{Papandreou2015, Dai2015, Pinheiro2015, Xu2015,
Pathak2015a, Pathak2015b, Bearman2015, Wei2015} lately have explored
different forms of weak supervision, such as bounding boxes or
image-level labels, for training CNNs for pixel-level semantic labeling.
We hope our coarse annotations can further advance this area.

\subsection{Baselines}\label{sec:semantic-baselines}

\ctable[
% star,
caption = {Quantitative results of
control experiments for semantic labeling using the metrics
presented in \cref{sec:semantic-metrics}.},
label = tab:controlResults,
pos = t,
width=\columnwidth,
doinside=\small,
]{Xcccc}{
}{
\FL
 \multicolumn{1}{r}{Average over}
 & \multicolumn{2}{c}{Classes}
 & \multicolumn{2}{c}{Categories}
\NN \cmidrule(rl){2-3}\cmidrule(l){4-5}
 \multicolumn{1}{r}{Metric [\%]}
 & $\miou $
 & $\miiou$
 & $\miou $
 & $\miiou$
\ML
static fine (SF)                     & $10.1$ & $ 4.7$ & $26.3$ & $19.9$ \NN
static coarse (SC)                   & $10.3$ & $ 5.0$ & $27.5$ & $21.7$ \NN
GT segmentation with SF              & $10.1$ & $ 6.3$ & $26.5$ & $25.0$ \NN
GT segmentation with SC              & $10.9$ & $ 6.3$ & $29.6$ & $27.0$ \NN[3pt]
GT segmentation with \cite{Long2015} & $79.4$ & $52.6$ & $93.3$ & $80.9$ \NN[3pt]
GT subsampled by $2$                 & $97.2$ & $92.6$ & $97.6$ & $93.3$ \NN
GT subsampled by $4$                 & $95.2$ & $90.4$ & $96.0$ & $91.2$ \NN
GT subsampled by $8$                 & $90.7$ & $82.8$ & $92.1$ & $83.9$ \NN
GT subsampled by $16$                & $84.6$ & $70.8$ & $87.4$ & $72.9$ \NN
GT subsampled by $32$                & $75.4$ & $53.7$ & $80.2$ & $58.1$ \NN
GT subsampled by $64$                & $63.8$ & $35.1$ & $71.0$ & $39.6$ \NN
GT subsampled by $128$               & $50.6$ & $21.1$ & $60.6$ & $29.9$ \NN[3pt]
nearest training neighbor            & $21.3$ & $ 5.9$ & $39.7$ & $18.6$ \LL
}

\newcommand{\emphL}[1]{\textbf{\MakeUppercase #1}}
\newcommand{\yes}{\checkmark}

\ctable[
% star,
caption = {Quantitative results of
baselines for semantic labeling using the metrics
presented in \cref{sec:semantic-metrics}. The first block lists results from our own experiments,
the second from those provided by \nth{3} parties. All numbers are given in percent and we
indicate the used training data for each method, \ie train fine, val fine, coarse extra
as well as a potential downscaling factor (sub) of the input image.\vspace{-1.5em}},
label = tab:baselineResults,
pos = t,
width=\columnwidth,
doinside=\small,
]{Xc@{\hskip 5pt}c@{\hskip 5pt}c@{\hskip 5pt}ccccc}{
}{
\FL
 & \multirow{2}{*}{\begin{sideways}\parbox{\widthof{extraii}}{train }\end{sideways}}
 & \multirow{2}{*}{\begin{sideways}\parbox{\widthof{extraii}}{val   }\end{sideways}}
 & \multirow{2}{*}{\begin{sideways}\parbox{\widthof{extraii}}{coarse}\end{sideways}}
 & \multirow{2}{*}{\begin{sideways}\parbox{\widthof{extraii}}{sub   }\end{sideways}}
 & \multicolumn{2}{c}{Classes}
 & \multicolumn{2}{c}{Categories}
\NN \cmidrule(rl){6-7}\cmidrule(l){8-9}
 &
 &
 &
 &
 & $\miou $
 & $\miiou$
 & $\miou $
 & $\miiou$
\ML

FCN-32s                             & \yes & \yes &      &     & $61.3$   & $38.2$   & $82.2$   & $65.4$    \NN
FCN-16s                             & \yes & \yes &      &     & $64.3$   & $41.1$   & $84.5$   & $69.2$    \NN
FCN-8s                              & \yes & \yes &      &     & $65.3$   & $41.7$   & $85.7$   & $70.1$    \NN
FCN-8s                              & \yes & \yes &      & $2$ & $61.9$   & $33.6$   & $81.6$   & $60.9$    \NN
FCN-8s                              &      & \yes &      &     & $58.3$   & $37.4$   & $83.4$   & $67.2$    \NN
FCN-8s                              &      &      & \yes &     & $58.0$   & $31.8$   & $78.2$   & $58.4$    \ML

\cite{Badrinarayanan2015} extended  & \yes &      &      & $4$ & $56.1$   & $34.2$   & $79.8$   & $66.4$    \NN
\cite{Badrinarayanan2015} basic     & \yes &      &      & $4$ & $57.0$   & $32.0$   & $79.1$   & $61.9$    \NN
\cite{Liu2015}                      & \yes & \yes & \yes & $3$ & $59.1$   & $28.1$   & $79.5$   & $57.9$    \NN
\cite{Zheng2015}                    & \yes &      &      & $2$ & $62.5$   & $34.4$   & $82.7$   & $66.0$    \NN
\cite{Chen2015}                     & \yes & \yes &      & $2$ & $63.1$   & $34.5$   & $81.2$   & $58.7$    \NN
\cite{Papandreou2015}               & \yes & \yes & \yes & $2$ & $64.8$   & $34.9$   & $81.3$   & $58.7$    \NN
\cite{Lin2015}                      & \yes &      &      &     & $66.4$   &\bst{46.7}& $82.8$   & $67.4$    \NN
\cite{Yu2016}                       & \yes &      &      &     &\bst{67.1}& $42.0$   &\bst{86.5}&\bst{71.1} \LL
}

Our own baseline experiments (\cref{tab:baselineResults}, top)
rely on fully convolutional networks (FCNs), as they are central to
most state-of-the-art
methods~\cite{Long2015,Schwing2015,Chen2015,Lin2015,Zheng2015}. We
adopted VGG16 \cite{Simonyan2014} and utilize the PASCAL-context setup
\cite{Long2015} with a modified learning rate to match our image
resolution under an unnormalized loss. According to the notation in \cite{Long2015},
we denote the different models as \textit{FCN-32s}, \textit{FCN-16s}, and
\textit{FCN-8s}, where the numbers are the stride of the finest heatmap.
Since VGG16 training on \SI{2}{MP} images exceeds even the largest GPU
memory available, we split each image into two halves with sufficiently
large overlap.
Additionally, we trained a model on images downscaled by a factor of \num{2}.
We first train on our training set (\textit{train})
until the performance on our validation set (\textit{val}) saturates,
and then retrain on \textit{train+val} with the same number of
epochs.

To obtain further baseline results, we asked selected groups that have
proposed state-of-the-art semantic labeling approaches to optimize
their methods on our dataset and evaluated their predictions on our
test set.
The resulting scores are given in \cref{tab:baselineResults} (bottom)
and qualitative examples of three selected methods are shown in
\cref{fig:exampleresults}.
Interestingly enough, the performance ranking in terms of the main
$\miou_{\finelevel}$ score on Cityscapes is highly different from
PASCAL VOC~\cite{Everingham2015}.
While DPN~\cite{Liu2015} is the \nth{2} best method on PASCAL, it is
only the \nth{6} best on Cityscapes.
FCN-8s~\cite{Long2015} is last on PASCAL, but \nth{3} best on
Cityscapes.
Adelaide~\cite{Lin2015} performs consistently well on both datasets
with rank \num{1} on PASCAL and \num{2} on Cityscapes.

From studying these results, we draw several conclusions:
(1) The amount of downscaling applied during training and testing has a strong
and consistent negative influence on performance (\cf \textit{FCN-8s} \vs
\textit{FCN-8s} at half resolution, as well as the \nth{2} half of the table).
The ranking according to $\miou_{\finelevel}$ is strictly consistent with the
degree of downscaling. We attribute this to the large scale variation present
in our dataset, \cf \cref{fig:disthist}. This observation clearly indicates
the demand for additional research in the direction of memory and
computationally efficient CNNs when facing such a large-scale
dataset with high-resolution images.
(2) Our novel $\miiou$ metric treats instances of any size equally and is
therefore more sensitive to errors in predicting small objects compared to the
$\miou$. Methods that leverage a CRF for regularization
\cite{Zheng2015,Liu2015,Chen2015,Papandreou2015} tend to over smooth small
objects, \cf \cref{fig:exampleresults}, hence show a larger drop from IoU to
iIoU than \cite{Badrinarayanan2015} or FCN-8s~\cite{Long2015}. \cite{Lin2015}
is the only exception; its specific FCN-derived pairwise terms apparently
allow for a more selective regularization.
(3) When considering $\miou_{\coarselevel}$, Dilated10~\cite{Yu2016}
and FCN-8s~\cite{Long2015}
perform particularly well, indicating that these approaches produce
comparatively many confusions between the classes within the same
category, \cf the buses in~\cref{fig:exampleresults} (top).
(4) Training FCN-8s~\cite{Long2015} with \num{500} densely annotated images
(\SI{750}{h} of annotation) yields comparable $\miou$ performance to a model
trained on \num{20000} weakly annotated images (\SI{1300}{h} annot.), \cf
rows \num{5} \& \num{6} in \cref{tab:baselineResults}. However, in both cases
the performance is significantly lower than \textit{FCN-8s} trained on all
\num{3475} densely annotated images. Many fine labels are thus important for
training standard methods as well as for testing, but the performance using
coarse annotations only does not collapse and presents a viable option.
(5) Since the coarse annotations do not include small or distant instances,
their $\miiou$ performance is worse.
(6) Coarse labels can complement the dense labels if applying appropriate
methods as evidenced by \cite{Papandreou2015} outperforming \cite{Chen2015},
which it extends by exploiting both dense and weak annotations (\eg bounding
boxes). Our dataset will hopefully
stimulate research on exploiting the coarse labels further, especially given
the interest in this area, \eg \cite{Oquab2015,Hattori2015,Misra2015}.

Overall, we believe that the unique characteristics of our dataset (\eg scale
variation, amount of small objects, focus on urban street scenes) allow for
more such novel insights.

\begin{figure*}[tb]
    \captionsetup[subfigure]{aboveskip=2pt,belowskip=2pt}
    \centering%
    \begin{subfigure}[b]{0.196\linewidth}
        \includegraphics[width=1\textwidth, trim= 0 1pt 0 0, clip]{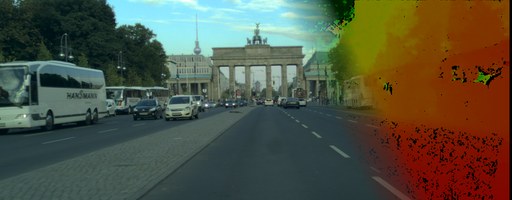}
    \end{subfigure}
    \begin{subfigure}[b]{0.196\linewidth}
        \includegraphics[width=1\textwidth, trim= 0 1pt 0 0, clip]{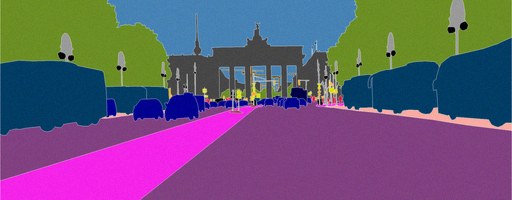}
    \end{subfigure}
    \begin{subfigure}[b]{0.196\linewidth}
        \includegraphics[width=1\textwidth, trim= 0 1pt 0 0, clip]{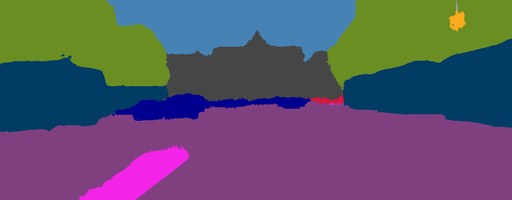}
    \end{subfigure}
    \begin{subfigure}[b]{0.196\linewidth}
        \includegraphics[width=1\textwidth, trim= 0 1pt 0 0, clip]{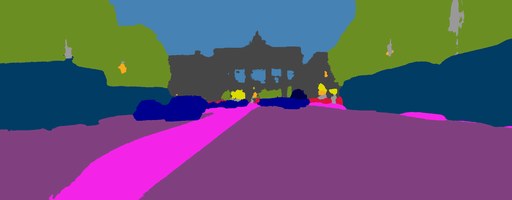}
    \end{subfigure}
    \begin{subfigure}[b]{0.196\linewidth}
        \includegraphics[width=1\textwidth, trim= 0 1pt 0 0, clip]{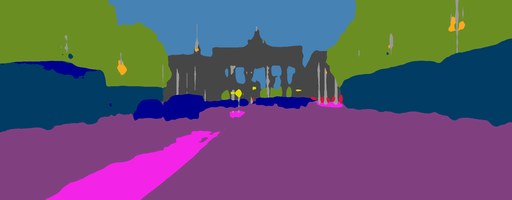}
    \end{subfigure}\\
    \begin{subfigure}[b]{0.196\linewidth}
        \includegraphics[width=1\textwidth, trim= 0 1pt 0 0, clip]{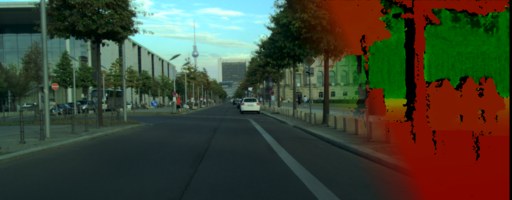}
    \end{subfigure}
    \begin{subfigure}[b]{0.196\linewidth}
        \includegraphics[width=1\textwidth, trim= 0 1pt 0 0, clip]{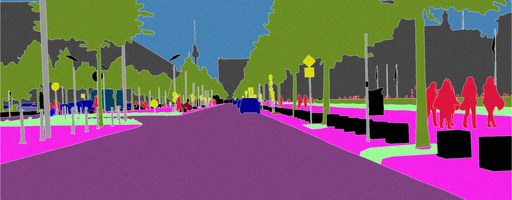}
    \end{subfigure}
    \begin{subfigure}[b]{0.196\linewidth}
        \includegraphics[width=1\textwidth, trim= 0 1pt 0 0, clip]{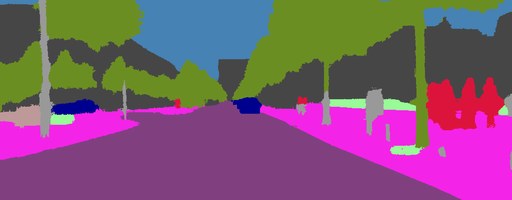}
    \end{subfigure}
    \begin{subfigure}[b]{0.196\linewidth}
        \includegraphics[width=1\textwidth, trim= 0 1pt 0 0, clip]{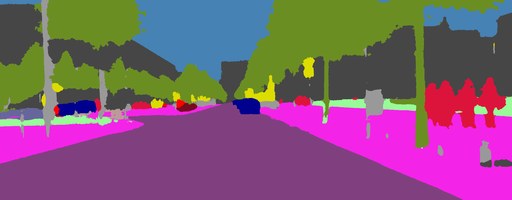}
    \end{subfigure}
    \begin{subfigure}[b]{0.196\linewidth}
        \includegraphics[width=1\textwidth, trim= 0 1pt 0 0, clip]{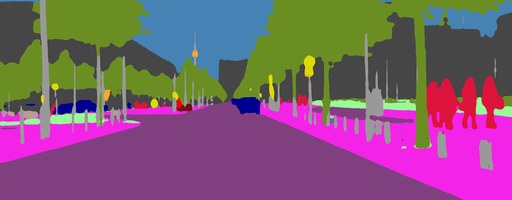}
    \end{subfigure}\\
    \begin{subfigure}[b]{0.196\linewidth}
        \includegraphics[width=1\textwidth, trim= 0 1pt 0 0, clip]{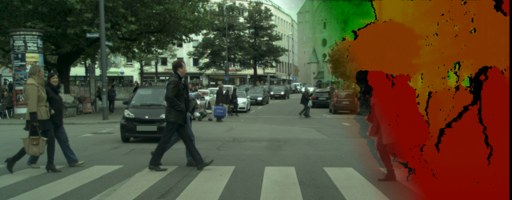}
    \end{subfigure}
    \begin{subfigure}[b]{0.196\linewidth}
        \includegraphics[width=1\textwidth, trim= 0 1pt 0 0, clip]{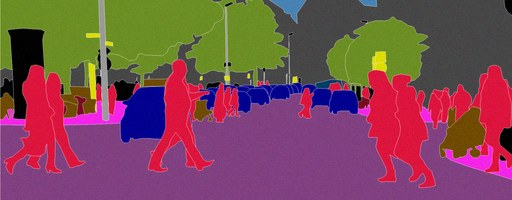}
    \end{subfigure}
    \begin{subfigure}[b]{0.196\linewidth}
        \includegraphics[width=1\textwidth, trim= 0 1pt 0 0, clip]{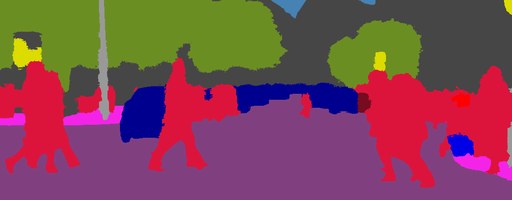}
    \end{subfigure}
    \begin{subfigure}[b]{0.196\linewidth}
        \includegraphics[width=1\textwidth, trim= 0 1pt 0 0, clip]{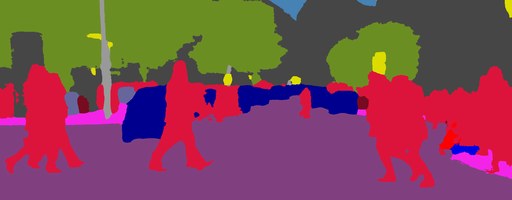}
    \end{subfigure}
    \begin{subfigure}[b]{0.196\linewidth}
        \includegraphics[width=1\textwidth, trim= 0 1pt 0 0, clip]{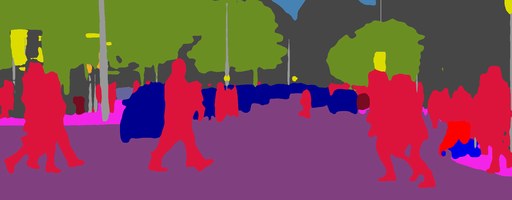}
    \end{subfigure}\\
    \begin{subfigure}[b]{0.196\linewidth}
        \includegraphics[width=1\textwidth, trim= 0 1pt 0 0, clip]{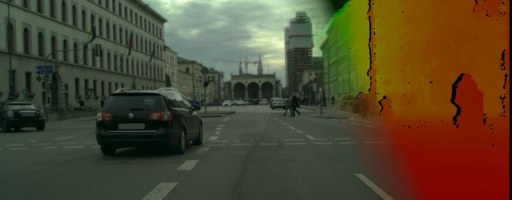}
    \end{subfigure}
    \begin{subfigure}[b]{0.196\linewidth}
        \includegraphics[width=1\textwidth, trim= 0 1pt 0 0, clip]{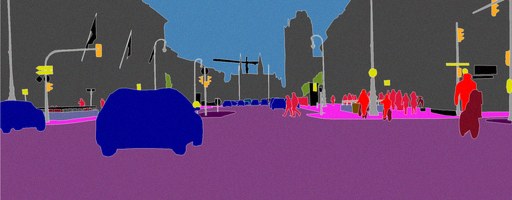}
    \end{subfigure}
    \begin{subfigure}[b]{0.196\linewidth}
        \includegraphics[width=1\textwidth, trim= 0 1pt 0 0, clip]{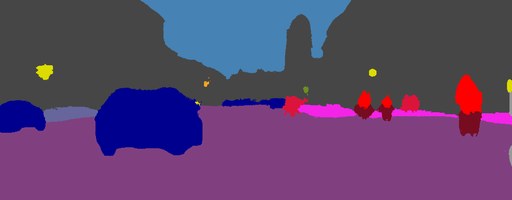}
    \end{subfigure}
    \begin{subfigure}[b]{0.196\linewidth}
        \includegraphics[width=1\textwidth, trim= 0 1pt 0 0, clip]{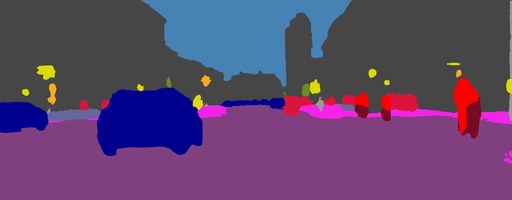}
    \end{subfigure}
    \begin{subfigure}[b]{0.196\linewidth}
        \includegraphics[width=1\textwidth, trim= 0 1pt 0 0, clip]{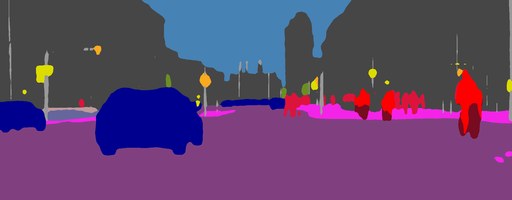}
    \end{subfigure}\\
    \vspace{0.5mm}
    \caption{Qualitative examples of selected baselines.
             From left to right: image with stereo depth maps partially overlayed, annotation, DeepLab~\cite{Papandreou2015},
             Adelaide~\cite{Lin2015}, and Dilated10~\cite{Yu2016}. The color coding of the semantic classes matches \cref{fig:pixeldistr}.}
    \label{fig:exampleresults}
    \vspace{-0.7em}
\end{figure*}

\subsection{Cross-dataset evaluation}\label{sec:semantic-datasets}

In order to show the compatibility and complementarity of Cityscapes regarding
related datasets, we applied an FCN model trained on our data to Camvid
\cite{Brostow2009} and two subsets of KITTI \cite{Ros2015, Sengupta2013}. We use the half-resolution
model (\cf \nth{4} row in \cref{tab:baselineResults}) to better match the target datasets, but we
do not apply any specific training or fine-tuning. In all cases, we
follow the evaluation of the respective
dataset to be able to compare to previously reported results \cite{Badrinarayanan2015, Vineet2015}.
The obtained results in \cref{tab:evalOtherDatasets} show that our
large-scale dataset enables us to train models that are on a par with or even
outperforming methods that are specifically trained on another benchmark
and specialized for its test data.
Further, our analysis shows that our new dataset integrates well with
existing ones and allows for cross-dataset research.

\ctable[
% star,
caption = {Quantitative results (avg.~recall in percent) of our half-resolution FCN-8s model trained on
Cityscapes images and tested on Camvid and KITTI.\vspace{-1.5em}},
label = tab:evalOtherDatasets,
pos = tb,
width=0.85\columnwidth,
doinside=\small,
]{Xcc}{
}{
\FL
 Dataset                    & Best reported result                               & Our result
\ML
Camvid \cite{Brostow2009}   & $62.9$  \cite{Badrinarayanan2015}     & $72.6$\NN
KITTI  \cite{Ros2015}       & $61.6$  \cite{Badrinarayanan2015}     & $70.9$\NN
KITTI  \cite{Sengupta2013}  & $82.2$  \cite{Vineet2015}             & $81.2$\LL
}

\section{Instance-Level Semantic Labeling}
\label{sec:instance-level}

The pixel-level task, \cf \cref{sec:semantic-labeling}, does not aim to
segment individual object instances. In contrast, in the instance-level
semantic labeling task, we focus on simultaneously detecting objects and
segmenting them. This is an extension to both traditional object detection, since
per-instance segments must be provided, and semantic labeling, since
each instance is treated as a separate label.

\subsection{Tasks and metrics}

For instance-level semantic labeling, algorithms are required to deliver a set
of detections of traffic participants in the scene, each associated with a
confidence score and a per-instance segmentation mask.
To assess instance-level performance, we compute the average precision on the
region level ($\apr$ \cite{Hariharan2014}) for each class and average it
across a range of overlap thresholds to avoid a bias towards a specific
value. Specifically, we follow \cite{Lin2014} and use 10 different
overlaps ranging from $0.5$ to $0.95$ in steps of $0.05$. The overlap is
computed at the region level, making it equivalent to the $\iou$ of a single
instance. We penalize multiple predictions of the same ground truth instance
as false positives. To obtain a single, easy to compare compound score, we
report the mean average precision $\mapr$, obtained by also averaging over the
class label set. As minor scores, we add $\mapr^{50\%}$ for an
overlap value of \SI{50}{\percent}, as well as $\mapr^{100\text{m}}$ and
$\mapr^{50\text{m}}$ where the evaluation is restricted to objects within
\SI{100}{\metre} and \SI{50}{\metre} distance, respectively.

\subsection{State of the art}
\label{sub:instance-level-s-o-a}

As detection results have matured (\SI{70}{\percent} mean $\mapr$ on PASCAL \cite{Everingham2015,RenHG015}),
the last years have seen a rising interest in more
difficult settings. Detections with pixel-level segments rather than traditional bounding boxes
provide a richer output and allow (in principle) for better occlusion handling.
We group existing methods into three categories.

The first encompasses \textit{segmentation, then detection} and most
prominently the R-CNN detection framework \cite{Girshick2014CVPR}, relying
on object proposals for generating detections. Many of the commonly used bounding
box proposal methods \cite{HosangBDS15,PTVG2015} first generate a set
of overlapping segments, \eg Selective Search \cite{UijlingsSGS13} or
MCG \cite{cArbelaez14}. In R-CNN, bounding boxes of each segment are then scored
using a CNN-based classifier, while each segment is treated independently.

The second category encompasses \textit{detection, then segmentation},
where bounding-box detections are refined to instance
specific segmentations. Either CNNs
\cite{Hariharan2014,HariharanAGM15} or non-parametric methods
\cite{Chen2015b} are typically used,
however, in both cases without coupling between individual
predictions.

Third, simultaneous \textit{detection and segmentation} is significantly more
delicate. Earlier methods relied on Hough voting
\cite{Leibe2008IJCV, Riemenschneider2012}. More recent works formulate
a joint inference problem on pixel and instance level
using CRFs \cite{Maire2011ICCV, Yao2012CVPR,
He2014CVPR, Dai2015, Tighe2015, Zhang2015}. Differences lie in the generation
of proposals (exemplars, average class shape, direct regression),
the cues considered (pixel-level labeling, depth ordering), and the
inference method (probabilistic, heuristics).

\subsection{Lower bounds, oracles \& baselines}
\label{sec:instance-seg-baselines}

In \cref{table:instanceseg}, we provide
lower-bounds that any sensible method should improve upon, as well as
oracle-case results (\ie using the test time ground truth).
For our experiments, we rely on publicly available implementations.
We train a Fast-R-CNN (\textit{FRCN}) detector \cite{Girshick15ICCV} on our training data
in order to score MCG object proposals \cite{cArbelaez14}.
Then, we use either its output bounding boxes as (rectangular) segmentations, the associated region
proposal, or its convex hull as a per-instance segmentation.
The best main score $\mapr$ is \SI{4.6}{\percent}, is obtained with convex hull proposals, and becomes
larger when restricting the evaluation to \SI{50}{\percent}
overlap or close instances. We contribute these rather low scores to our
challenging dataset, biased
towards busy and cluttered scenes, where
many, often highly occluded, objects occur at various scales, \cf
\cref{sec:dataset}. Further, the MCG bottom-up proposals seem to
be unsuited for such street scenes and cause extremely low scores
when requiring large overlaps.

We confirm this interpretation with oracle experiments, where
we replace the proposals at test-time with ground truth segments
or replace the FRCN classifier with an oracle. In doing so,
the task of object localization is decoupled from the classification task.
The results in \cref{table:instanceseg} show that when bound to MCG proposals,
the oracle classifier is only slightly better than FRCN. On the other hand,
when the proposals are perfect, FRCN achieves decent results.
Overall, these observations unveil that the instance-level performance of
our baseline is bound by the region proposals.

\ctable[
% star,
caption = {Baseline results on instance-level semantic labeling task using the metrics described in \cref{sec:instance-level}. All numbers in \%.\vspace{-2em}},
label = table:instanceseg,
pos = tb,
width=\columnwidth,
doinside=\small,
]{lXcccc}{
}{
\FL
Proposals & Classif.      & $\mapr$     & $\mapr^{50\%}$  & $\mapr^{100\text{m}}$ & $\mapr^{50\text{m}}$
\ML
MCG regions & FRCN      & $ 2.6$ & $ 9.0$ & $ 4.4$ & $ 5.5$ \NN
MCG bboxes  & FRCN      & $ 3.8$ & $11.3$ & $ 6.5$ & $ 8.9$ \NN
MCG hulls   & FRCN      &\bst{4.6}&\bst{12.9}&\bst{7.7}&\bst{10.3}\NN[3pt]

GT bboxes  & FRCN       & $ 8.2$ & $23.7$ & $12.6$ & $15.2$ \NN
GT regions & FRCN       & $41.3$ & $41.3$ & $58.1$ & $64.9$ \NN[3pt]

MCG regions  & GT       & $10.5$ & $27.0$ & $16.0$ & $18.7$ \NN
MCG bboxes   & GT       & $ 9.9$ & $25.8$ & $15.3$ & $18.9$ \NN
MCG hulls    & GT       & $11.6$ & $29.1$ & $17.7$ & $21.4$ \LL
}

\section{Conclusion and Outlook}
\label{sec:conclusion}

In this work, we presented Cityscapes, a comprehensive benchmark suite that
has been carefully designed to spark progress in semantic urban scene
understanding by: (i) creating the largest and most diverse dataset of street
scenes with high-quality and coarse annotations to date; (ii) developing a
sound evaluation methodology for pixel-level and instance-level semantic
labeling; (iii) providing an in-depth analysis of the characteristics of our
dataset; (iv) evaluating several state-of-the-art approaches on our benchmark.
To keep pace with the rapid progress in scene understanding,
we plan to adapt Cityscapes to future needs over
time.

The significance of Cityscapes is all the more apparent from three
observations. First, the relative order of performance for state-of-the-art
methods on our dataset is notably different than on more generic datasets such
as PASCAL VOC. Our conclusion is that serious progress in urban scene
understanding may not be achievable through such generic datasets. Second, the
current state-of-the-art in semantic labeling on KITTI and CamVid is easily
reached and to some extent even outperformed by applying an off-the-shelf
fully-convolutional network~\cite{Long2015} trained on Cityscapes only, as
demonstrated in \cref{sec:semantic-datasets}. This underlines the
compatibility and unique benefit of our dataset. Third, Cityscapes will pose a
significant new challenge for our field given that it is currently far from
being solved. The best performing baseline for pixel-level semantic segmentation
obtains an $\miou$ score of
\SI{67.1}{\percent}, whereas the best current methods on PASCAL VOC and KITTI
reach $\miou$ levels of \SI{77.9}{\percent}~\cite{Arnab2015} and
\SI{72.5}{\percent}~\cite{Vineet2015}, respectively.
In addition, the instance-level task is particularly challenging with an
$\mapr$ score of \SI{4.6}{\percent}.

{\small
\myparagraph{Acknowledgments.}
S. Roth was supported in part by the European Research Council under the EU's \nth{7} Framework Programme (FP/2007-2013)/ERC Grant agreement no. 307942.
The authors acknowledge the support of the Bundesministerium f\"ur Wirtschaft und Technologie (BMWi) in the context of the UR:BAN initiative.\\
We thank the \nth{3}-party authors for their valuable submissions.

\appendix
\bibliographystyle{ieee}
\bibliography{bib}
}
\clearpage
\pagenumbering{roman}
\section{Related Datasets}

In \cref{tab:relatedDatasets} we provide a comparison to other related
datasets in terms of the type of annotations, the meta information provided,
the camera perspective, the type of scenes, and their size. The selected
datasets are either of large scale or focus on street scenes.

\FloatBarrier

\newcommand{\no}{$\times$}

\ctable[
star,
caption = {
Comparison to related datasets. We list the type of labels provided, \ie
object bounding boxes (B), dense pixel-level semantic labels (D), coarse
labels (C) that do not aim to label the whole image. Further, we mark if
color, video, and depth information are available. We list the camera
perspective, the scene type, the number of images, and the number of semantic
classes.
},
label = tab:relatedDatasets,
pos = tb,
doinside=\footnotesize,
captionskip=1ex,
]{lccccccccc}{
\tnote[a]{Including the annotations of \nth{3} party groups~\cite{Xu2013,Ladicky2014,Guney2015,Sengupta2013,Ros2015,Kundu2014,Hu2013,Zhang2015}}
}{
\FL
Dataset                      & Labels        & Color & Video & Depth
& Camera     & Scene    & \#{}images                       & \#{}classes
\ML
\cite{Russakovsky2014}       & B             & \yes  & \no   & \no            & Mixed      & Mixed    & \SI{150}{k}                    & 1000            \NN % ImageNet
\cite{Everingham2015}        & B, C          & \yes  & \no   & \no            & Mixed      & Mixed    & \SI{20}{k} (B), \SI{10}{k} (C) & 20              \NN % PASCAL VOC
\cite{Mottaghi2014}          & D             & \yes  & \no   & \no            & Mixed      & Mixed    & \SI{20}{k}                     & 400             \NN % PASCAL-Context
\cite{Lin2014}               & C             & \yes  & \no   & \no            & Mixed      & Mixed    & \SI{300}{k}                    & 80              \NN % Microsoft COCO
\cite{Song2015}              & D, C          & \yes  & \no   & Kinect         & Pedestrian & Indoor   & \SI{10}{k}                     & 37              \NN % SUN RGB-D
\cite{Geiger2013a}           & B, D\tmark[a] & \yes  & \yes  & Laser, Stereo  & Car        & Suburban & \SI{15}{k} (B), \num{700} (D)  & 3 (B), 8 (D)    \NN % KITTI
\cite{Brostow2009}           & D             & \yes  & \yes  & \no            & Car        & Urban    & \num{701}                      & 32              \NN % CamVid
\cite{Leibe2007}             & D             & \yes  & \yes  & Stereo, Manual & Car        & Urban    & \num{70}                       & 7               \NN % Leuven
\cite{Scharwachter2013}      & D             & \no   & \yes  & Stereo         & Car        & Urban    & \num{500}                      & 5               \NN % DUS
\cite{Ardeshir2015}          & D             & \yes  & \no   & \no            & Pedestrian & Urban    & \num{200}                      & 2               \NN % Geosemantic segmentation
\cite{Sengupta2012automatic} & C             & \yes  & \no   & Stereo         & Car        & Facades  & 86                             & 13              \NN %  Automatic dense visual semantic mapping from street-level imagery
\cite{Riemenschneider2014}   & D             & \yes  & \no   & 3D mesh        & Pedestrian & Urban    & 428                            & 8               \NN
\cite{Xie2015}               & D             & \yes  & \yes  & Laser          & Car        & Suburban & \SI{400}{k}                    & 27              \NN % Andreas' submission
% \cite{Tighe2012}             & D             & \yes  & \no   & \no            & Pedestrian & Urban    & ??                             & ??              \NN % skipped, I don't understand the data and it seems offline anyway
Ours                         & D, C          & \yes  & \yes  & Stereo         & Car        & Urban    & \SI{5}{k} (D), \SI{20}{k} (C)  & 30              \LL
}

%\todo{replace rail track example}
%\todo{check spelling of labeling not labelling}

\section{Class Definitions}
\label{sec:classes}

\Cref{tab:Annotated-labels} provides precise definitions of our annotated classes.
These definitions were used to guide our labeling process, as well as quality control.
In addition, we include a typical example for each class.

The annotators were instructed to make use of the depth ordering and occlusions of the scene
to accelerate labeling, analogously to LabelMe~\cite{Russell2008IJCV}; see \cref{fig:labelProcess}
for an example. In doing so, distant objects are annotated first, while occluded
parts are annotated with a coarser, conservative boundary (possibly larger than the actual object).
Subsequently, the occluder is annotated with a polygon that lies in front of the occluded part. Thus, the boundary between these objects is shared and
consistent.

Holes in an object through which a background region can be seen are considered to be part of the object.
This allows keeping the labeling effort within reasonable bounds such that
objects can be described via simple polygons forming simply-connected sets.

\newcolumntype{J}[1]{>{\centering\arraybackslash}m{#1}}
% \begin{table}[ht]
%   \centering
%     \begin{tabular}{p{20mm}p{20mm}J{20mm}} \toprule
%         Type   & Specs   & Uses \\
%         --     & Model 1 & \bla \\
%                & Model 2 & \bla \\
%         Spring & Model 3 & \bla \\
%                & Model 4 & \bla \\
%     \end{tabular}
% \end{table}

\newcolumntype{L}[1]{>{\raggedright\arraybackslash}p{#1}}
\newcommand{\groupsep}{\ML}

\ctable[
star,
caption = List of annotated classes including their definition and typical example.,
label = tab:Annotated-labels,
pos = p,
width=\textwidth,
doinside=\small,
captionskip=0.6ex,
]
{p{0.09\textwidth}p{0.09\textwidth}m{0.36\textwidth}m{0.36\textwidth}}
{
	\tnote[1]{Single instance annotation available.}
	\tnote[2]{Not included in challenges.}
}{ \FL
Category & Class & Definition & Examples \ML
%%%%%%%%%%%%%%%%%%%%%%%%%%%%%%%%%%%%%%%%%%%%%%%%%%%%%%%%%%%%%%%%%%%%%%%%%%%%%%%%%%%%%%%%%
human
& person\tmark[1]
    & All humans that would primarily rely on their legs to move if necessary.
Consequently, this label includes people who are standing/sitting, or otherwise stationary.
This class also includes babies, people pushing a bicycle, or standing next to it
with both legs on the same side of the bicycle.
    &
    \raisebox{-\totalheight}{\includegraphics[height=3.15cm]{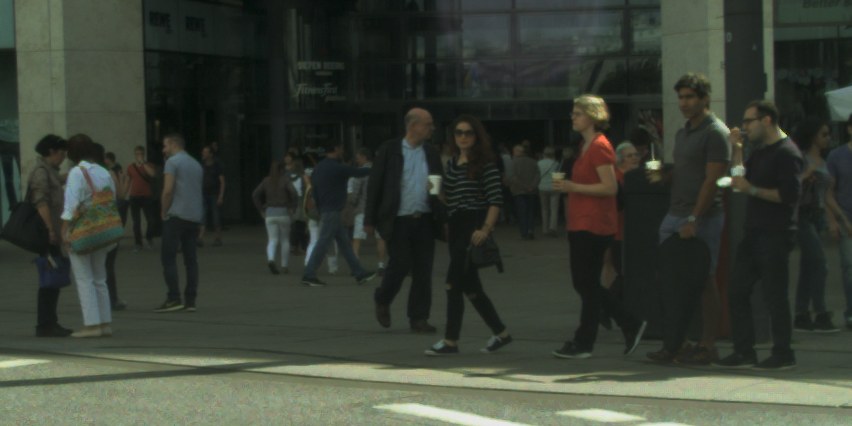}}
    \NN
& rider\tmark[1]
    & Humans relying on some device for movement. This includes drivers,
passengers, or riders of bicycles, motorcycles, scooters, skateboards, horses, Segways,
(inline) skates, wheelchairs, road cleaning cars, or convertibles. Note that a
visible driver of a closed car can only be seen through the window. Since holes are considered part of the surrounding object, the human is included in the \textit{car} label.
    &
    \raisebox{-\totalheight}{\includegraphics[height=3.15cm]{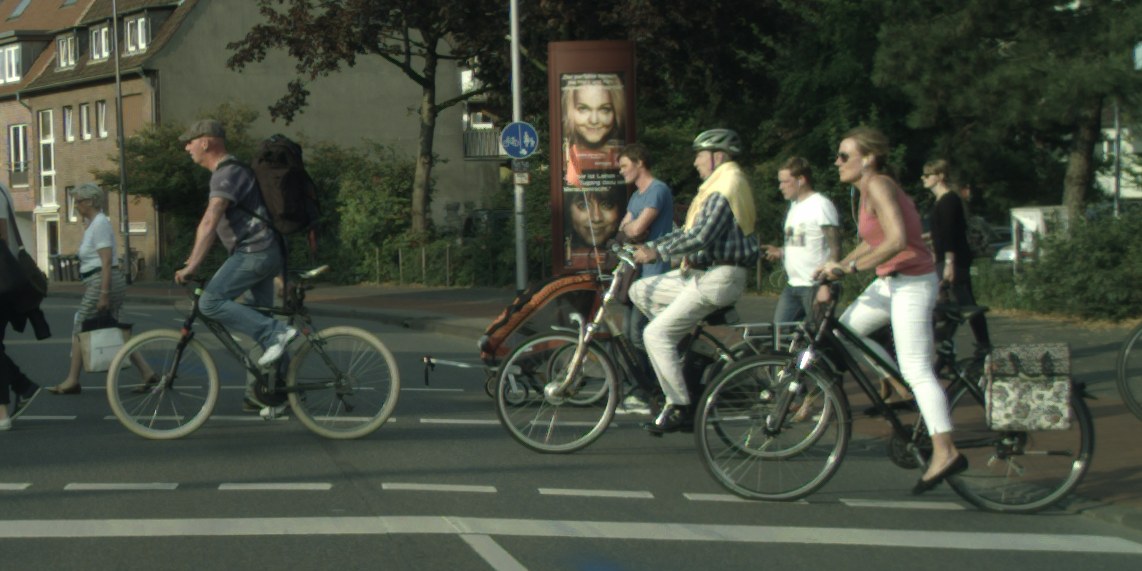}}
    \groupsep
%%%%%%%%%%%%%%%%%%%%%%%%%%%%%%%%%%%%%%%%%%%%%%%%%%%%%%%%%%%%%%%%%%%%%%%%%%%%%%%%%%%%%%%%%
vehicle
& car\tmark[1]
    & This includes cars, jeeps, SUVs, vans with a continuous body shape (\ie the
    driver's cabin and cargo compartment are one). Does not include trailers, which have
    their own separate class.
    &
    \includegraphics[height=3.15cm]{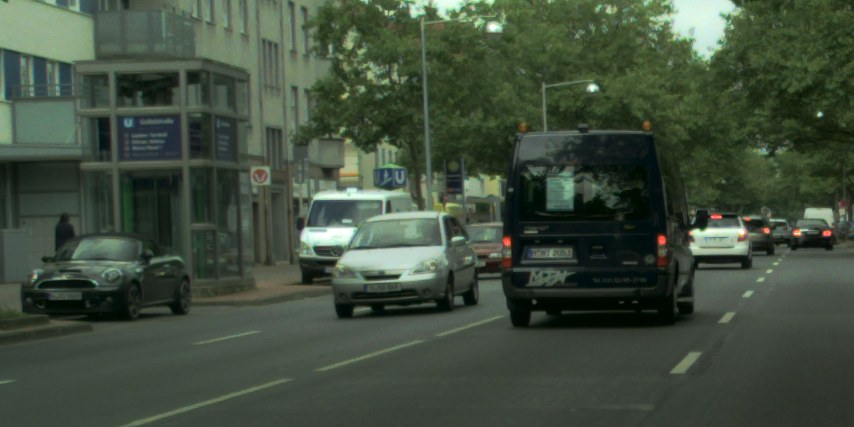}
    \NN
& truck\tmark[1]
    & This includes trucks, vans with a body that is separate from the driver's
    cabin, pickup trucks, as well as their trailers.
    &
    \includegraphics[height=3.15cm]{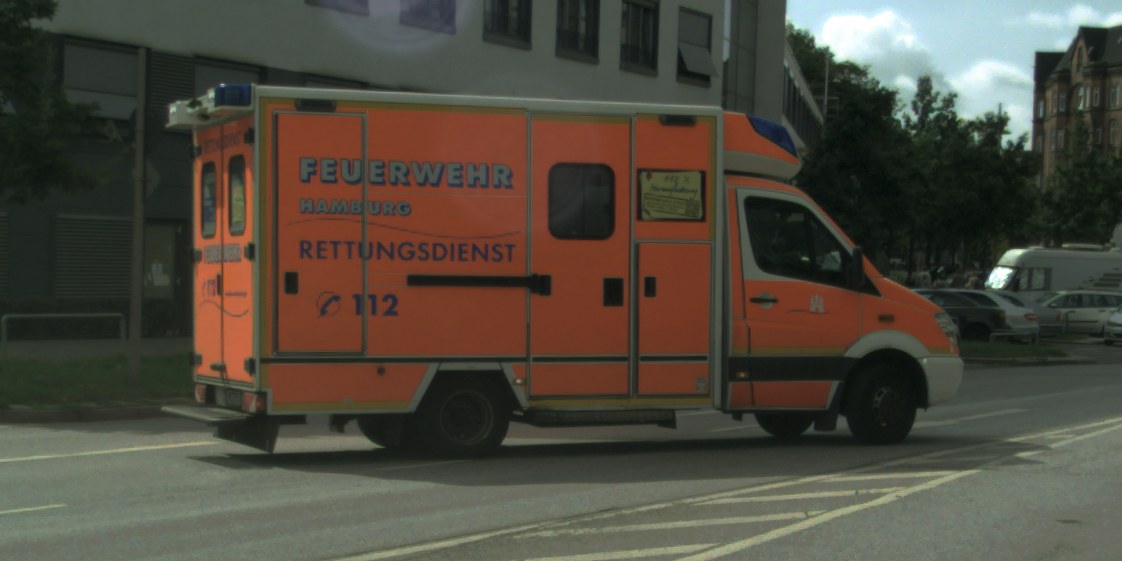}
    \NN
& bus\tmark[1]
    & This includes buses that are intended for 9+ persons for public or long-distance transport.
    &
    \includegraphics[height=3.15cm]{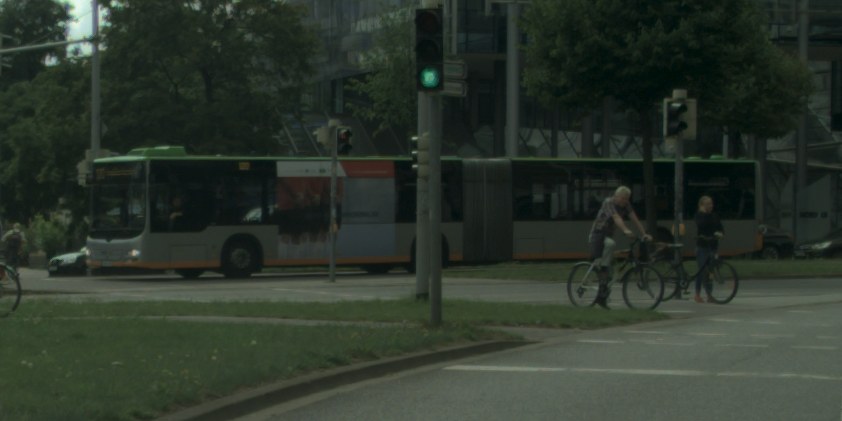}
    \NN
& train\tmark[1]
    & All vehicles that move on rails, \eg trams, trains.
    &
    \includegraphics[height=3.15cm]{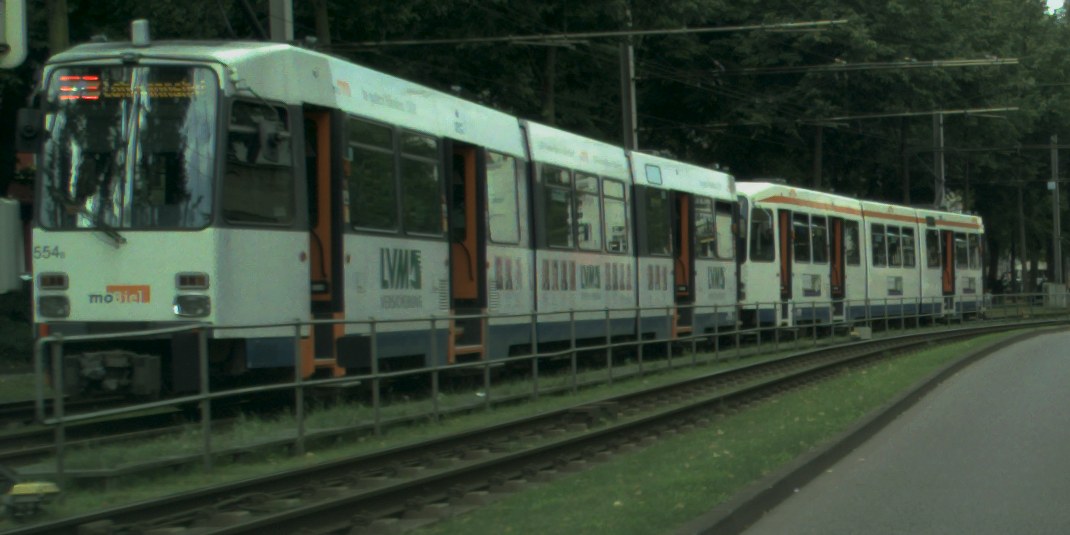}
    \LL
}

\ctable[
star,
caption = List of annotated classes including their definition and typical example.,
% label = tab:Annotated-labels,
pos = p,
width=\textwidth,
doinside=\small,
captionskip = 0.6ex,
continued
]
{p{0.09\textwidth}p{0.09\textwidth}m{0.36\textwidth}m{0.36\textwidth}}
{
    \tnote[1]{Single instance annotation available.}
    \tnote[2]{Not included in challenges.}
}{ \FL
Category & Class & Definition & Examples \ML
vehicle
& motorcycle\tmark[1]
    & This includes motorcycles, mopeds, and scooters without the driver or other
    passengers. The latter receive the label \textit{rider}.
    &
    \includegraphics[height=3.15cm]{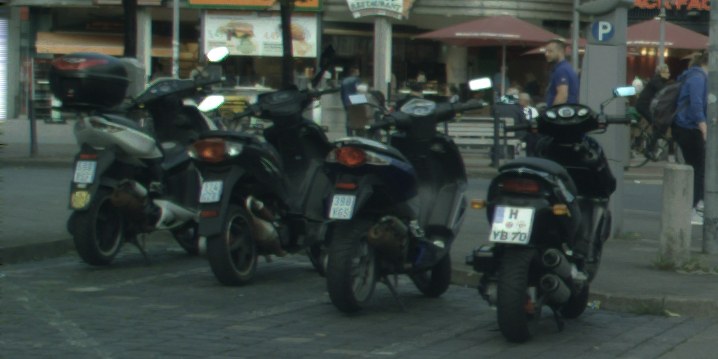}
    \NN
& bicycle\tmark[1]
    & This includes bicycles without the cyclist or other passengers. The latter
    receive the label \textit{rider}.
    &
    \includegraphics[height=3.15cm]{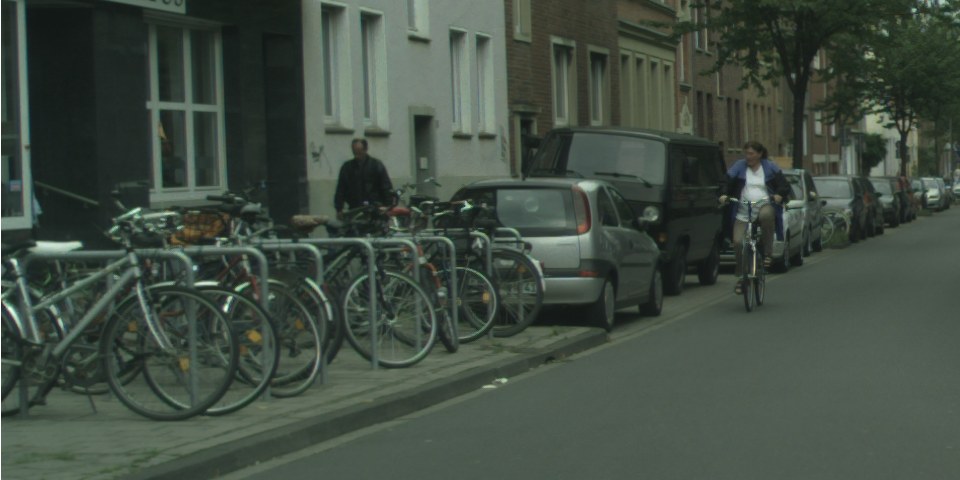}
    \NN
& caravan\tmark[1,2]
    & Vehicles that (appear to) contain living quarters. This also includes trailers that are
    used for living and has priority over the \textit{trailer} class.
    &
    \includegraphics[height=3.15cm]{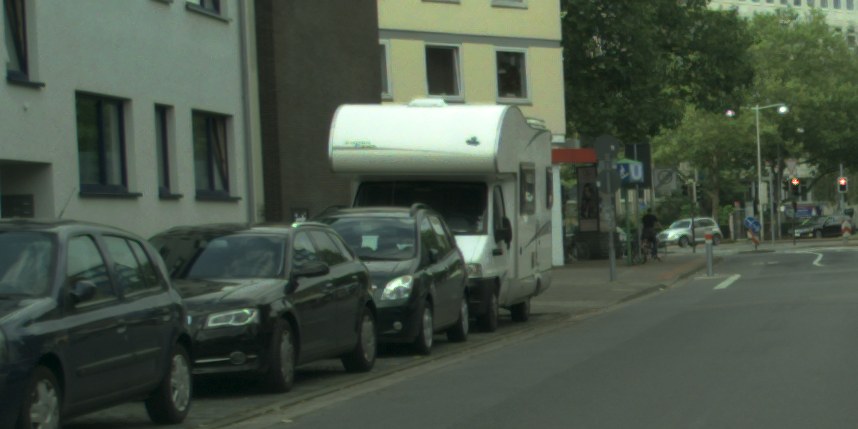}
    \NN
& trailer\tmark[1,2]
    & Includes trailers that can be attached to any vehicle, but excludes trailers
    attached to trucks. The latter are included in the \textit{truck} label.
    &
    \includegraphics[height=3.15cm]{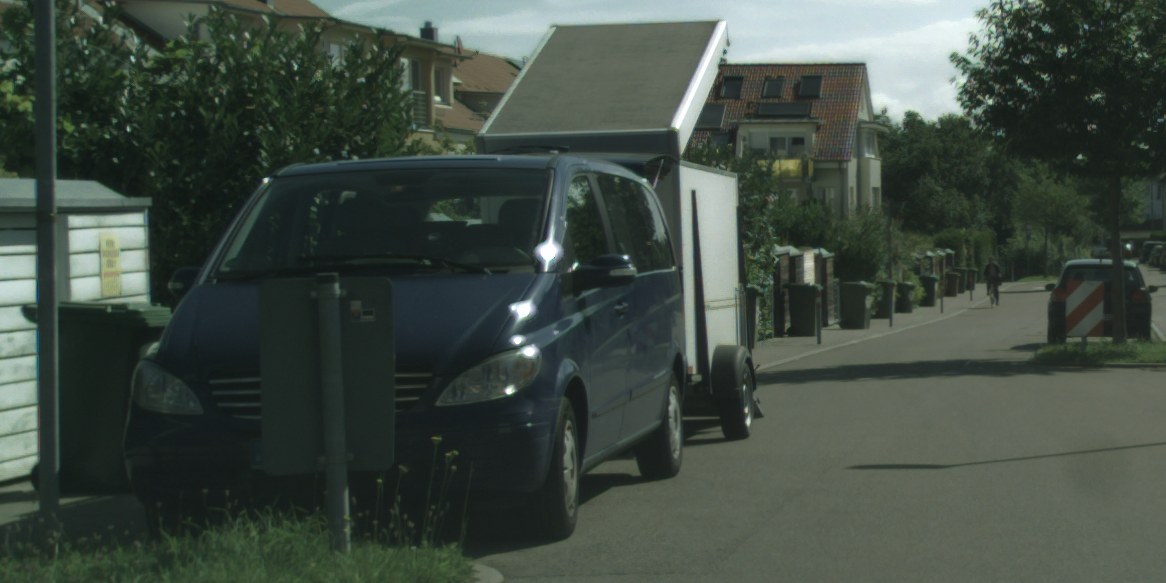}
%     \NN
% & license plate\tmark[1,2]
%     & Only applies to license plates with legible text.
%     &
    \groupsep
%%%%%%%%%%%%%%%%%%%%%%%%%%%%%%%%%%%%%%%%%%%%%%%%%%%%%%%%%%%%%%%%%%%%%%%%%%%%%%%%%%%%%%%%%
nature
& vegetation
    & Trees, hedges, and all kinds of vertically growing vegetation. Plants
    attached to buildings/walls/fences are not annotated separately, and
    receive the same label as the surface they are supported by.
    &
    \includegraphics[height=3.15cm]{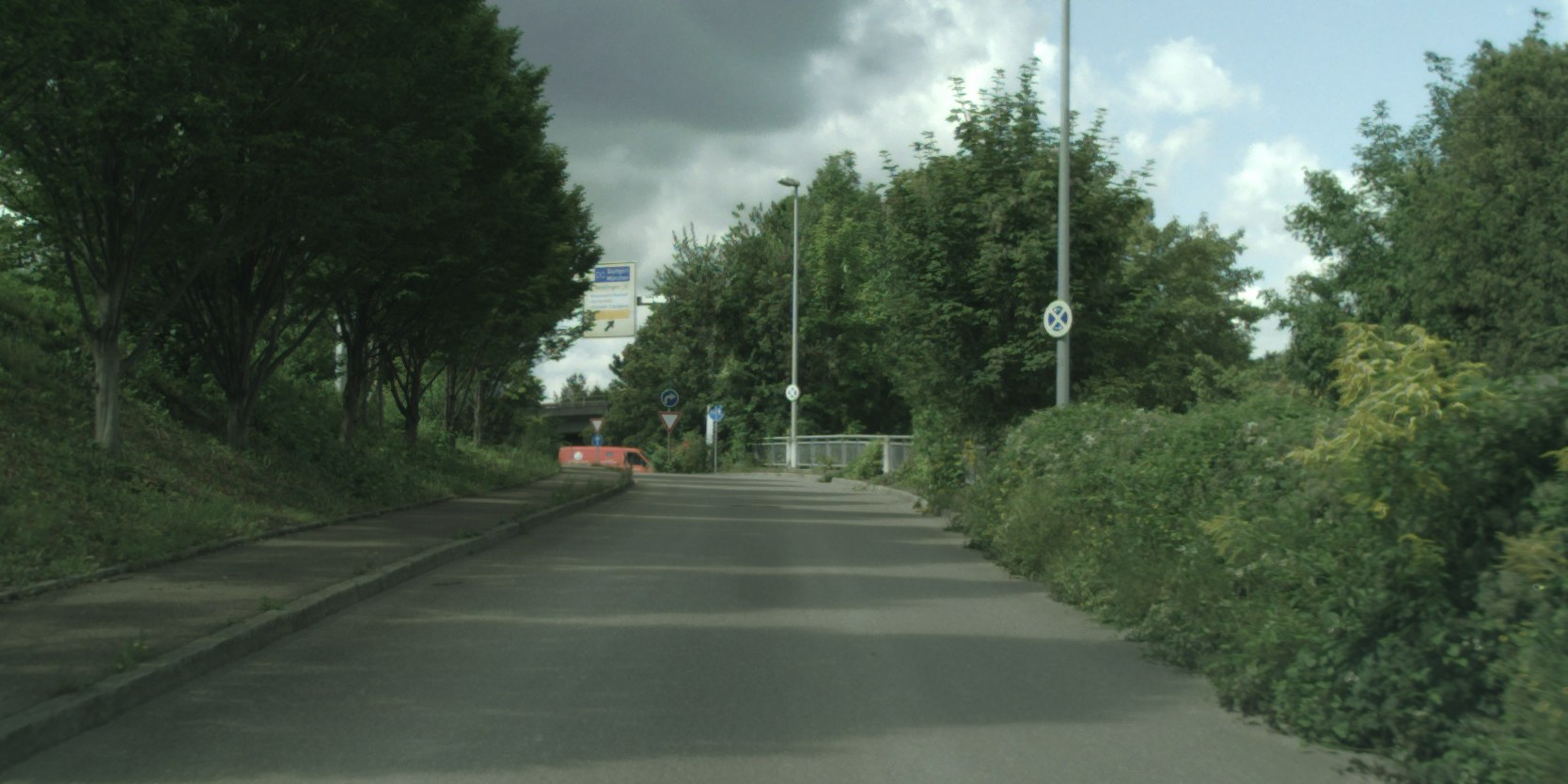}
    \NN
& terrain
    & Grass, all kinds of horizontally spreading vegetation, soil, or sand. These are areas that are
    not meant to be driven on. This label may also include a possibly adjacent curb.
    Single grass stalks or very small patches of grass are not annotated separately and thus are assigned
    to the label of the region they are growing on.
    &
    \includegraphics[height=3.15cm]{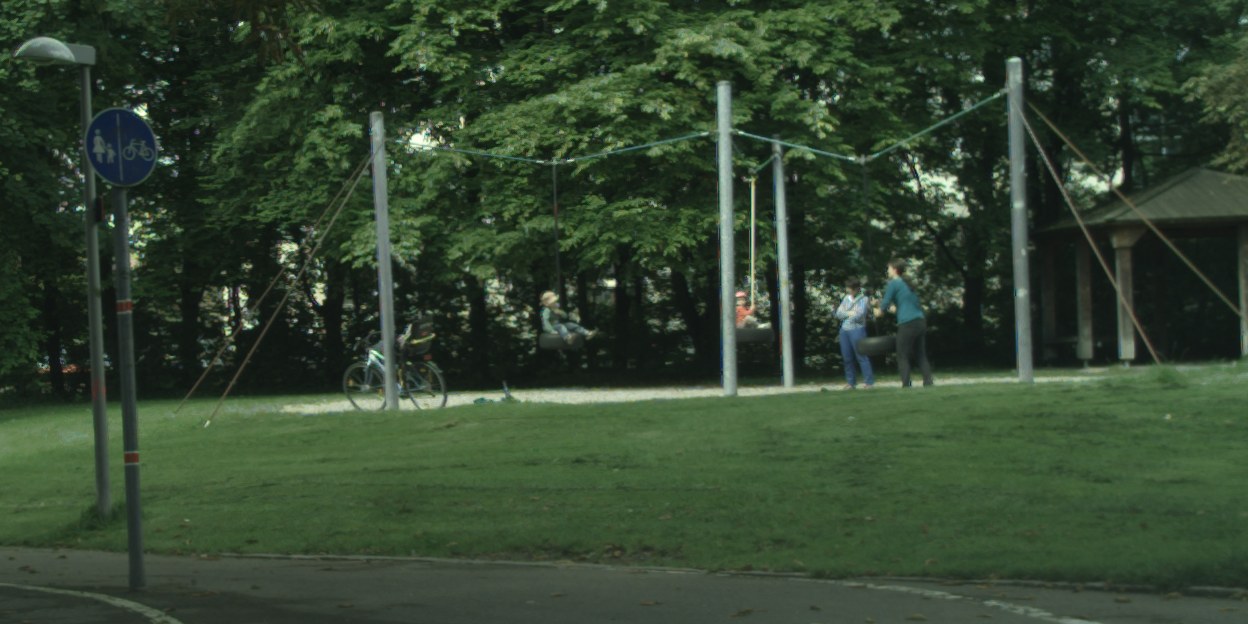}
    \groupsep
}

\ctable[
star,
caption = List of annotated classes including their definition and typical example.,
% label = tab:Annotated-labels,
pos = p,
width=\textwidth,
doinside=\small,
captionskip = 0.6ex,
continued
]
{p{0.09\textwidth}p{0.09\textwidth}m{0.36\textwidth}m{0.36\textwidth}}
{
    \tnote[1]{Single instance annotation available.}
    \tnote[2]{Not included in challenges.}
}{ \FL
Category & Class & Definition & Examples \ML
%%%%%%%%%%%%%%%%%%%%%%%%%%%%%%%%%%%%%%%%%%%%%%%%%%%%%%%%%%%%%%%%%%%%%%%%%%%%%%%%%%%%%%%%%
construction
& building
    & Includes structures that house/shelter humans, \eg low-rises, skyscrapers, bus
    stops, car ports. Translucent buildings made of glass still receive the label
    \textit{building}. Also includes scaffolding attached to buildings.
    &
    \includegraphics[height=3.15cm]{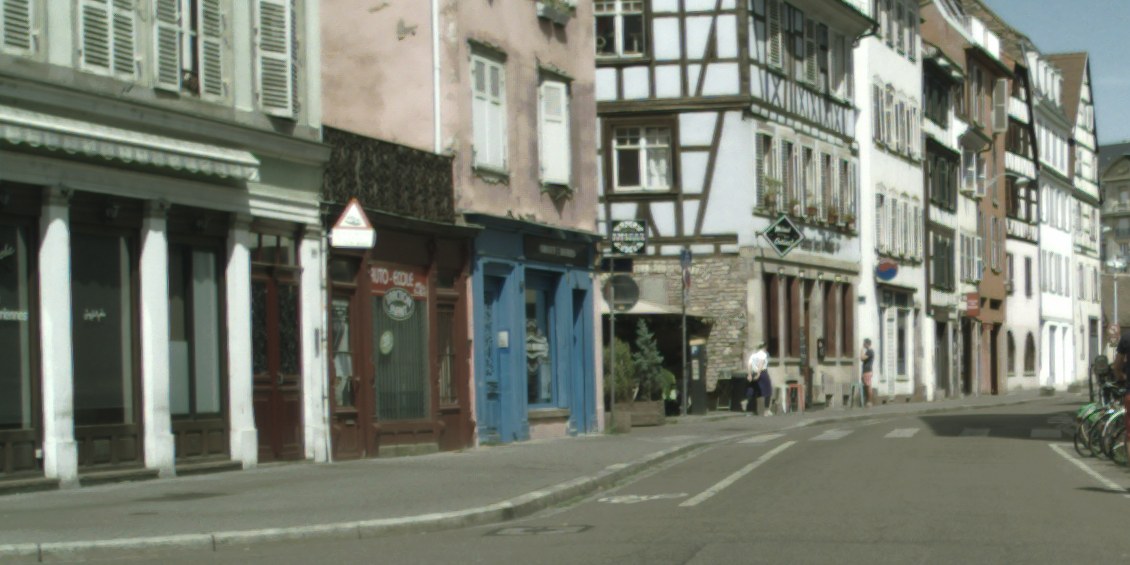}
    \NN
& wall
    & Individually standing walls that separate
    two (or more) outdoor areas, and do not provide support for a building.
    &
    \includegraphics[height=3.15cm]{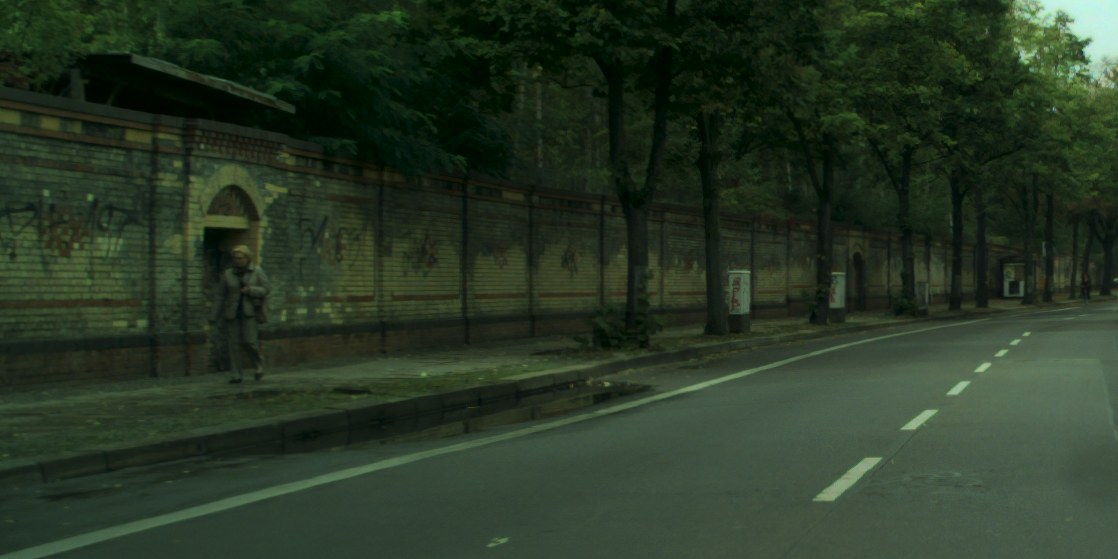}
    \NN
& fence
    & Structures with holes that separate two (or more) outdoor areas, sometimes temporary.
    &
    \includegraphics[height=3.15cm]{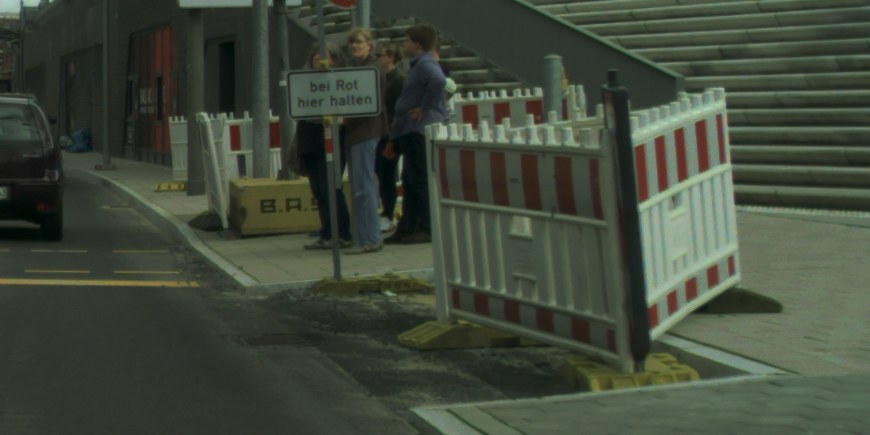}
    \NN
& guard rail\tmark[2]
    & Metal structure located on the side of the road to prevent serious accidents. Rare in inner cities, but
    occur sometimes in curves. Includes the bars holding the rails.
    &
    \includegraphics[height=3.15cm]{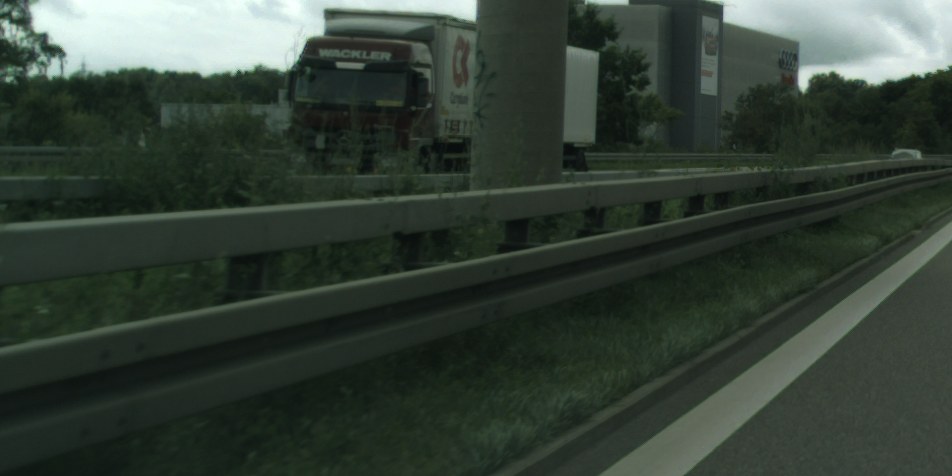}
\NN
& bridge\tmark[2]
    & Bridges (on which the ego-vehicle is not driving) including everything (fences, guard rails) permanently attached to them.
    &
    \includegraphics[height=3.15cm]{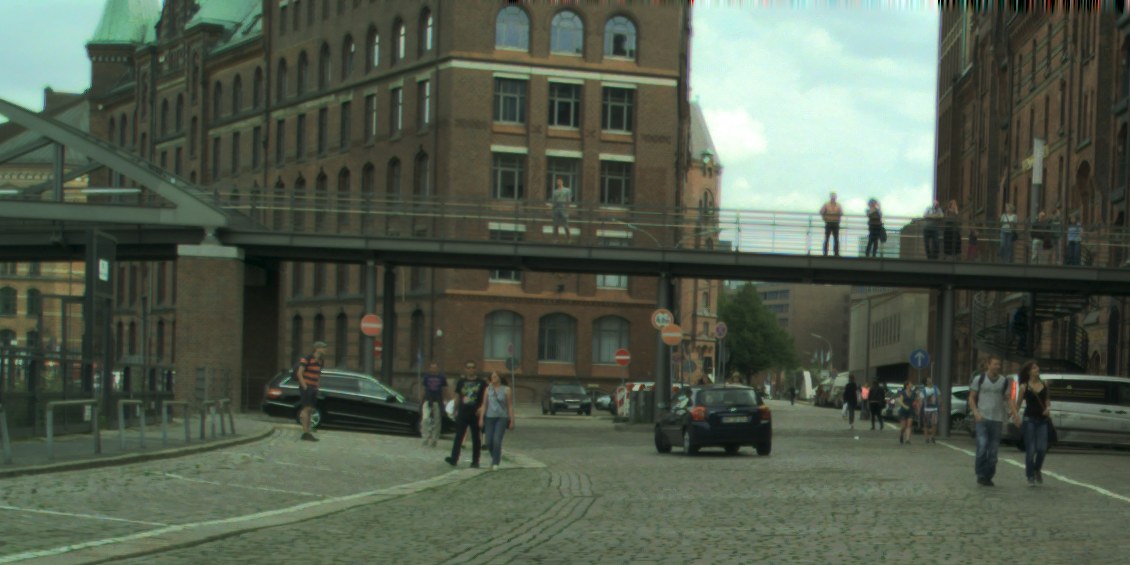}
\NN
& tunnel\tmark[2]
    & Tunnel walls and the (typically dark) space encased by the tunnel, but excluding vehicles.
    &
    \includegraphics[height=3.15cm]{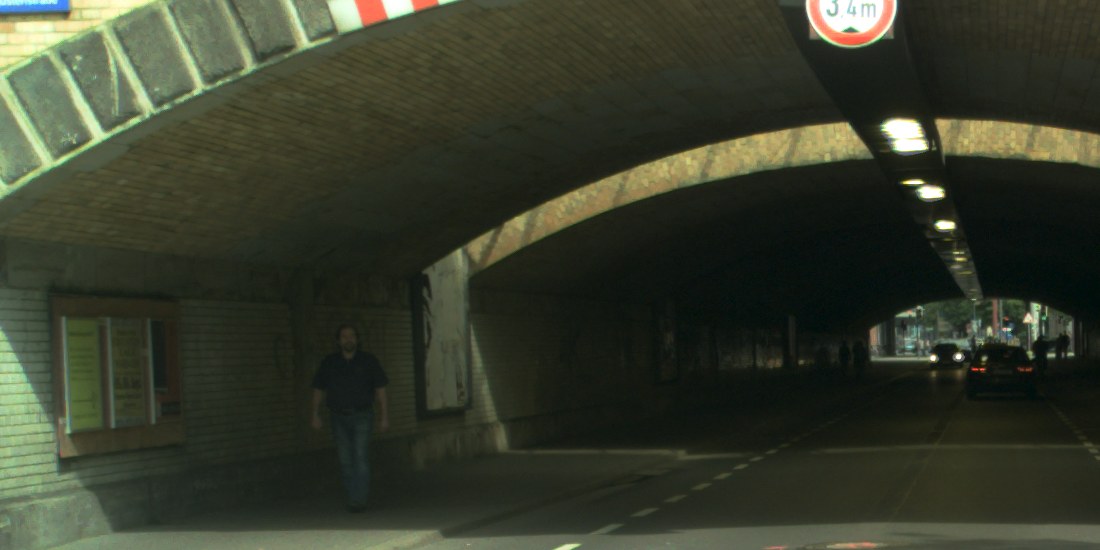}
    \groupsep
}

\ctable[
star,
caption = List of annotated classes including their definition and typical example.,
% label = tab:Annotated-labels,
pos = p,
width=\textwidth,
doinside=\small,
captionskip = 0.6ex,
continued
]
{p{0.09\textwidth}p{0.09\textwidth}m{0.36\textwidth}m{0.36\textwidth}}
{
    \tnote[1]{Single instance annotation available.}
    \tnote[2]{Not included in challenges.}
}{ \FL
Category & Class & Definition & Examples \ML
%%%%%%%%%%%%%%%%%%%%%%%%%%%%%%%%%%%%%%%%%%%%%%%%%%%%%%%%%%%%%%%%%%%%%%%%%%%%%%%%%%%%%%%%%
object
& traffic sign
    & Front part of signs installed by the state/city authority with the purpose of
    conveying information to drivers/cyclists/pedestrians, \eg traffic signs, parking signs, direction signs,
    or warning reflector posts.
    &
    \includegraphics[height=3.15cm]{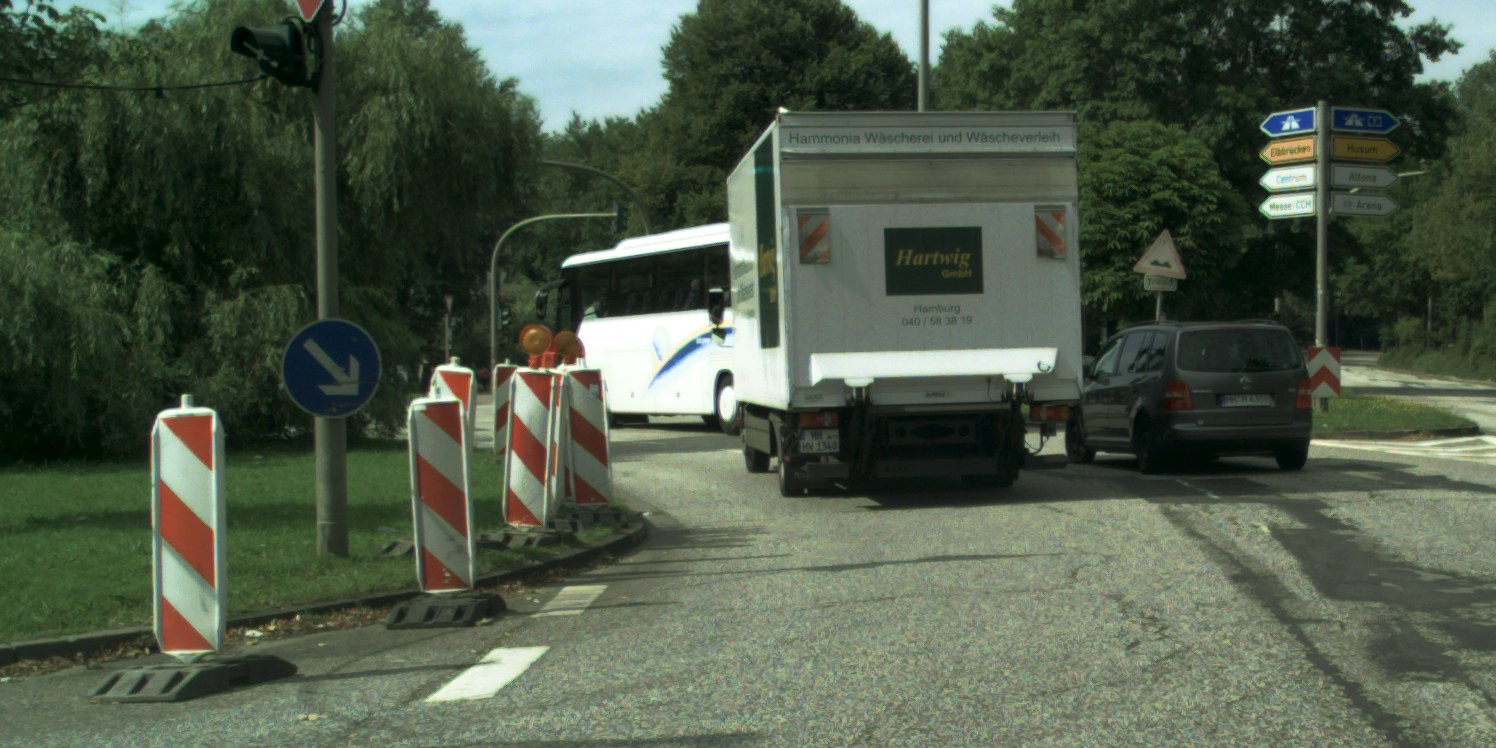}
    \NN
& traffic light
    & The traffic light box without its poles in all orientations and for all types of traffic participants, \eg
    regular traffic light, bus traffic light, train traffic light.
    &
    \includegraphics[height=3.15cm]{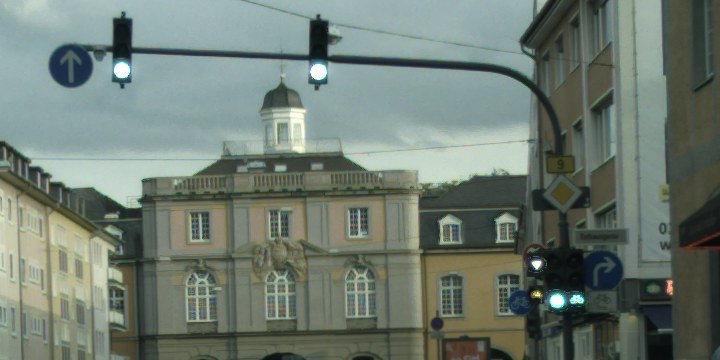}
    \NN
& pole
    & Small, mainly vertically oriented poles, \eg sign poles or traffic light poles.
    This does not include objects mounted on the pole, which have a larger
    diameter than the pole itself (\eg most street lights).
    &
    \includegraphics[height=3.15cm]{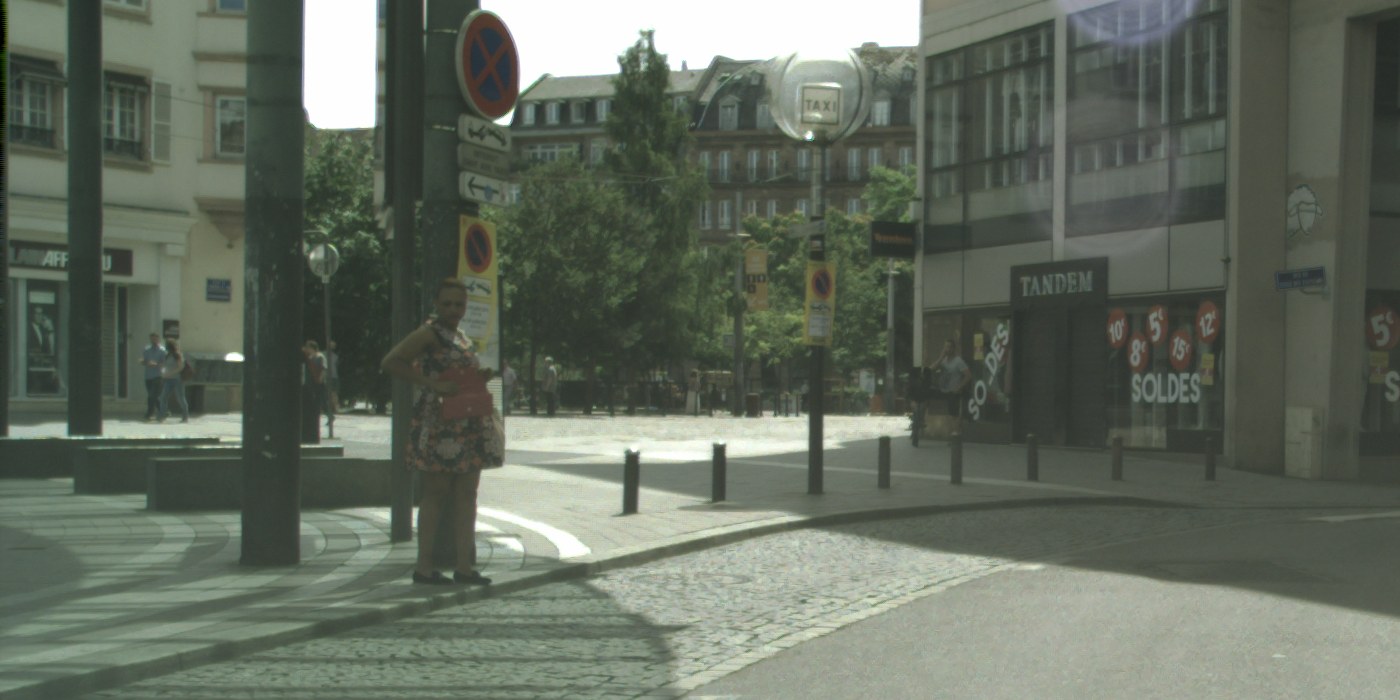}
    \NN
& pole group\tmark[2]
    & Multiple poles that are cumbersome to label individually, but where the background can be seen in their gaps.
    &
    \includegraphics[height=3.15cm]{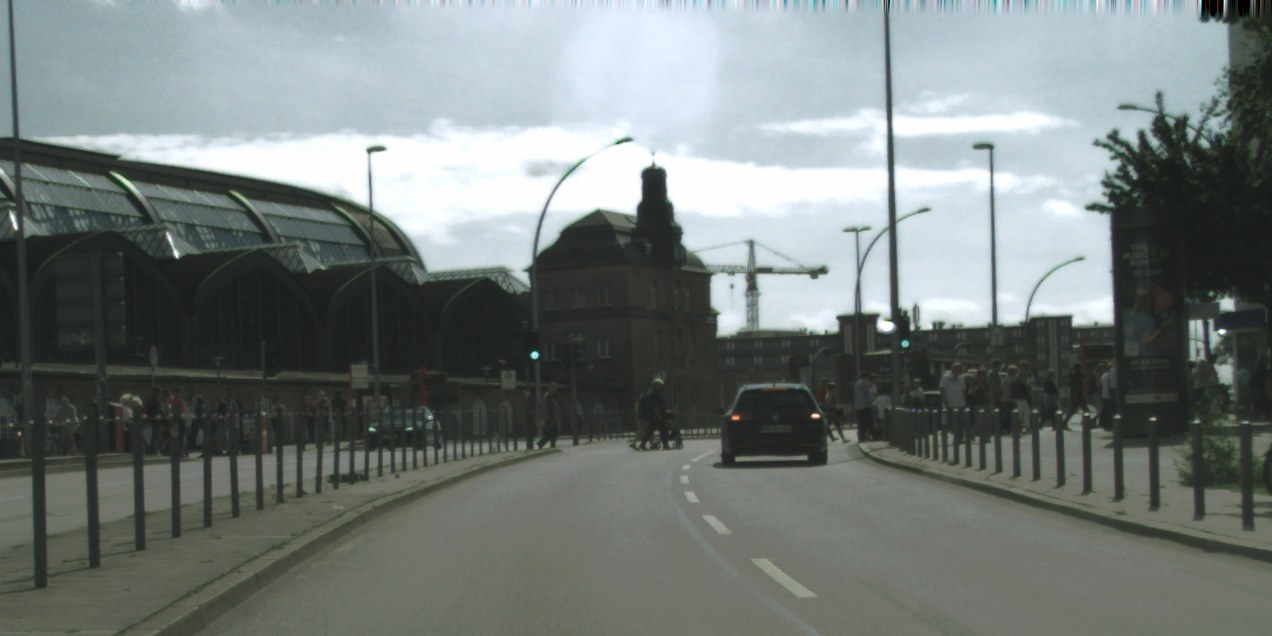}
    \groupsep
%%%%%%%%%%%%%%%%%%%%%%%%%%%%%%%%%%%%%%%%%%%%%%%%%%%%%%%%%%%%%%%%%%%%%%%%%%%%%%%%%%%%%%%%%
sky
& sky
    & Open sky (without tree branches/leaves)
    &
    \includegraphics[height=3.15cm]{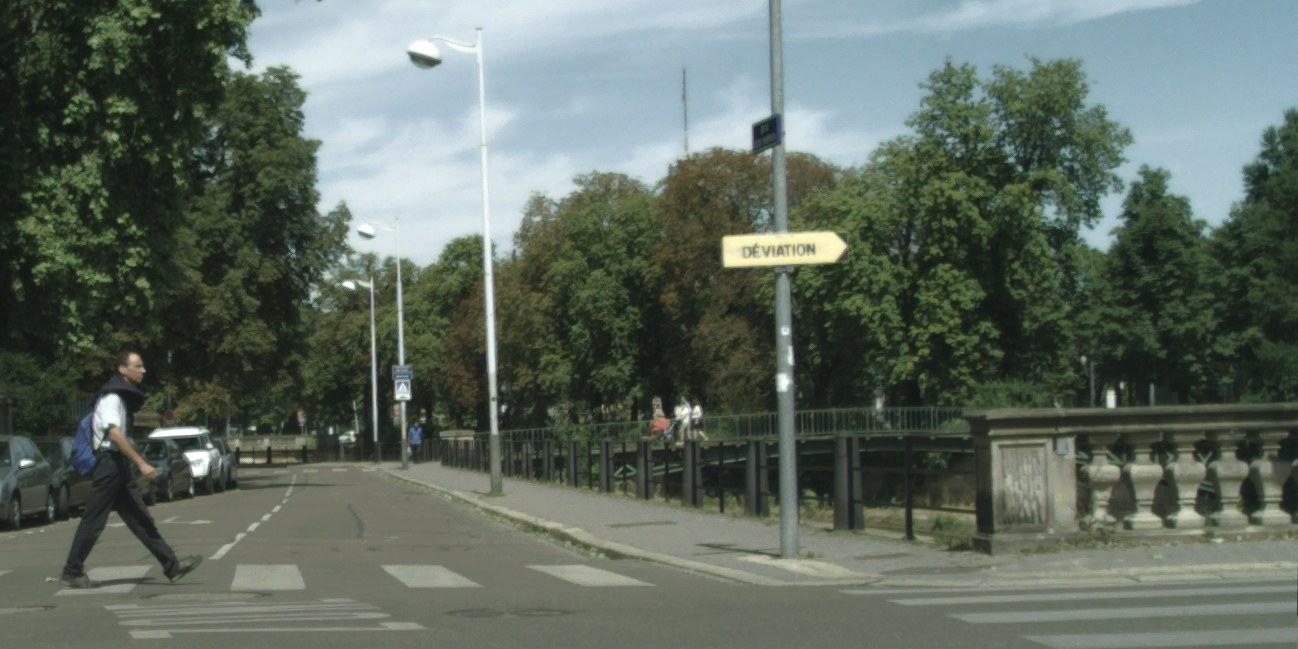}
    \groupsep
}

\ctable[
star,
caption = List of annotated classes including their definition and typical example.,
% label = tab:Annotated-labels,
pos = p,
width=\textwidth,
doinside=\small,
captionskip = 0.6ex,
continued
]
{p{0.09\textwidth}p{0.09\textwidth}m{0.36\textwidth}m{0.36\textwidth}}
{
    \tnote[1]{Single instance annotation available.}
    \tnote[2]{Not included in challenges.}
}{ \FL
Category & Class & Definition & Examples \ML
%%%%%%%%%%%%%%%%%%%%%%%%%%%%%%%%%%%%%%%%%%%%%%%%%%%%%%%%%%%%%%%%%%%%%%%%%%%%%%%%%%%%%%%%%
flat
& road
    & Horizontal surfaces on which cars usually drive, including road markings.
    Typically delimited by curbs, rail tracks, or parking areas. However, \textit{road} is
    not delimited by road markings and thus may include bicycle lanes or roundabouts.%, but not curbs
    & \includegraphics[height=3.15cm]{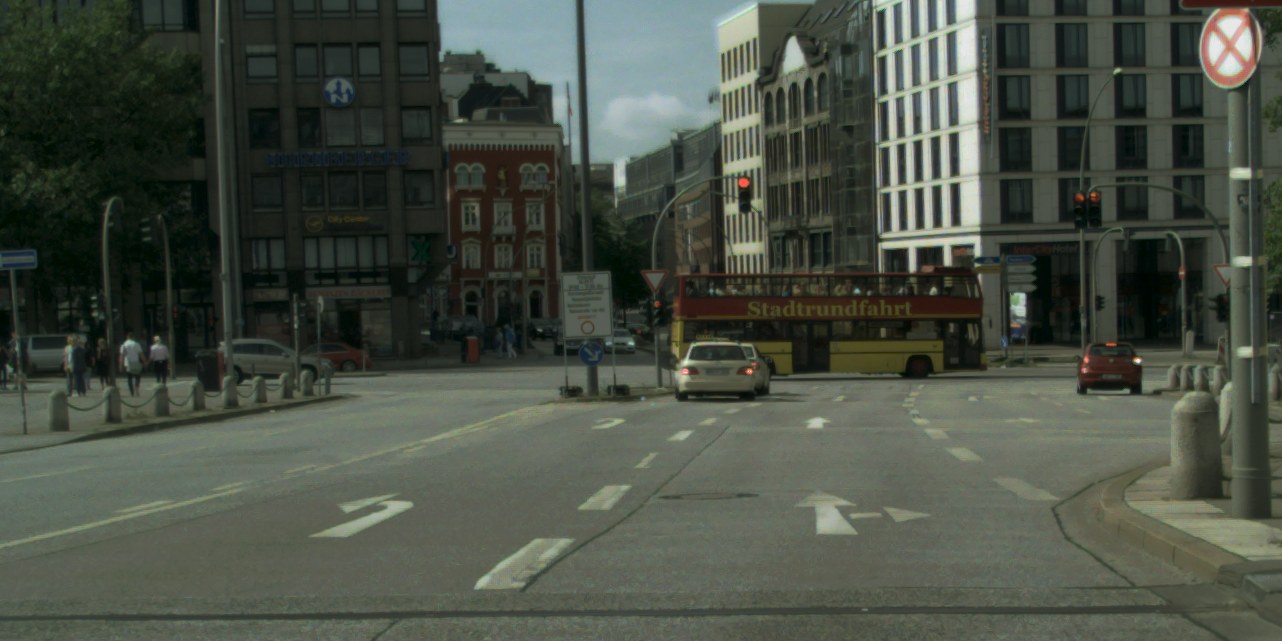} %could include example with bicycle lane
    \NN
& sidewalk
    & Horizontal surfaces designated for pedestrians or cyclists. Delimited from the
    road by some obstacle, \eg curbs or poles (might be small), but not only by
    markings. Often elevated compared to the road and often located at the side of a
    road. The curbs are included in the \textit{sidewalk} label. Also includes
    the walkable part of traffic islands, as well as pedestrian-only zones,
    where cars are not allowed to drive during regular business hours.
    If it's an all-day mixed pedestrian/car area, the correct label is \textit{ground}.
    & \includegraphics[height=3.15cm]{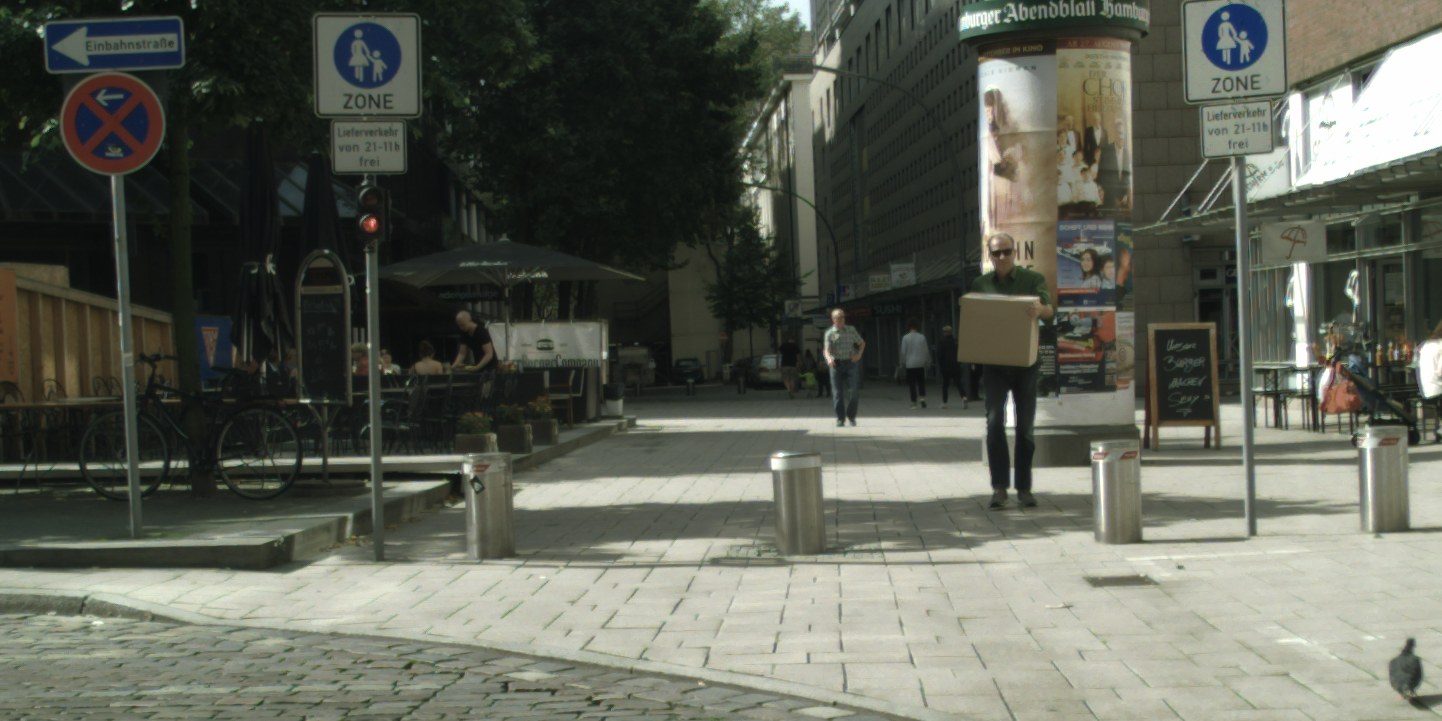}
    \NN
& parking\tmark[2]
    & Horizontal surfaces that are intended for parking and separated from the
    road, either via elevation or via a different texture/material, but not separated
    merely by markings.
    &
    \includegraphics[height=3.15cm]{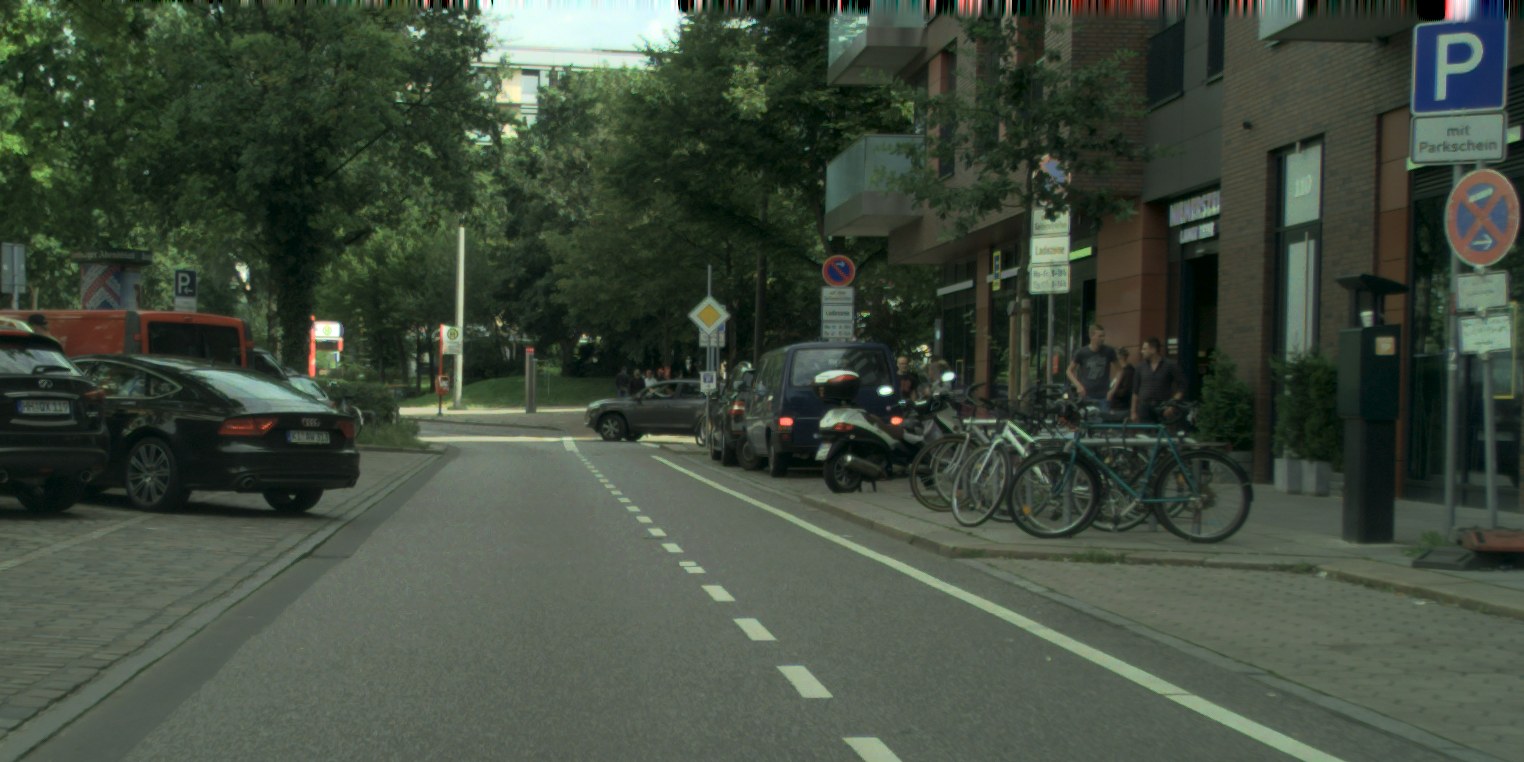}
    \NN
& rail track\tmark[2]
    & Horizontal surfaces on which only rail cars can normally drive. If rail tracks for trams
    are embedded in a standard road, they are included in the \textit{road} label.
    &
    \includegraphics[height=3.15cm]{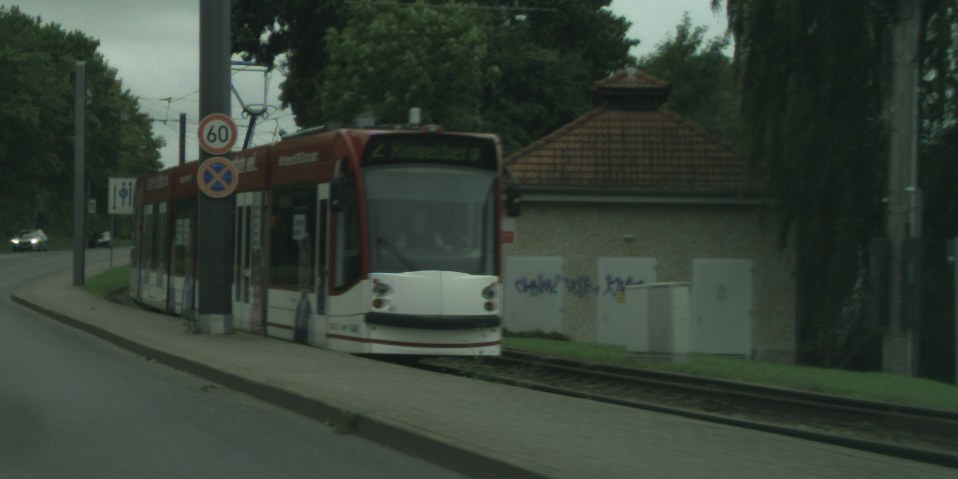}
    \groupsep
}

\ctable[
star,
caption = List of annotated classes including their definition and typical example.,
% label = tab:Annotated-labels,
pos = p,
width=\textwidth,
doinside=\small,
captionskip = 0.6ex,
continued
]
{p{0.09\textwidth}p{0.09\textwidth}m{0.36\textwidth}m{0.36\textwidth}}
{
    \tnote[1]{Single instance annotation available.}
    \tnote[2]{Not included in challenges.}
}{ \FL
Category & Class & Definition & Examples \ML
%%%%%%%%%%%%%%%%%%%%%%%%%%%%%%%%%%%%%%%%%%%%%%%%%%%%%%%%%%%%%%%%%%%%%%%%%%%%%%%%%%%%%%%%%
void
& ground\tmark[2]
    & All other forms of horizontal ground-level structures that do not match any of
    the above, for example mixed zones (cars and pedestrians), roundabouts that are
    flat but delimited from the road by a curb, or in general a fallback label for
    horizontal surfaces that are difficult to classify, \eg due to having a
    dual purpose.
    & \includegraphics[height=3.15cm]{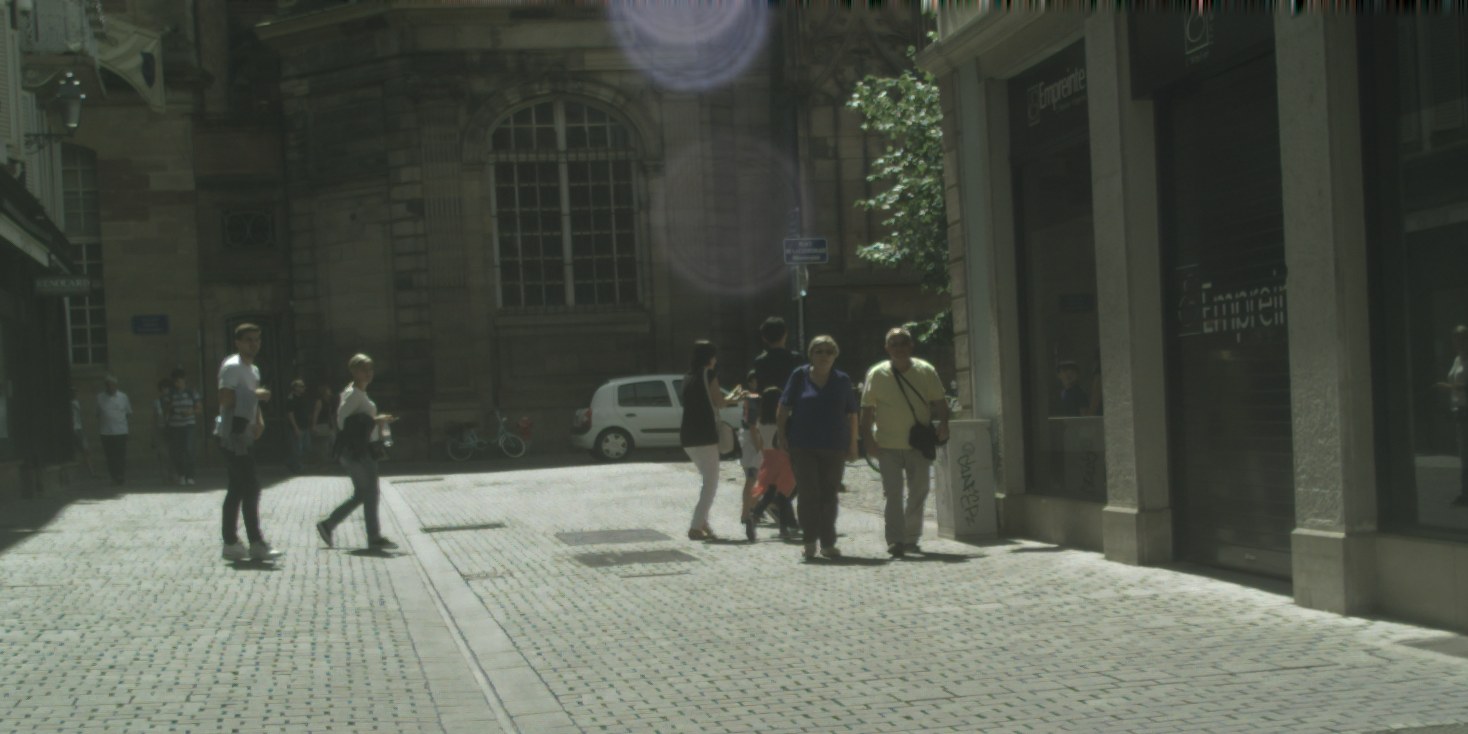}
    \NN
& dynamic\tmark[2]
    & Movable objects that do not correspond to any of the other non-void categories
    and might not be in the same position in the next day/hour/minute, \eg movable
    trash bins, buggies, luggage, animals, chairs, or tables.
    & \includegraphics[height=3.15cm]{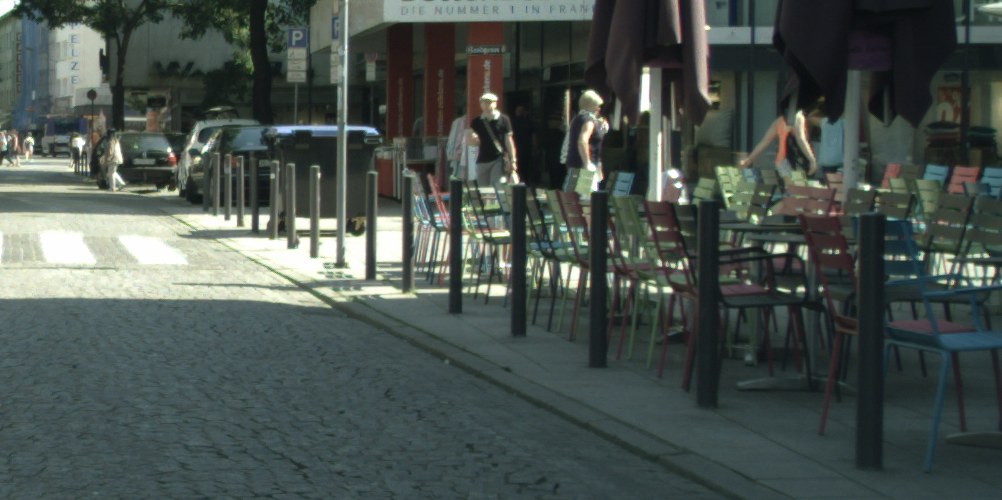}
    \NN
& static\tmark[2]
    & This includes areas of the image that are difficult to identify/label due to
    occlusion/distance, as well as non-movable objects that do not match any of the
    non-void categories, \eg mountains, street lights, reverse sides of
    traffic signs, or permanently mounted commercial signs.
    & \includegraphics[height=3.15cm]{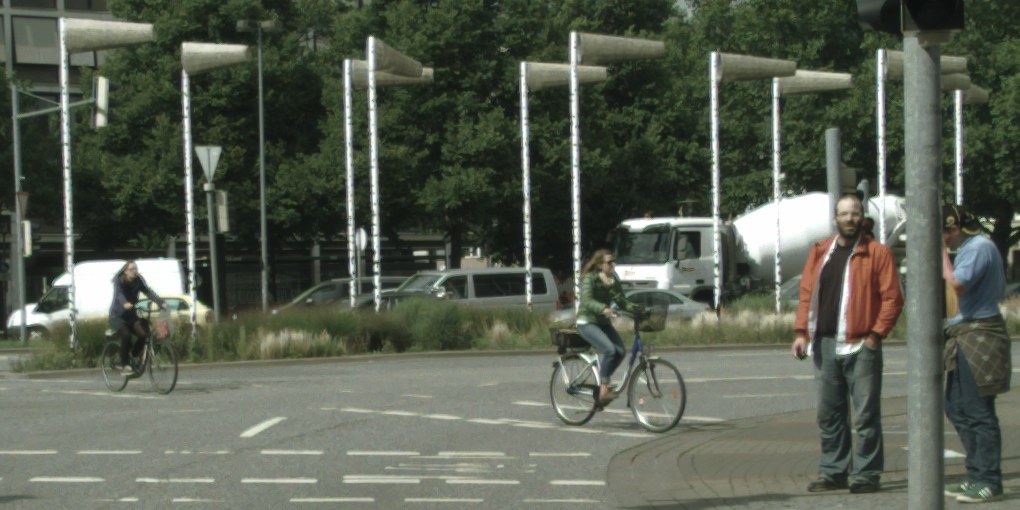}
    \NN
& ego vehicle\tmark[2]
    & Since a part of the vehicle from which our data was recorded is visible in all
    frames, it is assigned to this special label. This label is also available at test time.
    &
    \NN[1em]
& unlabeled\tmark[2]
    &Pixels that were not explicitly assigned to a label.
    &
    \NN[1em]
& out of roi\tmark[2]
    &Narrow strip of \SI{5}{pixels} along the image borders that is not considered for training or evaluation.
    This label is also available at test-time.
    &
    \NN[3em]
& rectification border\tmark[2]
    &Areas close to the image border that contain artifacts resulting from the stereo pair rectification.
    This label is also available at test time.
    &
    \LL
}

%%%%%%%%%%%%%%%%%%%%%%%%%%%%%%%%%%%%%

%%%%%%%%%%%%%%%%%%%%%%%%%%%%%%%%%%%%%

%%%%%%%%%%%%%%%%%%%%%%%%%%%%%%%%%%%%%

%%%%%%%%%%%%%%%%%%%%%%%%%%%%%%%%%%%%%

\section{Example Annotations}
\label{sec:examples}

\Cref{fig:exampleannotations} presents several examples of annotated
frames from our dataset that exemplify its diversity and difficulty.
All examples are taken from the \textit{train} and \textit{val} splits
and were chosen by searching for the extremes in terms of the number
of traffic participant instances in the scene;
see \cref{fig:exampleannotations} for details.

% \TODO{choose interesting frames where from various cities, preferably where a lot is going on, but perhaps also a few depicting sparsely populated scenes}

\begin{figure}[t]
    \centering
    \includegraphics[width=0.8\columnwidth]{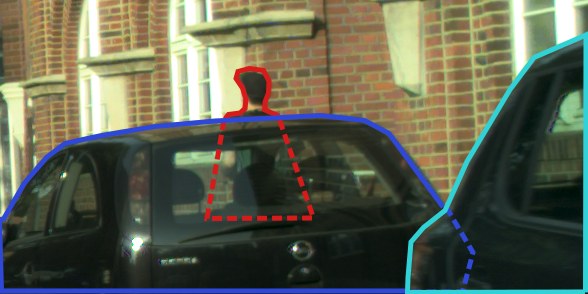}
    \caption{Exemplary labeling process. Distant objects are annotated
      first and subsequently their occluders. This ensures the
      boundary between these objects to be shared and consistent.}
    \label{fig:labelProcess}
\end{figure}

\begin{figure*}[p]
    \captionsetup[subfigure]{aboveskip=2pt,belowskip=2pt}
    \centering%
    \begin{subfigure}[b]{\linewidth}
        \begin{overpic}[width=\textwidth]{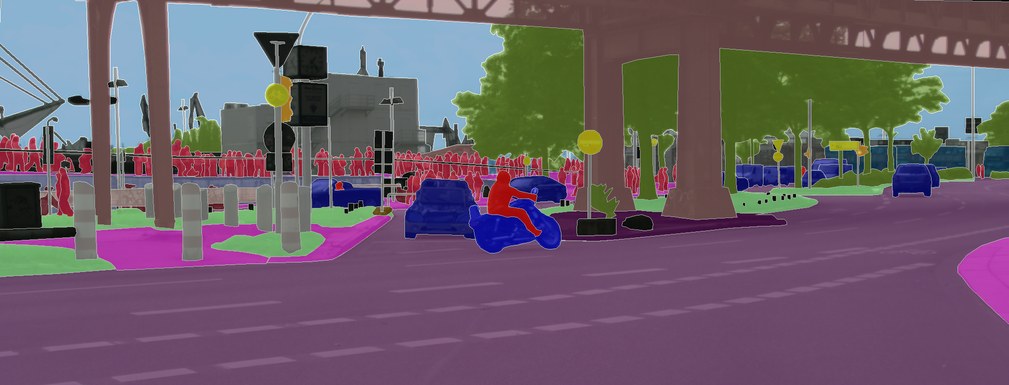}
        \put (50,2) {\makebox(0,0){\textcolor{white}{\textsf{\footnotesize Largest number of instances and persons}}}}
        \end{overpic}
    \end{subfigure}\\
    \begin{subfigure}[b]{0.499\linewidth}
        \begin{overpic}[width=\textwidth]{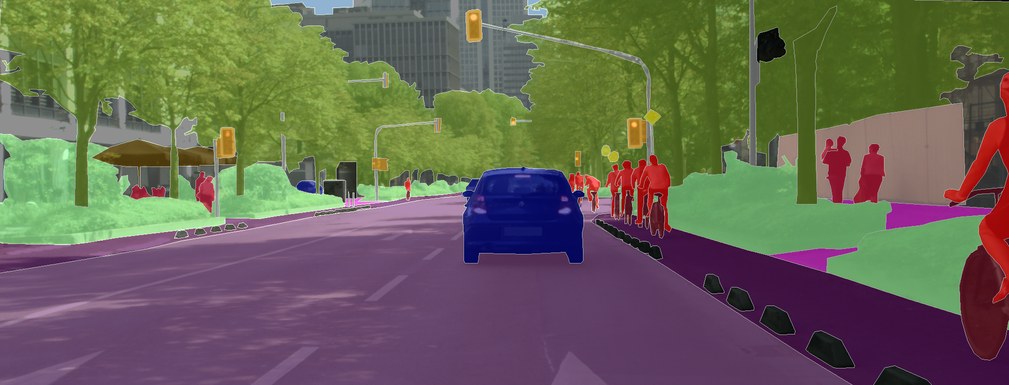}
        \put (50,4) {\makebox(0,0){\textcolor{white}{\textsf{\footnotesize Largest number of riders}}}}
        \end{overpic}
    \end{subfigure}\hfill
    \begin{subfigure}[b]{0.499\linewidth}
        \begin{overpic}[width=\textwidth]{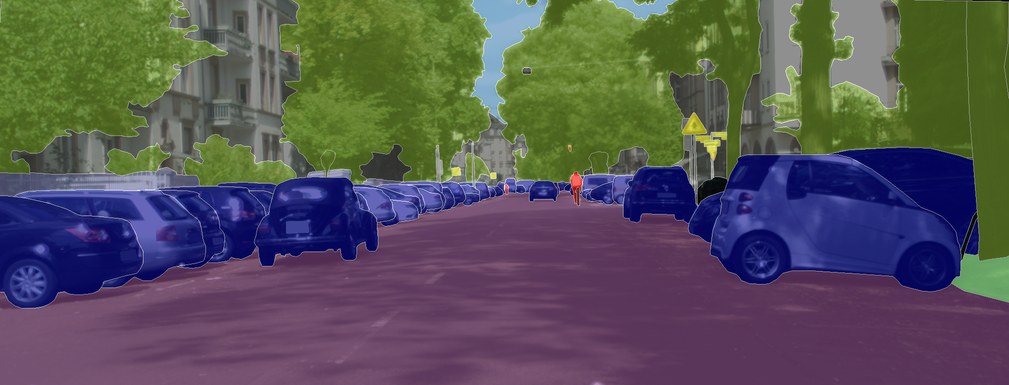}
        \put (50,4) {\makebox(0,0){\textcolor{white}{\textsf{\footnotesize Largest number of cars}}}}
        \end{overpic}
    \end{subfigure}\\
    \begin{subfigure}[b]{0.499\linewidth}
        \begin{overpic}[width=\textwidth]{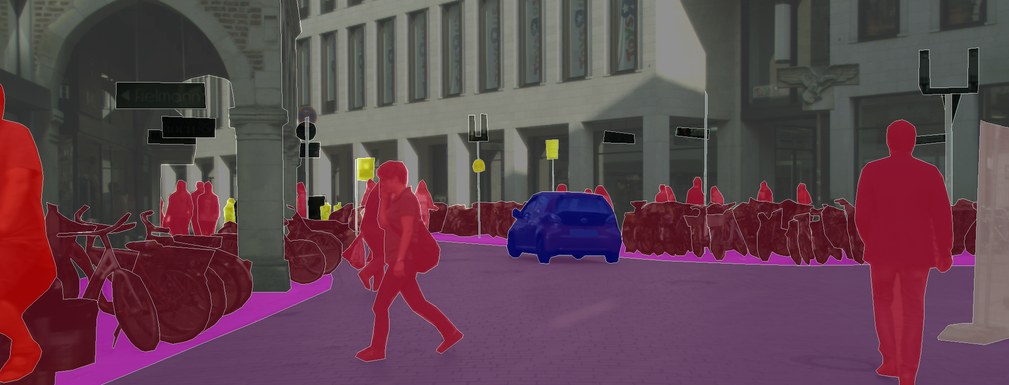}
        \put (50,4) {\makebox(0,0){\textcolor{white}{\textsf{\footnotesize Largest number of bicycles}}}}
        \end{overpic}
    \end{subfigure}\hfill
    \begin{subfigure}[b]{0.499\linewidth}
        \begin{overpic}[width=\textwidth]{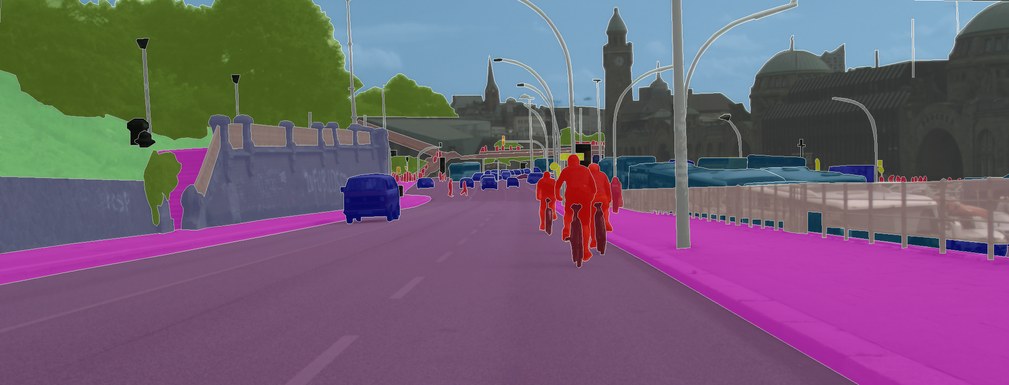}
        \put (50,4) {\makebox(0,0){\textcolor{white}{\textsf{\footnotesize Largest number of buses}}}}
        \end{overpic}
    \end{subfigure}\\
    \begin{subfigure}[b]{0.499\linewidth}
        \begin{overpic}[width=\textwidth]{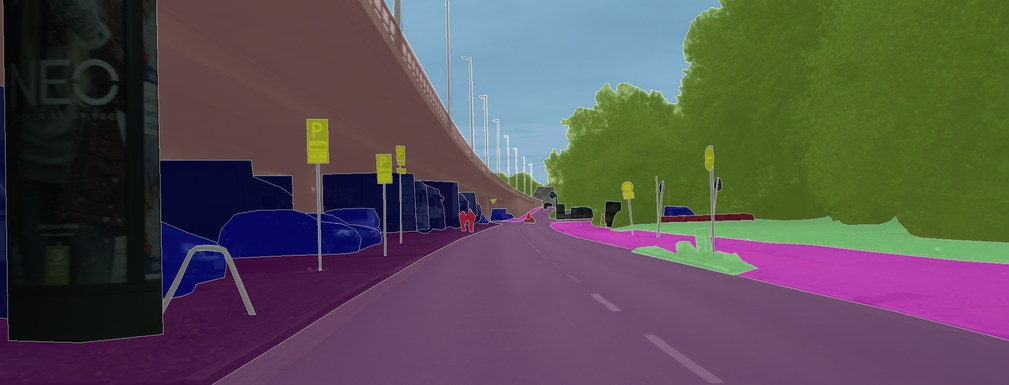}
        \put (50,4) {\makebox(0,0){\textcolor{white}{\textsf{\footnotesize Largest number of trucks}}}}
        \end{overpic}
    \end{subfigure}\hfill
    \begin{subfigure}[b]{0.499\linewidth}
        \begin{overpic}[width=\textwidth]{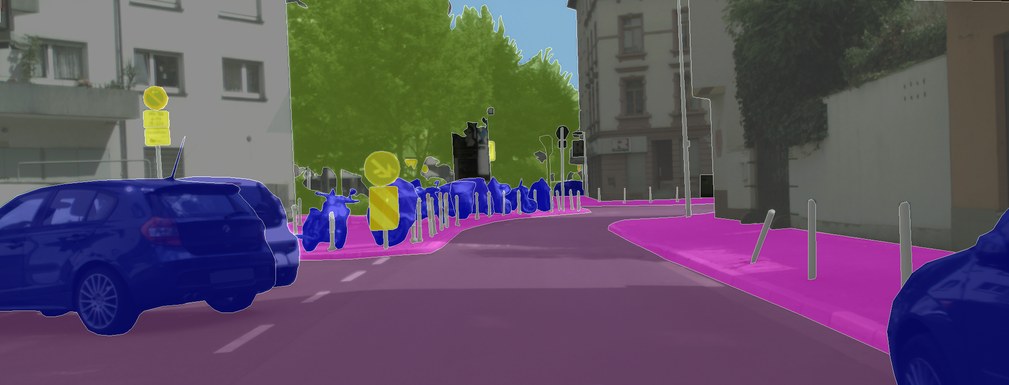}
        \put (50,4) {\makebox(0,0){\textcolor{white}{\textsf{\footnotesize Largest number of motorcycles}}}}
        \end{overpic}
    \end{subfigure}\\
    \begin{subfigure}[b]{0.499\linewidth}
        \begin{overpic}[width=\textwidth]{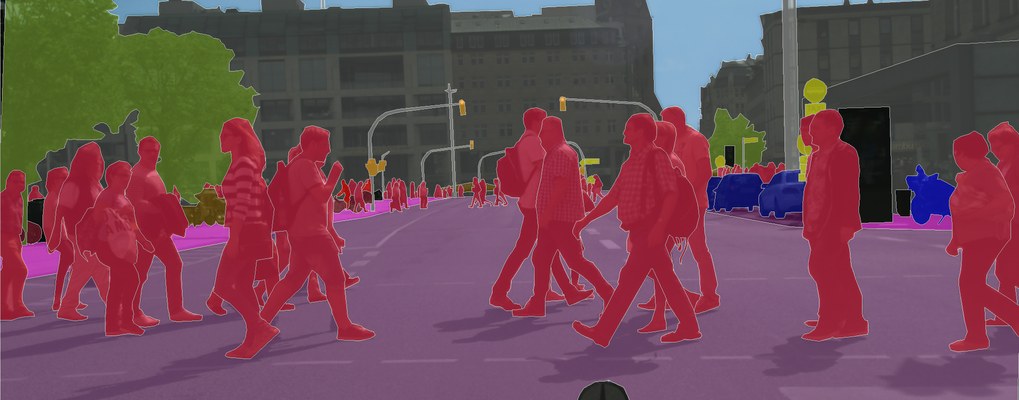}
        \put (50,4) {\makebox(0,0){\textcolor{white}{\textsf{\footnotesize Large spatial variation of persons}}}}
        \end{overpic}
    \end{subfigure}\hfill
    \begin{subfigure}[b]{0.499\linewidth}
        \begin{overpic}[width=\textwidth]{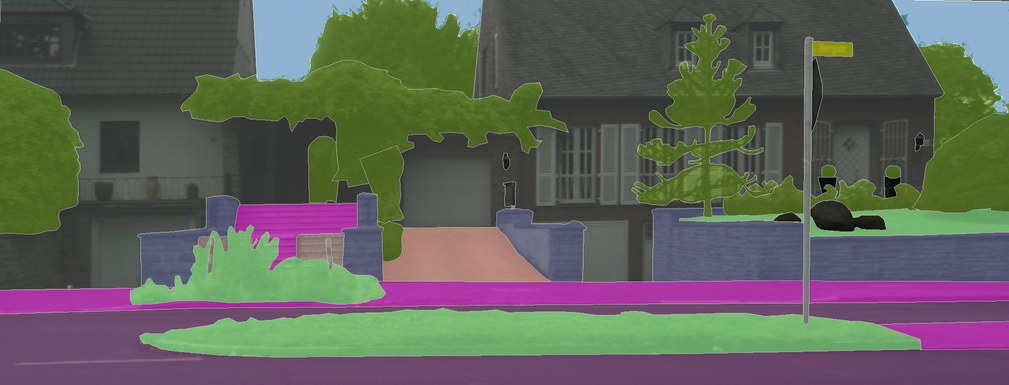}
        \put (50,4) {\makebox(0,0){\textcolor{white}{\textsf{\footnotesize Fewest number of instances}}}}
        \end{overpic}
    \end{subfigure}\\
    \vspace{2mm}
    \caption{Examples of our annotations on various images of our
      \textit{train} and \textit{val} sets. The images were selected
      based on criteria overlayed on each image.}
    \label{fig:exampleannotations}
\end{figure*}

% \section{Additional Statistics}
% \label{sec:statistics}

% In this section, we present additional dataset statistics that didn't make it into the main paper due to space constraints.

% \TODO{more statistics (distance between frames (time/space), object density, etc.)}

\section{Detailed Results}
In this section, we present additional details regarding our control
experiments and baselines.
Specifically, we give individual class scores that complement the
aggregated scores in the main paper.
Moreover, we provide details on the training procedure for all
baselines.
Finally, we show additional qualitative results of all methods.

\subsection{Semantic labeling}
\label{subsec:classlevel}

\Cref{tab:pixellevel_control,tab:pixellevel_methods} list all
individual class-level $\iou$ scores for all control experiments and
baselines.
\Cref{tab:instancelevelfromclasses_control,tab:instancelevelfromclasses_methods}
give the corresponding instance-normalized $\iiou$ scores.
In addition, \cref{fig:mostallcontrol,fig:mostallbaselines} contain qualitative examples of these methods.

\myparagraph{Basic setup.}
All baselines relied on single frame, monocular LDR images and were
pretrained on ImageNet~\cite{Russakovsky2014}, \ie their underlying
CNN was generally initialized with ImageNet VGG
weights~\cite{Simonyan2014}.
Subsequently, the CNNs were finetuned on Cityscapes using the
respective portions listed in \cref{tab:baselineResults}.
In our own FCN~\cite{Long2015} experiments, we additionally
investigated first pretraining on PASCAL-Context~\cite{Mottaghi2014},
but found this to not influence performance given a sufficiently large
number of training iterations.
Most baselines applied a subsampling of the input image, \cf
\cref{tab:baselineResults}, probably due to time or memory
constraints.
Only Adelaide~\cite{Lin2015}, Dilated10~\cite{Yu2016}, and our FCN experiments were conducted on
the full-resolution images.
In the first case, a new random patch of size $614\times 614$ pixels
was drawn at each iteration. In our FCN training, we split each image
into two halves (left and right) with an overlap that is sufficiently
large considering the network's receptive field.

\myparagraph{Own baselines.}
The training procedure of all our FCN experiments
follows~\cite{Long2015}.
We use three-stage training with subsequently smaller strides, \ie
first FCN-32s, then FCN-16s, and then FCN-8s, always initializing with
the parameters from the previous stage.
We add a \nth{4} stage for which we reduce the learning rate by a factor
of \num{10}.
The training parameters are identical to those publicly available for
training on PASCAL-Context~\cite{Mottaghi2014}, except that we reduce
the learning rate to account for the increased image resolution.
Each stage is trained until convergence on the validation set; pixels
with \textit{void} ground truth are ignored such that they do not
induce any gradient.
Eventually, we retrain on \textit{train} and \textit{val} together
with the same number of epochs, yielding \num{243250}, \num{69500},
\num{62550}, and \num{5950} iterations for stages \num{1} through
\num{4}.
Note that each iteration corresponds to half of an image (see above).
For the variant with factor \num{2} downsampling, no image splitting
is necessary, yielding \num{80325}, \num{68425}, \num{35700}, and
\num{5950} iterations in the respective stages.
The variant only trained on \textit{val} (full resolution) uses
\textit{train} for validation, leading to \num{130000}, \num{35700},
\num{47600}, and \num{0} iterations in the \num{4} stages.
Our last FCN variant is trained using the coarse annotations only, with
\num{386750}, \num{113050}, \num{35700}, and \num{0} iterations in the
respective stage;  pixels with \textit{void} ground truth are ignored
here as well.

\myparagraph{\nth{3}-party baselines.}
\emph{Note that for the following descriptions of the \nth{3}-party
  baselines, we have to rely on author-provided information.}

SegNet~\cite{Badrinarayanan2015} training for both the \textit{basic} and \textit{extended} variant was performed until convergence, yielding approximately \num{50} epochs. Inference takes \SI{0.12}{\second} per image.

DPN~\cite{Liu2015} was trained using the original procedure, while using all available Cityscapes annotations.

For training \textit{CRF as RNN}~\cite{Zheng2015}, an FCN-32s model was
trained for \num{3} days on \textit{train} using a GPU.
Subsequently an FCN-8s model was trained for \num{2} days, and eventually the model was
further finetuned including the CRF-RNN layers. Testing takes \SI{0.7}{\second} on
half-resolution images.

For training DeepLab on the fine annotations, denoted
\textit{DeepLab-LargeFOV-Strong}, the authors applied the training
procedure from~\cite{Chen2015}.
The model was trained on \textit{train} for \num{40000} iterations
until convergence on \textit{val}.
Then \textit{val} was included in the training set for another
\num{40000} iterations.
In both cases, a mini-batch size of \num{10} was applied.
Each training iteration lasts \SI{0.5}{\second}, while inference
including the dense CRF takes \SI{4}{\second} per image.
The DeepLab variant including our coarse annotations, termed
\textit{DeepLab-LargeFOV-StrongWeak}, followed the protocol
in~\cite{Papandreou2015} and is initialized from the
\textit{DeepLab-LargeFOV-Strong} model.
Each mini-batch consists of \num{5} finely and \num{5} coarsely
annotated images and training is performed for \num{20000} iterations
until convergence on \textit{val}.
Then, training was continued for another \num{20000} iterations on
\textit{train} and \textit{val}.

Adelaide~\cite{Lin2015} was trained for
\num{8} days using random crops of the input image as described
above.
Inference on a single image takes \SI{35}{\second}.

The best performing baseline, Dilated10~\cite{Yu2016},
is a convolutional network that consists of a front-end
prediction module and a context aggregation module. The front-end
module is an adaptation of the VGG-16 network based on dilated
convolutions. The context module uses dilated convolutions to
systematically expand the receptive field and aggregate contextual
information. This module is derived from the ``Basic" network, where
each layer has $C=19$ feature maps. The
total number of layers in the context module is 10, hence the name
Dilation10. The increased number of layers in the context module (10
for Cityscapes versus 8 for PASCAL VOC) is due to the higher input
resolution. The complete Dilation10 model is a pure convolutional
network: there is no CRF and no structured prediction.
The Dilation10 network was trained in three stages. First, the
front-end prediction module was trained for \num{40000} iterations on randomly
sampled crops of size $628\!\times\!628$, with learning rate
$10^{-4}$, momentum $0.99$, and batch size $8$. Second, the context
module was trained for \num{24000} iterations on whole (uncropped) images,
with learning rate $10^{-4}$, momentum $0.99$, and batch size \num{100}.
Third, the complete model (front-end + context) was jointly trained
for \num{60000} iterations on halves of images (input size
$1396\!\times\!1396$, including padding), with learning rate
$10^{-5}$, momentum $0.99$, and batch size 1.

\subsection{Instance-level semantic labeling}
\label{subsec:instancelevel}

%\TODO{(instance segmentation) detailed tables with results per-class, example results (must-have since promised in the paper), more training/testing details}

% This section contains more detailed results for all instance-level semantic
% labeling baselines presented in the main paper, as well as relevant training
% and testing details.

% Table \ref{tab:instancelevel_baselines} contains detailed results for each model,
% and figure \ref{fig:instanceresults} contains example results.

For our instance-level semantic labeling baselines and control
experiments, we rely on Fast R-CNN \cite{Girshick15ICCV} and
proposal regions from either MCG
(Multiscale Combinatorial Grouping \cite{cArbelaez14}) or from the ground truth annotations.

We use the standard training and testing parameters for Fast R-CNN. Training
starts with a model pre-trained on ImageNet~\cite{Russakovsky2014}.
We use a learning rate of \num{0.001} and
stop when the validation error plateaus after \num{120000} iterations.

At test time, one score per class is assigned to each object proposal.
% We then
% iterate over the different classes, and for each class select the set of
% detections for which the respective class score is above a threshold ($0.8$).
% This threshold was chosen using the validation error.
% To these detections we apply non-maximum suppression, eliminating those
% detections that have an IoU larger than $0.3$ with a higher-scoring detection.
% The remaining detections are added to the final set.
Subsequently, thresholding and non-maximum suppression is applied and either the bounding boxes,
the original proposal regions or their convex hull are used to generate the predicted masks of each instance.
Quantitative results of all classes can be found in \cref{tab:instancelevel_baselines_ap,tab:instancelevel_baselines_ap50,tab:instancelevel_baselines_ap100m,tab:instancelevel_baselines_ap50m}
and qualitative results in \cref{fig:instanceresults}.

\clearpage

\ctable[
star,
caption = {Detailed results of our control experiments for the
  pixel-level semantic labeling task in terms of the $\iou$ score on
  the class level.
  All numbers are given in percent.
  See the main paper for details on the listed methods.},
label = tab:pixellevel_control,
pos = p,
doinside=\scriptsize
]{U VVVVVVVVVVVVVVVVVVV W}{
}{
\FL
& \rotatedlabel{road}
& \rotatedlabel{sidewalk}
& \rotatedlabel{building}
& \rotatedlabel{wall}
& \rotatedlabel{fence}
& \rotatedlabel{pole}
& \rotatedlabel{traffic light}
& \rotatedlabel{traffic sign}
& \rotatedlabel{vegetation}
& \rotatedlabel{terrain}
& \rotatedlabel{sky}
& \rotatedlabel{person}
& \rotatedlabel{rider}
& \rotatedlabel{car}
& \rotatedlabel{truck}
& \rotatedlabel{bus}
& \rotatedlabel{train}
& \rotatedlabel{motorcycle}
& \rotatedlabel{bicycle}
& \rotatedlabel{\textbf{mean $\iou$}} \ML
static fine (SF)                     & $80.0$ & $13.2$ & $40.3$ & $ 0.0$ & $ 0.0$ & $ 0.0$ & $ 0.0$ & $ 0.0$ & $12.5$ & $ 0.0$ & $22.1$ & $ 0.0$ & $ 0.0$ & $23.4$ & $ 0.0$ & $ 0.0$ & $ 0.0$ & $ 0.0$ & $ 0.0$ & $10.1$ \NN % staticPrediction.json
static coarse (SC)                   & $80.1$ & $ 9.5$ & $39.5$ & $ 0.0$ & $ 0.0$ & $ 0.0$ & $ 0.0$ & $ 0.0$ & $16.4$ & $ 0.0$ & $24.3$ & $ 0.0$ & $ 0.0$ & $26.2$ & $ 0.0$ & $ 0.0$ & $ 0.0$ & $ 0.0$ & $ 0.0$ & $10.3$ \NN % staticPredictionCoarseExclUnlabeled.json
GT segmentation with SF              & $80.8$ & $11.1$ & $44.5$ & $ 0.0$ & $ 0.0$ & $ 0.0$ & $ 0.0$ & $ 0.0$ & $ 4.2$ & $ 0.0$ & $17.9$ & $ 0.0$ & $ 0.0$ & $32.9$ & $ 0.0$ & $ 0.0$ & $ 0.0$ & $ 0.0$ & $ 0.0$ & $10.1$ \NN % staticPredictionPerfectSegmentation.json
GT segmentation with SC              & $79.6$ & $ 5.1$ & $46.6$ & $ 0.0$ & $ 0.0$ & $ 0.0$ & $ 0.0$ & $ 0.0$ & $11.8$ & $ 0.0$ & $29.2$ & $ 0.0$ & $ 0.0$ & $34.1$ & $ 0.0$ & $ 0.0$ & $ 0.0$ & $ 0.0$ & $ 0.0$ & $10.9$ \NN[3pt] % staticPredictionPerfectSegmentationCoarse.json
GT segmentation with \cite{Long2015} & $99.3$ & $91.9$ & $94.8$ & $44.9$ & $62.0$ & $66.1$ & $81.2$ & $84.3$ & $96.5$ & $80.1$ & $99.1$ & $90.6$ & $69.2$ & $98.0$ & $59.0$ & $66.9$ & $71.6$ & $66.8$ & $85.8$ & $79.4$ \NN[3pt] % fcnPerfectSegmentation.json
GT subsampled by $2$                 & $99.6$ & $98.1$ & $98.6$ & $97.8$ & $97.4$ & $90.4$ & $94.1$ & $95.2$ & $98.7$ & $97.6$ & $98.3$ & $96.5$ & $95.7$ & $98.9$ & $98.9$ & $99.1$ & $98.9$ & $96.5$ & $95.8$ & $97.2$ \NN % f02.json
GT subsampled by $4$                 & $99.4$ & $96.8$ & $98.0$ & $96.1$ & $95.5$ & $83.1$ & $89.7$ & $91.6$ & $98.0$ & $96.0$ & $97.9$ & $94.1$ & $92.5$ & $98.2$ & $98.1$ & $98.5$ & $98.1$ & $94.1$ & $93.0$ & $95.2$ \NN % f04.json
GT subsampled by $8$                 & $98.6$ & $93.4$ & $95.4$ & $92.3$ & $91.1$ & $69.5$ & $80.9$ & $84.2$ & $95.5$ & $92.1$ & $94.5$ & $88.9$ & $86.1$ & $96.2$ & $95.9$ & $96.7$ & $96.1$ & $88.7$ & $86.8$ & $90.7$ \NN % f08.json
GT subsampled by $16$                & $97.8$ & $88.8$ & $93.1$ & $86.9$ & $84.9$ & $50.9$ & $68.4$ & $73.0$ & $93.4$ & $86.5$ & $93.1$ & $81.0$ & $76.0$ & $93.5$ & $93.0$ & $94.4$ & $93.4$ & $80.8$ & $78.0$ & $84.6$ \NN % f16.json
GT subsampled by $32$                & $96.0$ & $80.9$ & $88.7$ & $77.6$ & $75.2$ & $30.9$ & $51.6$ & $56.8$ & $89.2$ & $77.3$ & $88.7$ & $69.4$ & $62.3$ & $88.0$ & $87.4$ & $89.8$ & $88.5$ & $68.6$ & $65.6$ & $75.4$ \NN % f32.json
GT subsampled by $64$                & $92.1$ & $69.6$ & $83.0$ & $65.5$ & $61.0$ & $14.8$ & $32.1$ & $37.6$ & $83.3$ & $65.2$ & $81.6$ & $55.1$ & $46.4$ & $78.8$ & $78.9$ & $82.4$ & $80.2$ & $54.2$ & $50.7$ & $63.8$ \NN % f64.json
GT subsampled by $128$               & $86.2$ & $55.0$ & $75.2$ & $51.3$ & $45.9$ & $ 5.7$ & $13.6$ & $17.9$ & $75.2$ & $51.6$ & $69.9$ & $41.1$ & $31.5$ & $67.3$ & $66.3$ & $70.1$ & $68.3$ & $36.0$ & $33.3$ & $50.6$ \NN[3pt] % f128.json
nearest training neighbor            & $85.3$ & $35.6$ & $56.7$ & $15.6$ & $ 6.2$ & $ 1.3$ & $ 0.5$ & $ 1.0$ & $54.2$ & $23.3$ & $36.5$ & $ 4.0$ & $ 0.4$ & $42.0$ & $ 9.7$ & $18.3$ & $12.9$ & $ 0.3$ & $ 1.7$ & $21.3$ \LL % sub128MajorityTrue.json
}

\ctable[
star,
caption = {Detailed results of our control experiments for the
  pixel-level semantic labeling task in terms of the
  instance-normalized $\iiou$ score on
  the class level.
  All numbers are given in percent.
  See the main paper for details on the listed methods.},
label = tab:instancelevelfromclasses_control,
pos = p,
% width=\linewidth,
% mincapwidth=0.8\textwidth,
doinside=\footnotesize
]{U VVVVVVVV W}{
}{
\FL
& \rotatedlabel{person}
& \rotatedlabel{rider}
& \rotatedlabel{car}
& \rotatedlabel{truck}
& \rotatedlabel{bus}
& \rotatedlabel{train}
& \rotatedlabel{motorcycle}
& \rotatedlabel{bicycle}
& \rotatedlabel{\textbf{mean $\iiou$}} \ML
static fine (SF)                     & $ 0.0$ & $ 0.0$ & $38.0$ & $ 0.0$ & $ 0.0$ & $ 0.0$ & $ 0.0$ & $ 0.0$ & $ 4.7$ \NN % staticPrediction.json
static coarse (SC)                   & $ 0.0$ & $ 0.0$ & $39.8$ & $ 0.0$ & $ 0.0$ & $ 0.0$ & $ 0.0$ & $ 0.0$ & $ 5.0$ \NN % staticPredictionCoarseExclUnlabeled.json
GT segmentation with SF              & $ 0.0$ & $ 0.0$ & $50.3$ & $ 0.0$ & $ 0.0$ & $ 0.0$ & $ 0.0$ & $ 0.0$ & $ 6.3$ \NN % staticPredictionPerfectSegmentation.json
GT segmentation with SC              & $ 0.0$ & $ 0.0$ & $50.8$ & $ 0.0$ & $ 0.0$ & $ 0.0$ & $ 0.0$ & $ 0.0$ & $ 6.3$ \NN[3pt] % staticPredictionPerfectSegmentationCoarse.json
GT segmentation with \cite{Long2015} & $68.3$ & $44.4$ & $92.8$ & $32.3$ & $38.7$ & $41.5$ & $39.5$ & $63.1$ & $52.6$ \NN[3pt] % fcnPerfectSegmentation.json
GT subsampled by $2$                 & $91.4$ & $91.9$ & $95.1$ & $93.3$ & $94.1$ & $94.3$ & $91.4$ & $89.6$ & $92.6$ \NN % f02.json
GT subsampled by $4$                 & $88.1$ & $86.4$ & $94.4$ & $91.8$ & $93.1$ & $93.0$ & $88.9$ & $87.2$ & $90.4$ \NN % f04.json
GT subsampled by $8$                 & $78.4$ & $75.6$ & $89.7$ & $85.7$ & $87.8$ & $88.8$ & $79.4$ & $76.8$ & $82.8$ \NN % f08.json
GT subsampled by $16$                & $63.5$ & $58.5$ & $82.6$ & $73.4$ & $78.2$ & $81.5$ & $66.4$ & $62.3$ & $70.8$ \NN % f16.json
GT subsampled by $32$                & $45.5$ & $38.0$ & $71.0$ & $57.7$ & $62.1$ & $66.0$ & $46.2$ & $43.5$ & $53.7$ \NN % f32.json
GT subsampled by $64$                & $28.4$ & $19.1$ & $51.0$ & $37.0$ & $42.0$ & $51.4$ & $27.6$ & $24.4$ & $35.1$ \NN % f64.json
GT subsampled by $128$               & $19.1$ & $10.5$ & $41.9$ & $18.9$ & $24.5$ & $30.7$ & $11.0$ & $11.8$ & $21.1$ \NN[3pt] % f128.json
nearest training neighbor            & $ 3.6$ & $ 0.5$ & $32.7$ & $ 1.9$ & $ 4.0$ & $ 2.8$ & $ 0.3$ & $ 1.5$ & $ 5.9$ \LL % sub128MajorityTrue.json
}

\clearpage

\ctable[
star,
caption = {Detailed results of our baseline experiments for the
  pixel-level semantic labeling task in terms of the $\iou$ score on
  the class level.
  All numbers are given in percent and we indicate the used training
  data for each method, \ie \emph{train} fine, \emph{val} fine, \emph{coarse}
  extra, as well as a potential downscaling factor (\emph{sub})
  of the input image.
  See the main paper and \cref{subsec:classlevel} for details on the listed methods.},
label = tab:pixellevel_methods,
pos = t,
doinside=\tiny
]{U VVVV WVVVVVVVVVVVVVVVVVV W}{
}{
\FL
& \rotatedlabel{train}
& \rotatedlabel{val}
& \rotatedlabel{coarse}
& \rotatedlabel{sub}
& \rotatedlabel{road}
& \rotatedlabel{sidewalk}
& \rotatedlabel{building}
& \rotatedlabel{wall}
& \rotatedlabel{fence}
& \rotatedlabel{pole}
& \rotatedlabel{traffic light}
& \rotatedlabel{traffic sign}
& \rotatedlabel{vegetation}
& \rotatedlabel{terrain}
& \rotatedlabel{sky}
& \rotatedlabel{person}
& \rotatedlabel{rider}
& \rotatedlabel{car}
& \rotatedlabel{truck}
& \rotatedlabel{bus}
& \rotatedlabel{train}
& \rotatedlabel{motorcycle}
& \rotatedlabel{bicycle}
& \rotatedlabel{\textbf{mean $\iou$}} \ML
FCN-32s                         & \yes & \yes &      &     &  $97.1$  &  $76.0$  &  $87.6$  &  $33.1$  &  $36.3$  &  $35.2$  &  $53.2$  &  $58.1$  &  $89.5$  &  $66.7$  &  $91.6$  &  $71.1$  &  $46.7$  &  $91.0$  &  $33.3$  &  $46.6$  &  $43.8$  &  $48.2$  &  $59.1$  &  $61.3$  \NN % fcnOurs_32sTrainval.json
FCN-16s                         & \yes & \yes &      &     &  $97.3$  &  $77.6$  &  $88.7$  &  $34.7$  &  $44.0$  &  $43.0$  &  $57.7$  &  $62.0$  &  $90.9$  &  $68.6$  &  $92.9$  &  $75.4$  &  $50.5$  &  $91.9$  &  $35.3$  &  $49.1$  &  $45.9$  &  $50.7$  &  $65.2$  &  $64.3$  \NN % fcnOurs_16sTrainval.json
FCN-8s                          & \yes & \yes &      &     &  $97.4$  &  $78.4$  &  $89.2$  &  $34.9$  &  $44.2$  &  $47.4$  &\bst{60.1}&  $65.0$  &  $91.4$  &  $69.3$  &\bst{93.9}&  $77.1$  &  $51.4$  &  $92.6$  &  $35.3$  &  $48.6$  &  $46.5$  &  $51.6$  &\bst{66.8}&  $65.3$  \NN % fcnOurs_8srlTrainval.json
FCN-8s                          & \yes & \yes &      & $2$ &  $97.0$  &  $75.4$  &  $87.3$  &  $37.4$  &  $39.0$  &  $35.1$  &  $47.7$  &  $53.3$  &  $89.3$  &  $66.1$  &  $92.5$  &  $69.5$  &  $46.0$  &  $90.8$  &  $41.9$  &  $52.9$  &  $50.1$  &  $46.5$  &  $58.4$  &  $61.9$  \NN % fcnOurs_8srlTrainValSub2.json
FCN-8s                          &      & \yes &      &     &  $95.9$  &  $69.7$  &  $86.9$  &  $23.1$  &  $32.6$  &  $44.3$  &  $52.1$  &  $56.8$  &  $90.2$  &  $60.9$  &  $92.9$  &  $73.3$  &  $42.7$  &  $89.9$  &  $22.8$  &  $39.2$  &  $29.6$  &  $42.5$  &  $63.1$  &  $58.3$  \NN % fcnOurs_8srlVal.json
FCN-8s                          &      &      & \yes &     &  $95.3$  &  $67.7$  &  $84.6$  &  $35.9$  &  $41.0$  &  $36.0$  &  $44.9$  &  $52.7$  &  $86.6$  &  $60.2$  &  $90.2$  &  $59.6$  &  $37.2$  &  $86.1$  &  $35.4$  &  $53.1$  &  $39.7$  &  $42.6$  &  $52.6$  &  $58.0$  \ML % fcnOursCoarse_8sTrainCoarseExtraCoarse.json
\cite{Badrinarayanan2015} ext.  & \yes &      &      & $4$ &  $95.6$  &  $70.1$  &  $82.8$  &  $29.9$  &  $31.9$  &  $38.1$  &  $43.1$  &  $44.6$  &  $87.3$  &  $62.3$  &  $91.7$  &  $67.3$  &  $50.7$  &  $87.9$  &  $21.7$  &  $29.0$  &  $34.7$  &  $40.5$  &  $56.6$  &  $56.1$  \NN % segnet_extended.json
\cite{Badrinarayanan2015} basic & \yes &      &      & $4$ &  $96.4$  &  $73.2$  &  $84.0$  &  $28.5$  &  $29.0$  &  $35.7$  &  $39.8$  &  $45.2$  &  $87.0$  &  $63.8$  &  $91.8$  &  $62.8$  &  $42.8$  &  $89.3$  &  $38.1$  &  $43.1$  &  $44.2$  &  $35.8$  &  $51.9$  &  $57.0$  \NN % segnet.json
\cite{Liu2015}                  & \yes & \yes & \yes & $3$ &  $96.3$  &  $71.7$  &  $86.7$  &  $43.7$  &  $31.7$  &  $29.2$  &  $35.8$  &  $47.4$  &  $88.4$  &  $63.1$  &\bst{93.9}&  $64.7$  &  $38.7$  &  $88.8$  &  $48.0$  &  $56.4$  &  $49.4$  &  $38.3$  &  $50.0$  &  $59.1$  \NN % dpn.json
\cite{Zheng2015}                & \yes &    &      & $2$ &  $96.3$  &  $73.9$  &  $88.2$  &\bst{47.6}&  $41.3$  &  $35.2$  &  $49.5$  &  $59.7$  &  $90.6$  &  $66.1$  &  $93.5$  &  $70.4$  &  $34.7$  &  $90.1$  &  $39.2$  &  $57.5$  &  $55.4$  &  $43.9$  &  $54.6$  &  $62.5$  \NN % crfasrnn.json
\cite{Chen2015}                 & \yes & \yes &      & $2$ &  $97.3$  &  $77.7$  &  $87.7$  &  $43.6$  &  $40.5$  &  $29.7$  &  $44.5$  &  $55.4$  &  $89.4$  &  $67.0$  &  $92.7$  &  $71.2$  &  $49.4$  &  $91.4$  &  $48.7$  &  $56.7$  &  $49.1$  &  $47.9$  &  $58.6$  &  $63.1$  \NN % DeepLab_LargeFOV_Strong.json
\cite{Papandreou2015}           & \yes & \yes & \yes & $2$ &  $97.4$  &  $78.3$  &  $88.1$  &  $47.5$  &  $44.2$  &  $29.5$  &  $44.4$  &  $55.4$  &  $89.4$  &  $67.3$  &  $92.8$  &  $71.0$  &  $49.3$  &  $91.4$  &\bst{55.9}&\bst{66.6}&\bst{56.7}&  $48.1$  &  $58.1$  &  $64.8$  \NN % DeepLab_LargeFOV_StrongWeak.json
\cite{Lin2015}                  & \yes &      &      &     &  $97.3$  &  $78.5$  &  $88.4$  &  $44.5$  &\bst{48.3}&  $34.1$  &  $55.5$  &  $61.7$  &  $90.1$  &\bst{69.5}&  $92.2$  &  $72.5$  &  $52.3$  &  $91.0$  &  $54.6$  &  $61.6$  &  $51.6$  &\bst{55.0}&  $63.1$  &  $66.4$  \NN % adelaide.json
\cite{Yu2016}                   & \yes &      &      &     &\bst{97.6}&\bst{79.2}&\bst{89.9}&  $37.3$  &  $47.6$  &\bst{53.2}&  $58.6$  &\bst{65.2}&\bst{91.8}&  $69.4$  &  $93.7$  &\bst{78.9}&\bst{55.0}&\bst{93.3}&  $45.5$  &  $53.4$  &  $47.7$  &  $52.2$  &  $66.0$  &\bst{67.1}\LL % dilation10.json
}

\ctable[
star,
caption = {Detailed results of our baseline experiments for the
  pixel-level semantic labeling task in terms of the
  instance-normalized $\iiou$ score on the class level.
  All numbers are given in percent and we indicate the used training
  data for each method, \ie \emph{train} fine, \emph{val} fine, \emph{coarse}
  extra, as well as a potential downscaling factor (\emph{sub})
  of the input image.
  See the main paper and \cref{subsec:classlevel} for details on the
  listed methods.},
label = tab:instancelevelfromclasses_methods,
pos = t,
% width=\linewidth,
% mincapwidth=0.8\textwidth,
doinside=\footnotesize
]{U VVVV WVVVVVVV W}{
}{
\FL
& \rotatedlabel{train}
& \rotatedlabel{val}
& \rotatedlabel{coarse}
& \rotatedlabel{sub}
& \rotatedlabel{person}
& \rotatedlabel{rider}
& \rotatedlabel{car}
& \rotatedlabel{truck}
& \rotatedlabel{bus}
& \rotatedlabel{train}
& \rotatedlabel{motorcycle}
& \rotatedlabel{bicycle}
& \rotatedlabel{\textbf{mean $\iiou$}} \ML
FCN-32s                            & \yes & \yes &      &     &  $46.9$  &  $32.0$  &  $82.1$  &  $21.2$  &  $28.8$  &  $21.9$  &  $26.0$  &  $47.1$  &  $38.2$  \NN % fcnOurs_32sTrainval.json
FCN-16s                            & \yes & \yes &      &     &  $53.6$  &  $33.5$  &  $84.2$  &  $21.3$  &  $32.8$  &  $25.8$  &  $28.9$  &  $48.6$  &  $41.1$  \NN % fcnOurs_16sTrainval.json
FCN-8s                             & \yes & \yes &      &     &  $55.9$  &  $33.4$  &  $83.9$  &  $22.2$  &  $30.8$  &  $26.7$  &  $31.1$  &  $49.6$  &  $41.7$  \NN % fcnOurs_8srlTrainval.json
FCN-8s                             & \yes & \yes &      & $2$ &  $42.8$  &  $22.3$  &  $79.3$  &  $16.6$  &  $27.3$  &  $22.2$  &  $20.0$  &  $38.5$  &  $33.6$  \NN % fcnOurs_8srlTrainValSub2.json
FCN-8s                             &      & \yes &      &     &  $51.8$  &  $31.0$  &  $80.6$  &  $17.0$  &  $23.9$  &  $24.5$  &  $23.7$  &  $47.3$  &  $37.4$  \NN % fcnOurs_8srlVal.json
FCN-8s                             &      &      & \yes &     &  $43.2$  &  $18.9$  &  $72.5$  &  $18.2$  &  $24.2$  &  $20.1$  &  $20.9$  &  $36.2$  &  $31.8$  \ML % fcnOursCoarse_8sTrainCoarseExtraCoarse.json
\cite{Badrinarayanan2015} extended & \yes &      &      & $4$ &  $49.9$  &  $27.1$  &  $81.1$  &  $15.3$  &  $23.7$  &  $18.5$  &  $19.6$  &  $38.4$  &  $34.2$  \NN % segnet_extended.json
\cite{Badrinarayanan2015} basic    & \yes &      &      & $4$ &  $44.3$  &  $22.7$  &  $78.4$  &  $16.1$  &  $24.3$  &  $20.7$  &  $15.8$  &  $33.6$  &  $32.0$  \NN % segnet.json
\cite{Liu2015}                     & \yes & \yes & \yes & $3$ &  $38.9$  &  $12.8$  &  $78.6$  &  $13.4$  &  $24.0$  &  $19.2$  &  $10.7$  &  $27.2$  &  $28.1$  \NN % dpn.json
\cite{Zheng2015}                   & \yes &      &      & $2$ &  $50.6$  &  $17.8$  &  $81.1$  &  $18.0$  &  $25.0$  &  $30.3$  &  $22.3$  &  $30.1$  &  $34.4$  \NN % crfasrnn.json
\cite{Chen2015}                    & \yes & \yes &      & $2$ &  $40.5$  &  $23.3$  &  $78.8$  &  $20.3$  &  $31.9$  &  $24.8$  &  $21.1$  &  $35.2$  &  $34.5$  \NN % DeepLab_LargeFOV_Strong.json
\cite{Papandreou2015}              & \yes & \yes & \yes & $2$ &  $40.7$  &  $23.1$  &  $78.6$  &  $21.4$  &  $32.4$  &  $27.6$  &  $20.8$  &  $34.6$  &  $34.9$  \NN % DeepLab_LargeFOV_StrongWeak.json
\cite{Lin2015}                     & \yes &      &      &     &  $56.2$  &\bst{38.0}&  $77.1$  &\bst{34.0}&\bst{47.0}&\bst{33.4}&\bst{38.1}&\bst{49.9}&\bst{46.7}\NN % adelaide.json
\cite{Yu2016}                      & \yes &      &      &     &\bst{56.3}&  $34.5$  &\bst{85.8}&  $21.8$  &  $32.7$  &  $27.6$  &  $28.0$  &  $49.1$  &  $42.0$  \LL % dilation10.json
}

\ctable[
star,
caption = {Detailed results of our baseline experiments for the
  instance-level semantic labeling task in terms of the
  region-level average precision scores $\mapr$ on the class level.
  All numbers are given in percent.
  See the main paper and \cref{subsec:instancelevel} for details on the
  listed methods.},
label = tab:instancelevel_baselines_ap,
pos = p,
doinside=\footnotesize
]{UU WVVVVVVV W}{
}{
\FL
   Proposals
 & Classifier
 & \rotatedlabel{person}
 & \rotatedlabel{rider}
 & \rotatedlabel{car}
 & \rotatedlabel{truck}
 & \rotatedlabel{bus}
 & \rotatedlabel{train}
 & \rotatedlabel{motorcycle}
 & \rotatedlabel{bicycle}
 & \rotatedlabel{\textbf{mean $\mapr$}} \ML
 MCG regions & FRCN &\bst{ 1.9}&\bst{ 1.0}&  $ 6.2$  &  $ 4.0$  &  $ 3.1$  &  $ 2.8$  &  $ 1.5$  &\bst{ 0.6}&  $ 2.6$  \NN % 2016-02-29_14-29-52_cityscapes_CityscapesVGG_lr1e-3step90000_output_iter_120000_excl_rcrr_mask_proposals.json
 MCG bboxes  & FRCN &  $ 0.5$  &  $ 0.1$  &  $ 7.8$  &\bst{ 6.4}&\bst{10.3}&  $ 4.5$  &  $ 0.9$  &  $ 0.2$  &  $ 3.8$  \NN % 2016-02-29_14-29-52_cityscapes_CityscapesVGG_lr1e-3step90000_output_iter_120000_excl_rcrr_box_proposals.json
 MCG hulls   & FRCN &  $ 1.3$  &  $ 0.6$  &\bst{10.5}&  $ 6.1$  &  $ 9.7$  &\bst{ 5.9}&\bst{ 1.7}&  $ 0.5$  &\bst{ 4.6}\NN[3pt] % 2016-02-29_14-29-52_cityscapes_CityscapesVGG_lr1e-3step90000_output_iter_120000_excl_rcrr_mask_proposals_convexHull.json
 GT bboxes   & FRCN &  $ 7.6$  &  $ 0.5$  &  $17.5$  &  $10.7$  &  $15.7$  &  $ 8.4$  &  $ 2.6$  &  $ 2.9$  &  $ 8.2$  \NN % 2016-02-29_14-29-52_cityscapes_CityscapesVGG_lr1e-3step90000_output_iter_120000_excl_gt_crr_box_proposals.json
 GT regions  & FRCN &  $65.5$  &  $40.6$  &  $65.9$  &  $21.1$  &  $31.9$  &  $30.2$  &  $28.8$  &  $46.4$  &  $41.3$  \NN[3pt] % 2016-02-29_14-29-52_cityscapes_CityscapesVGG_lr1e-3step90000_output_iter_120000_excl_gt_crr_mask_proposals.json
 MCG regions & GT   &  $ 3.7$  &  $ 4.4$  &  $11.9$  &  $19.9$  &  $21.5$  &  $12.4$  &  $ 7.8$  &  $ 2.6$  &  $10.5$  \NN % resultSemanticLabelingGtOracleAllProposals.json
 MCG bboxes  & GT   &  $ 2.0$  &  $ 2.0$  &  $10.9$  &  $18.2$  &  $22.1$  &  $15.9$  &  $ 6.0$  &  $ 2.2$  &  $ 9.9$  \NN % resultSemanticLabelingGtOracleAllBoundingBoxes.json
 MCG hulls   & GT   &  $ 3.4$  &  $ 4.1$  &  $13.4$  &  $20.4$  &  $24.1$  &  $16.0$  &  $ 8.3$  &  $ 2.8$  &  $11.6$  \LL % resultSemanticLabelingGtOracleAllConvexHull.json
}

\clearpage

\ctable[
star,
caption = {Detailed results of our baseline experiments for the
  instance-level semantic labeling task in terms of the
  region-level average precision scores $\mapr^{50\%}$ for an overlap value of \SI{50}{\percent}.
  All numbers are given in percent.
  See the main paper and \cref{subsec:instancelevel} for details on the
  listed methods.},
label = tab:instancelevel_baselines_ap50,
pos = p,
doinside=\footnotesize
]{UU WVVVVVVV W}{
}{
\FL
   Proposals
 & Classifier
 & \rotatedlabel{person}
 & \rotatedlabel{rider}
 & \rotatedlabel{car}
 & \rotatedlabel{truck}
 & \rotatedlabel{bus}
 & \rotatedlabel{train}
 & \rotatedlabel{motorcycle}
 & \rotatedlabel{bicycle}
 & \rotatedlabel{\textbf{mean $\mapr^{50\%}$}} \ML
 MCG regions & FRCN &\bst{ 6.7}&\bst{ 5.4}&  $19.3$  &  $10.3$  &  $11.9$  &  $ 7.6$  &  $ 7.8$  &  $ 3.0$  &  $ 9.0$  \NN % 2016-02-29_14-29-52_cityscapes_CityscapesVGG_lr1e-3step90000_output_iter_120000_excl_rcrr_mask_proposals.json
 MCG bboxes  & FRCN &  $ 2.7$  &  $ 0.6$  &  $23.3$  &\bst{15.4}&\bst{27.2}&  $15.2$  &  $ 4.8$  &  $ 1.4$  &  $11.3$  \NN % 2016-02-29_14-29-52_cityscapes_CityscapesVGG_lr1e-3step90000_output_iter_120000_excl_rcrr_box_proposals.json
 MCG hulls   & FRCN &  $ 5.6$  &  $ 3.9$  &\bst{26.0}&  $13.8$  &  $26.3$  &\bst{15.8}&\bst{ 8.6}&\bst{ 3.1}&\bst{12.9}\NN[3pt] % 2016-02-29_14-29-52_cityscapes_CityscapesVGG_lr1e-3step90000_output_iter_120000_excl_rcrr_mask_proposals_convexHull.json
 GT bboxes   & FRCN &  $35.4$  &  $ 4.3$  &  $44.9$  &  $19.3$  &  $29.9$  &  $26.7$  &  $11.9$  &  $16.7$  &  $23.7$  \NN % 2016-02-29_14-29-52_cityscapes_CityscapesVGG_lr1e-3step90000_output_iter_120000_excl_gt_crr_box_proposals.json
 GT regions  & FRCN &  $65.5$  &  $40.6$  &  $65.9$  &  $21.1$  &  $31.9$  &  $30.2$  &  $28.8$  &  $46.4$  &  $41.3$  \NN[3pt] % 2016-02-29_14-29-52_cityscapes_CityscapesVGG_lr1e-3step90000_output_iter_120000_excl_gt_crr_mask_proposals.json
 MCG regions & GT   &  $12.3$  &  $18.1$  &  $29.6$  &  $43.9$  &  $44.6$  &  $31.4$  &  $25.9$  &  $10.0$  &  $27.0$  \NN % resultSemanticLabelingGtOracleAllProposals.json
 MCG bboxes  & GT   &  $ 9.2$  &  $11.5$  &  $29.0$  &  $41.8$  &  $46.0$  &  $36.0$  &  $23.3$  &  $ 9.6$  &  $25.8$  \NN % resultSemanticLabelingGtOracleAllBoundingBoxes.json
 MCG hulls   & GT   &  $12.0$  &  $18.4$  &  $31.4$  &  $46.1$  &  $46.3$  &  $40.7$  &  $27.7$  &  $10.7$  &  $29.1$  \LL % resultSemanticLabelingGtOracleAllConvexHull.json
}

\ctable[
star,
caption = {Detailed results of our baseline experiments for the
  instance-level semantic labeling task in terms of the
  region-level average precision scores $\mapr^{100\text{m}}$ for objects within \SI{100}{\metre}.
  All numbers are given in percent.
  See the main paper and \cref{subsec:instancelevel} for details on the
  listed methods.},
label = tab:instancelevel_baselines_ap100m,
pos = p,
doinside=\footnotesize
]{UU WVVVVVVV W}{
}{
\FL
   Proposals
 & Classifier
 & \rotatedlabel{person}
 & \rotatedlabel{rider}
 & \rotatedlabel{car}
 & \rotatedlabel{truck}
 & \rotatedlabel{bus}
 & \rotatedlabel{train}
 & \rotatedlabel{motorcycle}
 & \rotatedlabel{bicycle}
 & \rotatedlabel{\textbf{mean $\mapr^{100\text{m}}$}} \ML
 MCG regions & FRCN &\bst{ 3.7}&\bst{ 1.6}&  $10.2$  &  $ 6.8$  &  $ 5.4$  &  $ 4.2$  &  $ 2.2$  &\bst{ 1.1}&  $ 4.4$  \NN % 2016-02-29_14-29-52_cityscapes_CityscapesVGG_lr1e-3step90000_output_iter_120000_excl_rcrr_mask_proposals.json
 MCG bboxes  & FRCN &  $ 0.9$  &  $ 0.1$  &  $12.9$  &\bst{11.3}&\bst{18.5}&  $ 6.9$  &  $ 1.3$  &  $ 0.3$  &  $ 6.5$  \NN % 2016-02-29_14-29-52_cityscapes_CityscapesVGG_lr1e-3step90000_output_iter_120000_excl_rcrr_box_proposals.json
 MCG hulls   & FRCN &  $ 2.6$  &  $ 1.1$  &\bst{17.5}&  $10.6$  &  $17.4$  &\bst{ 9.2}&\bst{ 2.6}&  $ 0.9$  &\bst{ 7.7}\NN[3pt] % 2016-02-29_14-29-52_cityscapes_CityscapesVGG_lr1e-3step90000_output_iter_120000_excl_rcrr_mask_proposals_convexHull.json
 GT bboxes   & FRCN &  $ 8.8$  &  $ 0.8$  &  $25.3$  &  $18.4$  &  $27.1$  &  $13.0$  &  $ 3.9$  &  $ 3.6$  &  $12.6$  \NN % 2016-02-29_14-29-52_cityscapes_CityscapesVGG_lr1e-3step90000_output_iter_120000_excl_gt_crr_box_proposals.json
 GT regions  & FRCN &  $79.1$  &  $66.0$  &  $78.9$  &  $33.6$  &  $53.9$  &  $47.1$  &  $42.6$  &  $63.5$  &  $58.1$  \NN[3pt] % 2016-02-29_14-29-52_cityscapes_CityscapesVGG_lr1e-3step90000_output_iter_120000_excl_gt_crr_mask_proposals.json
 MCG regions & GT   &  $ 6.8$  &  $ 6.8$  &  $18.9$  &  $28.7$  &  $32.7$  &  $19.0$  &  $10.5$  &  $ 4.3$  &  $16.0$  \NN % resultSemanticLabelingGtOracleAllProposals.json
 MCG bboxes  & GT   &  $ 3.5$  &  $ 2.9$  &  $17.3$  &  $27.3$  &  $34.5$  &  $24.9$  &  $ 8.2$  &  $ 3.7$  &  $15.3$  \NN % resultSemanticLabelingGtOracleAllBoundingBoxes.json
 MCG hulls   & GT   &  $ 6.1$  &  $ 6.2$  &  $21.4$  &  $29.9$  &  $37.2$  &  $24.7$  &  $11.4$  &  $ 4.7$  &  $17.7$  \LL % resultSemanticLabelingGtOracleAllConvexHull.json
}

\ctable[
star,
caption = {Detailed results of our baseline experiments for the
  instance-level semantic labeling task in terms of the
  region-level average precision scores $\mapr^{50\text{m}}$ for objects within \SI{50}{\metre}.
  All numbers are given in percent.
  See the main paper and \cref{subsec:instancelevel} for details on the
  listed methods.},
label = tab:instancelevel_baselines_ap50m,
pos = p,
doinside=\footnotesize
]{UU WVVVVVVV W}{
}{
\FL
   Proposals
 & Classifier
 & \rotatedlabel{person}
 & \rotatedlabel{rider}
 & \rotatedlabel{car}
 & \rotatedlabel{truck}
 & \rotatedlabel{bus}
 & \rotatedlabel{train}
 & \rotatedlabel{motorcycle}
 & \rotatedlabel{bicycle}
 & \rotatedlabel{\textbf{mean $\mapr^{50\text{m}}$}} \ML
 MCG regions & FRCN &\bst{ 4.0}&\bst{ 1.7}&  $12.0$  &  $ 9.0$  &  $ 7.8$  &  $ 6.4$  &  $ 2.4$  &\bst{ 1.1}&  $ 5.5$  \NN % 2016-02-29_14-29-52_cityscapes_CityscapesVGG_lr1e-3step90000_output_iter_120000_excl_rcrr_mask_proposals.json
 MCG bboxes  & FRCN &  $ 1.0$  &  $ 0.1$  &  $15.5$  &\bst{14.9}&\bst{27.7}&  $10.0$  &  $ 1.4$  &  $ 0.4$  &  $ 8.9$  \NN % 2016-02-29_14-29-52_cityscapes_CityscapesVGG_lr1e-3step90000_output_iter_120000_excl_rcrr_box_proposals.json
 MCG hulls   & FRCN &  $ 2.7$  &  $ 1.1$  &\bst{21.2}&  $14.0$  &  $25.2$  &\bst{14.2}&\bst{ 2.7}&  $ 1.0$  &\bst{10.3}\NN[3pt] % 2016-02-29_14-29-52_cityscapes_CityscapesVGG_lr1e-3step90000_output_iter_120000_excl_rcrr_mask_proposals_convexHull.json
 GT bboxes   & FRCN &  $ 8.5$  &  $ 0.8$  &  $26.6$  &  $23.2$  &  $37.2$  &  $17.7$  &  $ 4.1$  &  $ 3.6$  &  $15.2$  \NN % 2016-02-29_14-29-52_cityscapes_CityscapesVGG_lr1e-3step90000_output_iter_120000_excl_gt_crr_box_proposals.json
 GT regions  & FRCN &  $79.1$  &  $68.3$  &  $80.5$  &  $42.9$  &  $69.4$  &  $67.9$  &  $46.2$  &  $64.7$  &  $64.9$  \NN[3pt] % 2016-02-29_14-29-52_cityscapes_CityscapesVGG_lr1e-3step90000_output_iter_120000_excl_gt_crr_mask_proposals.json
 MCG regions & GT   &  $ 7.2$  &  $ 7.0$  &  $21.7$  &  $32.4$  &  $42.4$  &  $23.6$  &  $11.1$  &  $ 4.5$  &  $18.7$  \NN % resultSemanticLabelingGtOracleAllProposals.json
 MCG bboxes  & GT   &  $ 3.7$  &  $ 3.0$  &  $19.9$  &  $33.0$  &  $46.0$  &  $32.9$  &  $ 8.6$  &  $ 3.8$  &  $18.9$  \NN % resultSemanticLabelingGtOracleAllBoundingBoxes.json
 MCG hulls   & GT   &  $ 6.5$  &  $ 6.4$  &  $24.8$  &  $35.4$  &  $49.6$  &  $31.8$  &  $12.2$  &  $ 4.9$  &  $21.4$  \LL % resultSemanticLabelingGtOracleAllConvexHull.json
}

\clearpage

%%%%%%%%%%%%%%%%%%%%%%%%%
% Largest number of instances and persons
%%%%%%%%%%%%%%%%%%%%%%%%%

\begin{figure*}[p]
    \captionsetup[subfigure]{aboveskip=2pt,belowskip=2pt}
    \centering%
    \begin{subfigure}[b]{0.499\linewidth}
        \begin{overpic}[width=\textwidth]{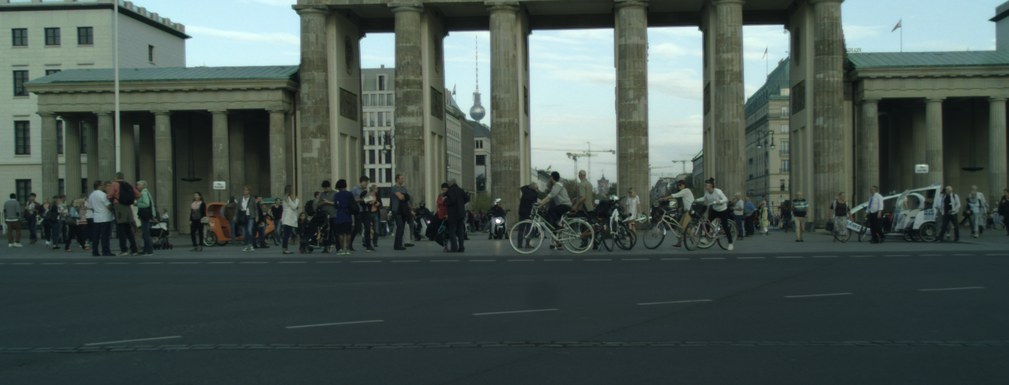}
        \put (50,4) {\makebox(0,0){\textcolor{white}{\textsf{\footnotesize Image}}}}
        \end{overpic}
    \end{subfigure}\hfill
    \begin{subfigure}[b]{0.499\linewidth}
        \begin{overpic}[width=\textwidth]{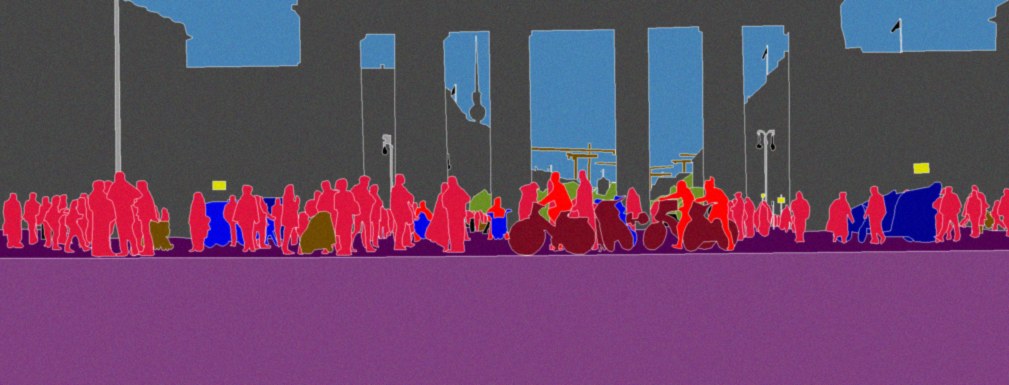}
        \put (50,4) {\makebox(0,0){\textcolor{white}{\textsf{\footnotesize Annotation}}}}
        \end{overpic}
    \end{subfigure}\\
    \begin{subfigure}[b]{0.499\linewidth}
        \begin{overpic}[width=\textwidth]{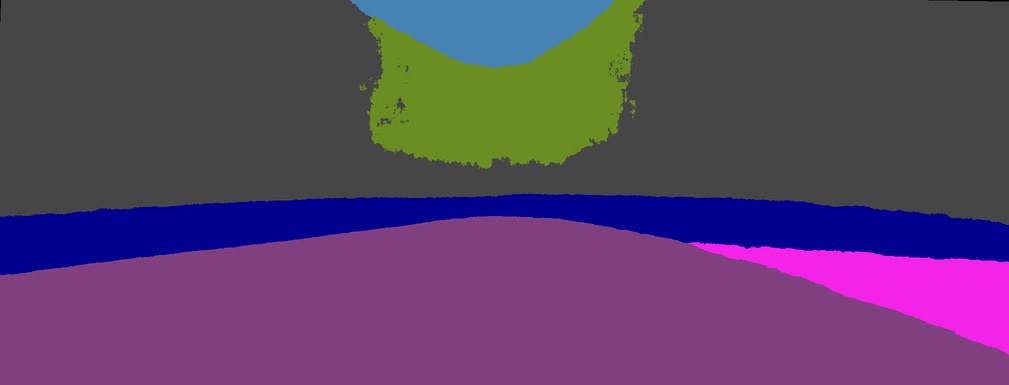}
        \put (50,4) {\makebox(0,0){\textcolor{white}{\textsf{\footnotesize static fine (SF)}}}}
        \end{overpic}
    \end{subfigure}\hfill
    \begin{subfigure}[b]{0.499\linewidth}
        \begin{overpic}[width=\textwidth]{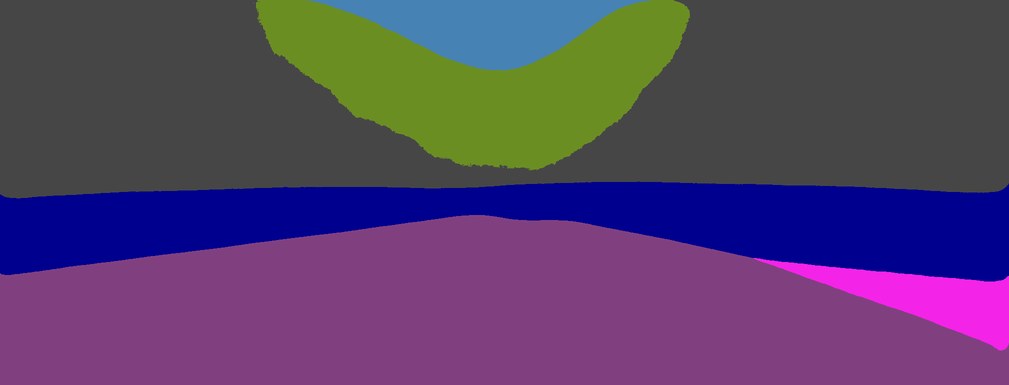}
        \put (50,4) {\makebox(0,0){\textcolor{white}{\textsf{\footnotesize static coarse (SC)}}}}
        \end{overpic}
    \end{subfigure}\\
    \begin{subfigure}[b]{0.499\linewidth}
        \begin{overpic}[width=\textwidth]{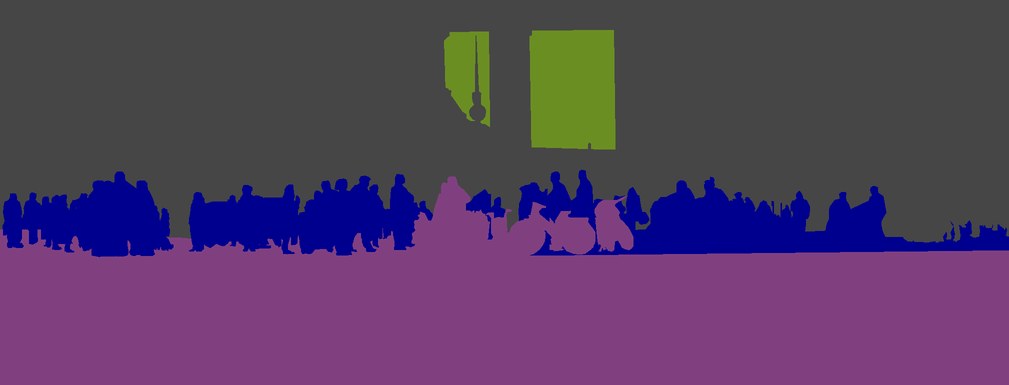}
        \put (50,4) {\makebox(0,0){\textcolor{white}{\textsf{\footnotesize GT segmentation w/ SF}}}}
        \end{overpic}
    \end{subfigure}\hfill
    \begin{subfigure}[b]{0.499\linewidth}
        \begin{overpic}[width=\textwidth]{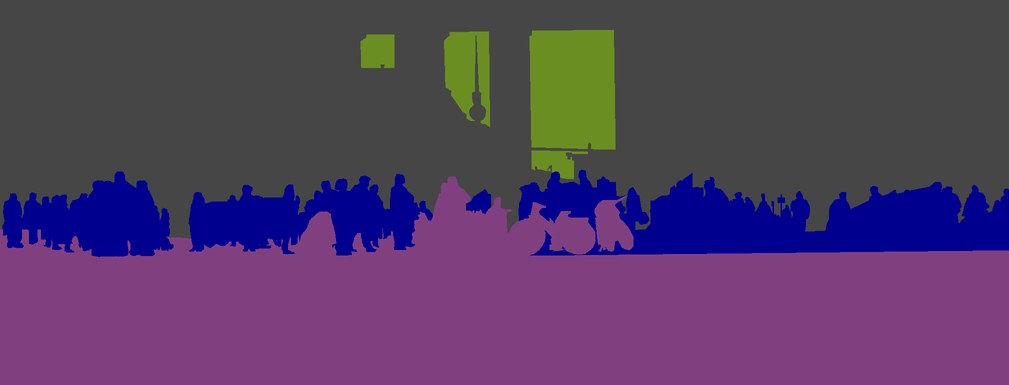}
        \put (50,4) {\makebox(0,0){\textcolor{white}{\textsf{\footnotesize GT segmentation w/ SC}}}}
        \end{overpic}
    \end{subfigure}\\
    \begin{subfigure}[b]{0.499\linewidth}
        \begin{overpic}[width=\textwidth]{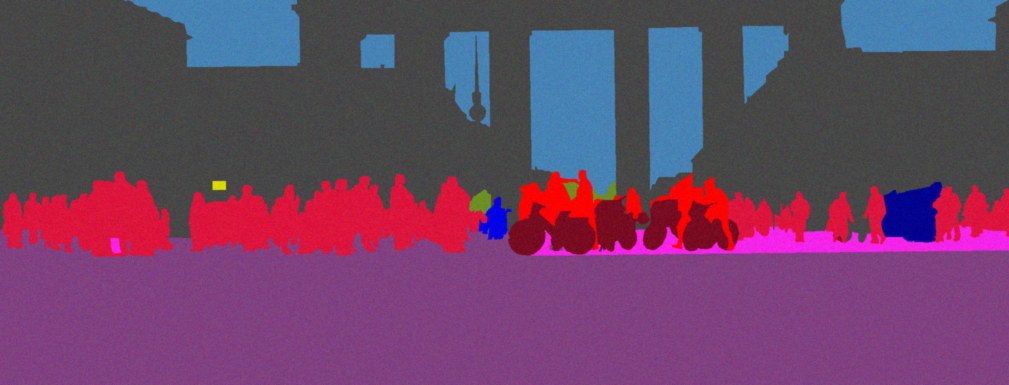}
        \put (50,4) {\makebox(0,0){\textcolor{white}{\textsf{\footnotesize GT segmentation w/ \cite{Long2015}}}}}
        \end{overpic}
    \end{subfigure}\hfill
    \begin{subfigure}[b]{0.499\linewidth}
        \begin{overpic}[width=\textwidth]{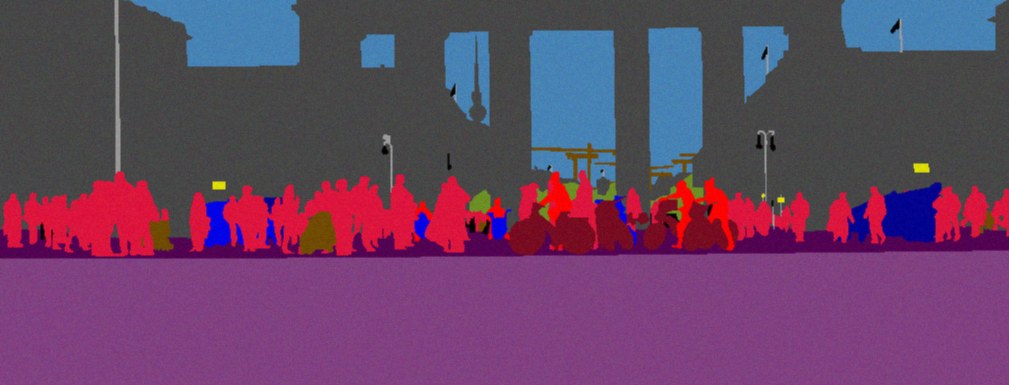}
        \put (50,4) {\makebox(0,0){\textcolor{white}{\textsf{\footnotesize GT subsampled by 2}}}}
        \end{overpic}
    \end{subfigure}\\
    \begin{subfigure}[b]{0.499\linewidth}
        \begin{overpic}[width=\textwidth]{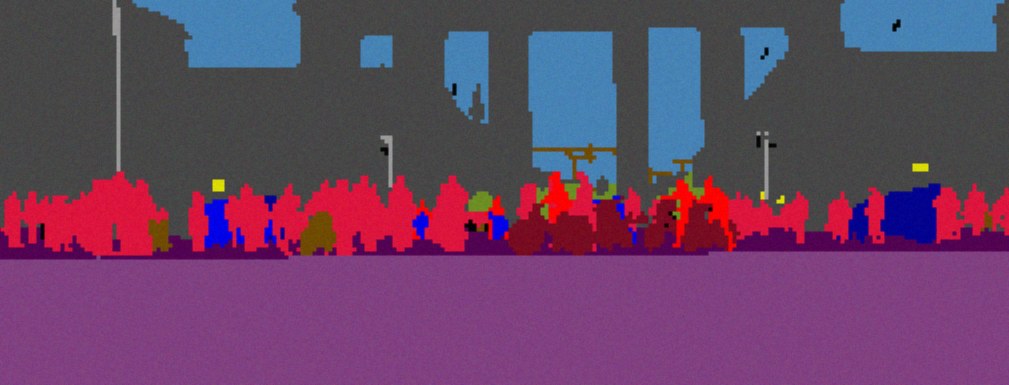}
        \put (50,4) {\makebox(0,0){\textcolor{white}{\textsf{\footnotesize GT subsampled by 8}}}}
        \end{overpic}
    \end{subfigure}\hfill
    \begin{subfigure}[b]{0.499\linewidth}
        \begin{overpic}[width=\textwidth]{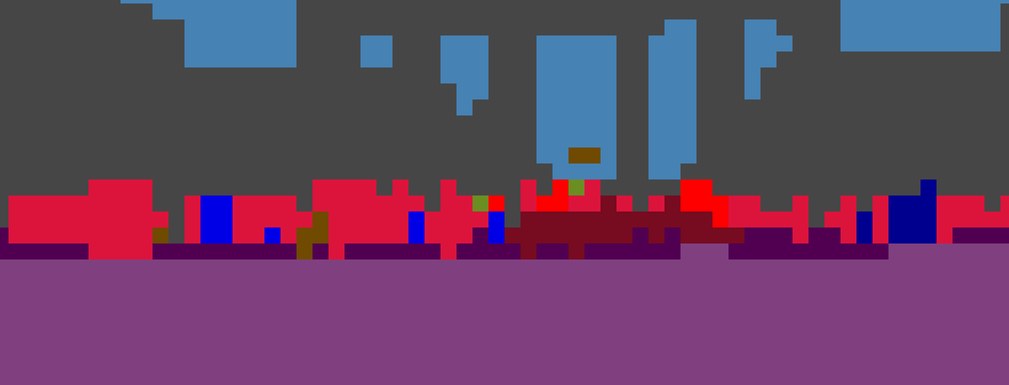}
        \put (50,4) {\makebox(0,0){\textcolor{white}{\textsf{\footnotesize GT subsampled by 32}}}}
        \end{overpic}
    \end{subfigure}\\
    \begin{subfigure}[b]{0.499\linewidth}
        \begin{overpic}[width=\textwidth]{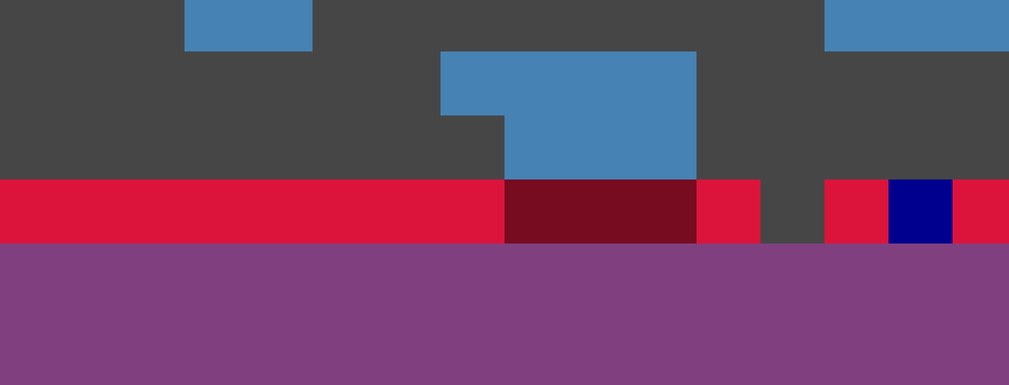}
        \put (50,4) {\makebox(0,0){\textcolor{white}{\textsf{\footnotesize GT subsampled by 128}}}}
        \end{overpic}
    \end{subfigure}\hfill
    \begin{subfigure}[b]{0.499\linewidth}
        \begin{overpic}[width=\textwidth]{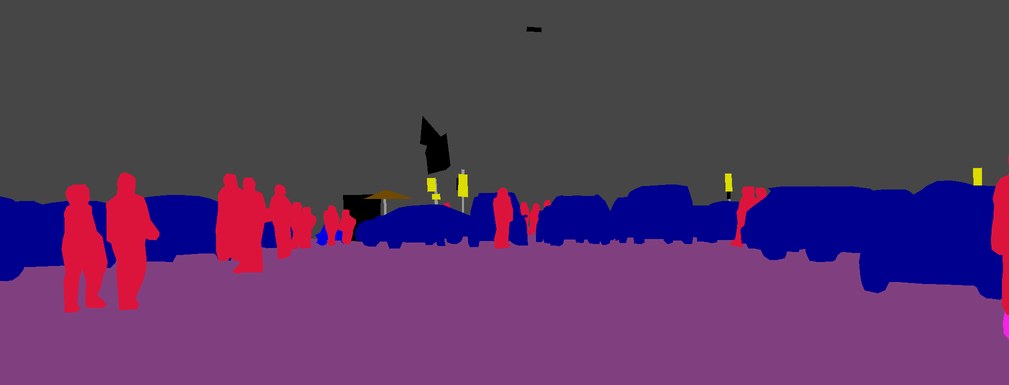}
        \put (50,4) {\makebox(0,0){\textcolor{white}{\textsf{\footnotesize nearest training neighbor}}}}
        \end{overpic}
    \end{subfigure}\\
    \vspace{2mm}
    \caption{Exemplary output of our control experiments for the pixel-level semantic labeling task, see the main paper for details. The image is part of our \textit{test} set and has both, the largest number of instances and persons.}
    \label{fig:mostallcontrol}
\end{figure*}

\begin{figure*}[p]
    \captionsetup[subfigure]{aboveskip=2pt,belowskip=2pt}
    \centering%
    \begin{subfigure}[b]{0.499\linewidth}
        \begin{overpic}[width=\textwidth]{figures/examplesAutoInclBaselineSmallRes/berlin00057_000019_most_all_img}
        \put (50,4) {\makebox(0,0){\textcolor{white}{\textsf{\footnotesize Image}}}}
        \end{overpic}
    \end{subfigure}\hfill
    \begin{subfigure}[b]{0.499\linewidth}
        \begin{overpic}[width=\textwidth]{figures/examplesAutoInclBaselineSmallRes/berlin00057_000019_most_all_annotation}
        \put (50,4) {\makebox(0,0){\textcolor{white}{\textsf{\footnotesize Annotation}}}}
        \end{overpic}
    \end{subfigure}\\
    \begin{subfigure}[b]{0.499\linewidth}
        \begin{overpic}[width=\textwidth]{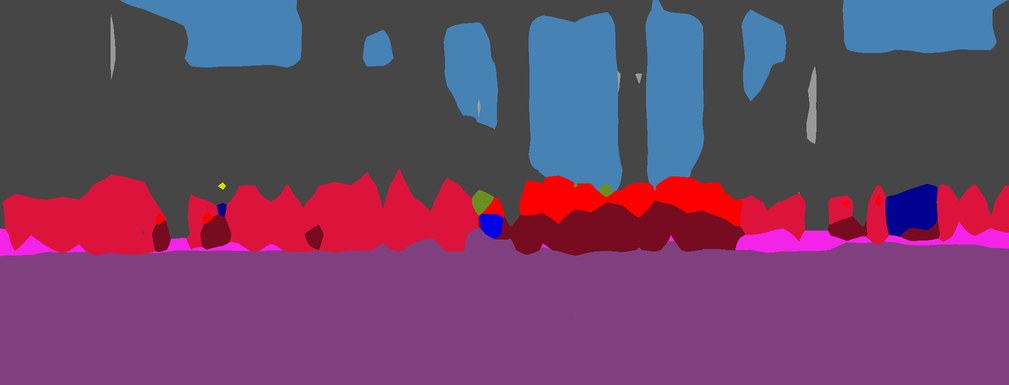}
        \put (50,4) {\makebox(0,0){\textcolor{white}{\textsf{\footnotesize FCN-32s}}}}
        \end{overpic}
    \end{subfigure}\hfill
    \begin{subfigure}[b]{0.499\linewidth}
        \begin{overpic}[width=\textwidth]{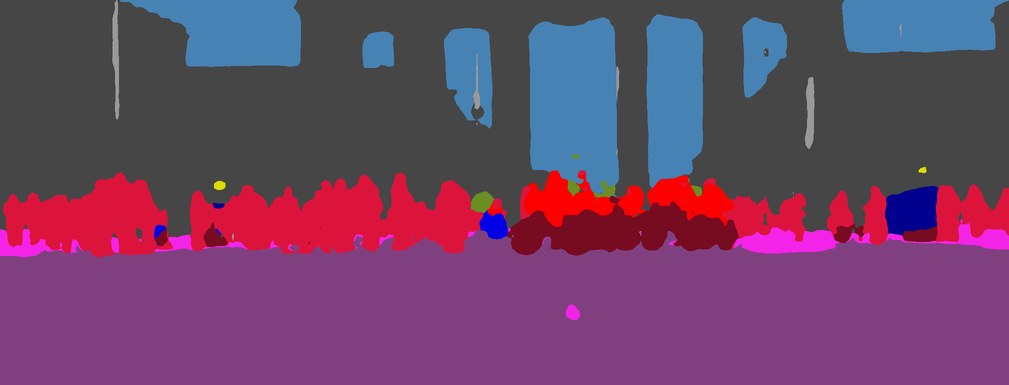}
        \put (50,4) {\makebox(0,0){\textcolor{white}{\textsf{\footnotesize FCN-8s}}}}
        \end{overpic}
    \end{subfigure}\\
    \begin{subfigure}[b]{0.499\linewidth}
        \begin{overpic}[width=\textwidth]{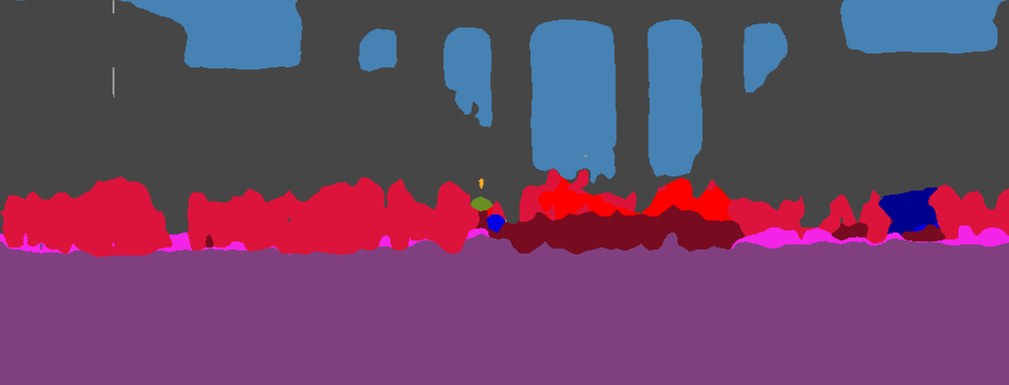}
        \put (50,4) {\makebox(0,0){\textcolor{white}{\textsf{\footnotesize FCN-8s half resolution}}}}
        \end{overpic}
    \end{subfigure}\hfill
    \begin{subfigure}[b]{0.499\linewidth}
        \begin{overpic}[width=\textwidth]{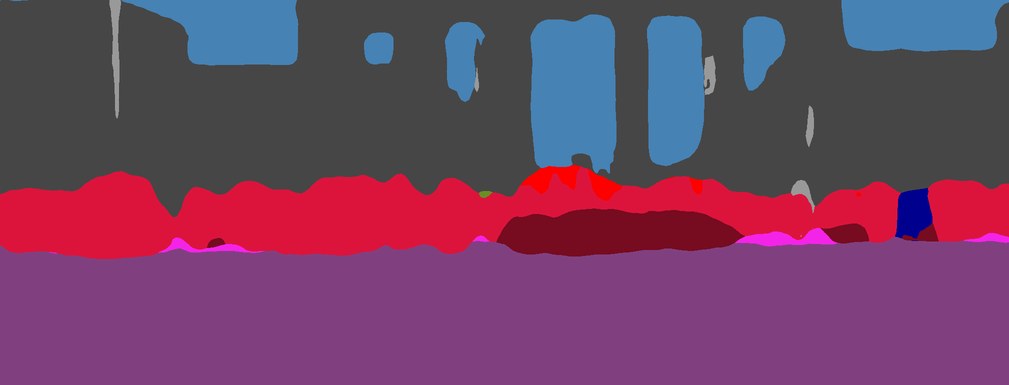}
        \put (50,4) {\makebox(0,0){\textcolor{white}{\textsf{\footnotesize FCN-8s trained on coarse}}}}
        \end{overpic}
    \end{subfigure}\\
    \begin{subfigure}[b]{0.499\linewidth}
        \begin{overpic}[width=\textwidth]{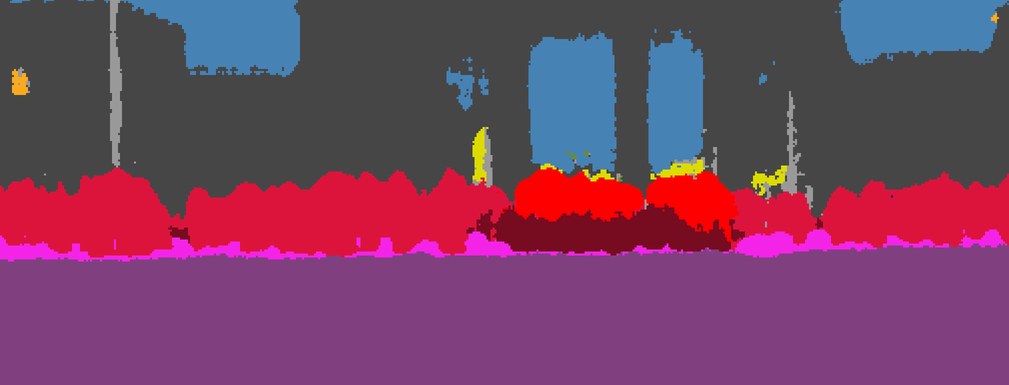}
        \put (50,4) {\makebox(0,0){\textcolor{white}{\textsf{\footnotesize SegNet basic \cite{Badrinarayanan2015}}}}}
        \end{overpic}
    \end{subfigure}\hfill
    \begin{subfigure}[b]{0.499\linewidth}
        \begin{overpic}[width=\textwidth]{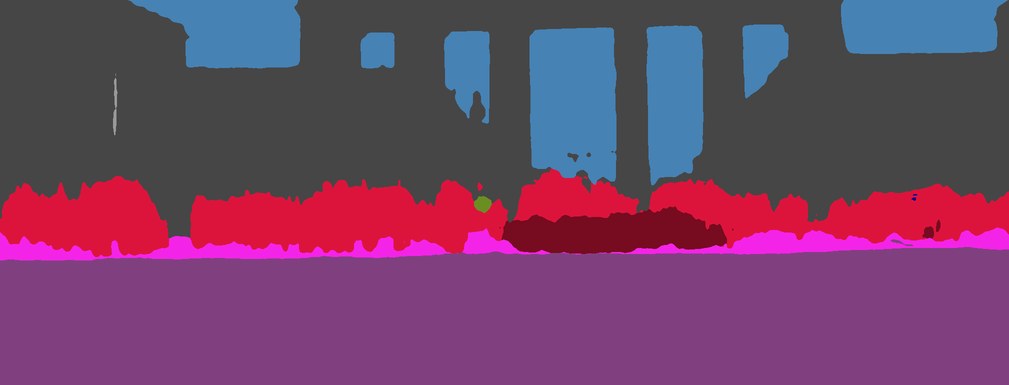}
        \put (50,4) {\makebox(0,0){\textcolor{white}{\textsf{\footnotesize DPN \cite{Liu2015}}}}}
        \end{overpic}
    \end{subfigure}\\
    \begin{subfigure}[b]{0.499\linewidth}
        \begin{overpic}[width=\textwidth]{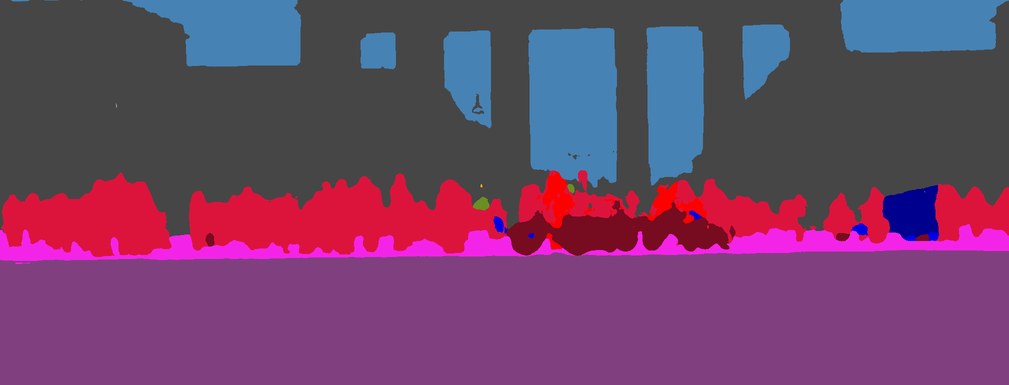}
        \put (50,4) {\makebox(0,0){\textcolor{white}{\textsf{\footnotesize CRF as RNN \cite{Zheng2015}}}}}
        \end{overpic}
    \end{subfigure}\hfill
    \begin{subfigure}[b]{0.499\linewidth}
        \begin{overpic}[width=\textwidth]{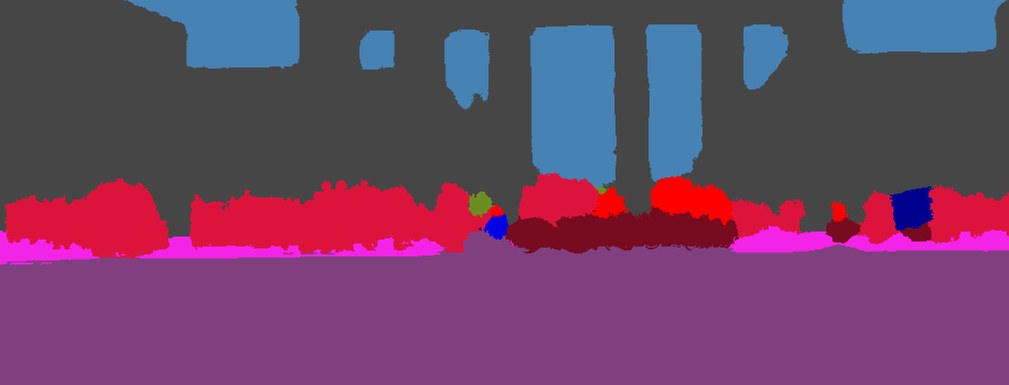}
        \put (50,4) {\makebox(0,0){\textcolor{white}{\textsf{\footnotesize DeepLab LargeFOV StrongWeak \cite{Papandreou2015}}}}}
        \end{overpic}
    \end{subfigure}\\
    \begin{subfigure}[b]{0.499\linewidth}
        \begin{overpic}[width=\textwidth]{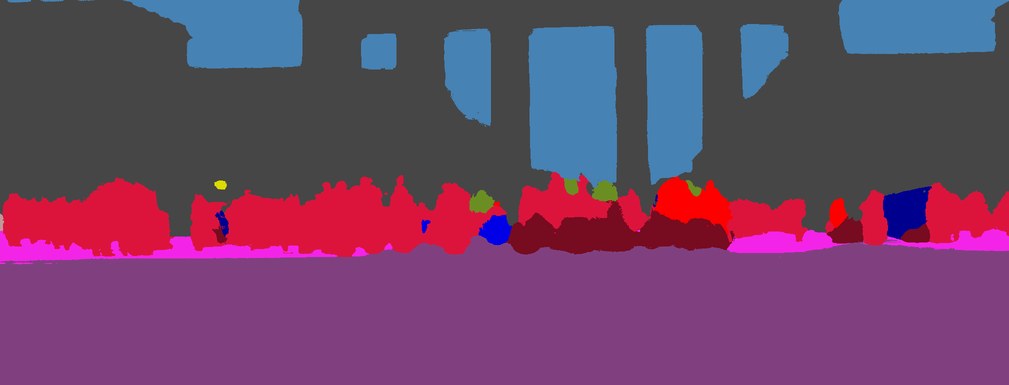}
        \put (50,4) {\makebox(0,0){\textcolor{white}{\textsf{\footnotesize Adelaide \cite{Lin2015}}}}}
        \end{overpic}
    \end{subfigure}\hfill
    \begin{subfigure}[b]{0.499\linewidth}
        \begin{overpic}[width=\textwidth]{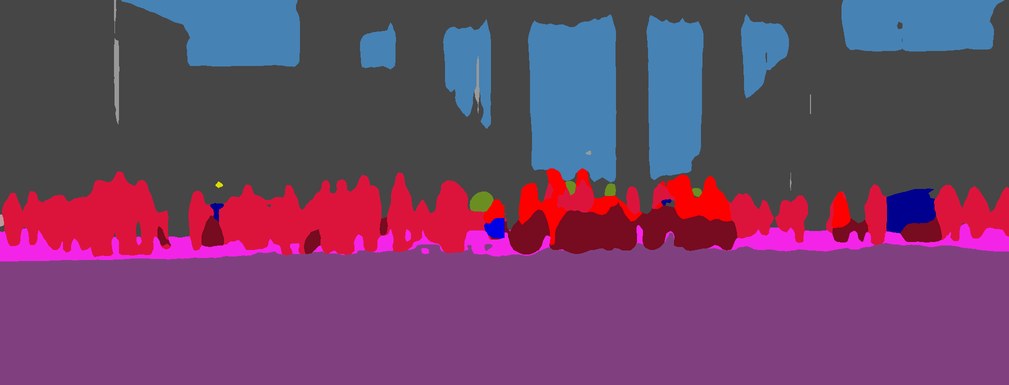}
        \put (50,4) {\makebox(0,0){\textcolor{white}{\textsf{\footnotesize Dilated10 \cite{Yu2016}}}}}
        \end{overpic}
    \end{subfigure}\\
    \vspace{2mm}
    \caption{Exemplary output of our baselines for the pixel-level semantic labeling task, see the main paper for details. The image is part of our \textit{test} set and has both, the largest number of instances and persons.}
    \label{fig:mostallbaselines}
\end{figure*}

%%%%%%%%%%%%%%%%%%%%%%%%%
% Largest number of cars
%%%%%%%%%%%%%%%%%%%%%%%%%

\begin{figure*}[p]
    \captionsetup[subfigure]{aboveskip=2pt,belowskip=2pt}
    \centering%
    \begin{subfigure}[b]{0.499\linewidth}
        \begin{overpic}[width=\textwidth]{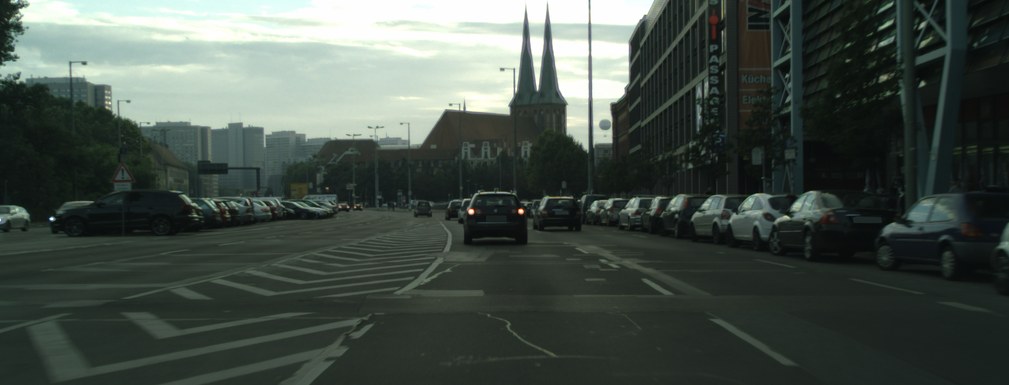}
        \put (50,4) {\makebox(0,0){\textcolor{white}{\textsf{\footnotesize Image}}}}
        \end{overpic}
    \end{subfigure}\hfill
    \begin{subfigure}[b]{0.499\linewidth}
        \begin{overpic}[width=\textwidth]{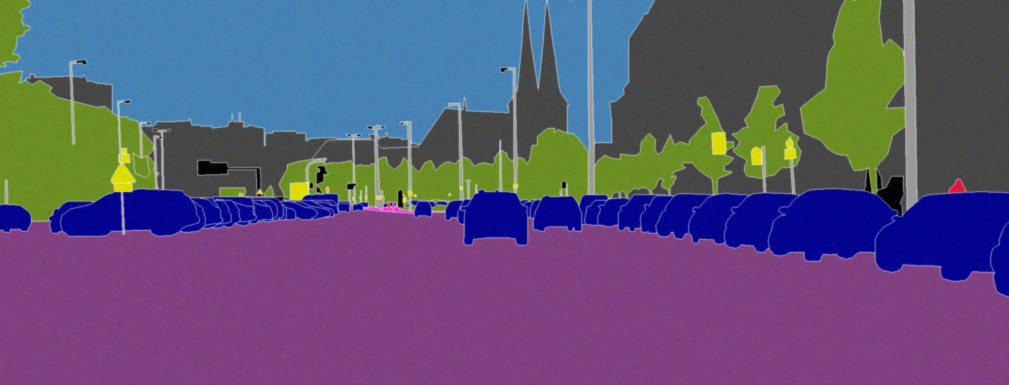}
        \put (50,4) {\makebox(0,0){\textcolor{white}{\textsf{\footnotesize Annotation}}}}
        \end{overpic}
    \end{subfigure}\\
    \begin{subfigure}[b]{0.499\linewidth}
        \begin{overpic}[width=\textwidth]{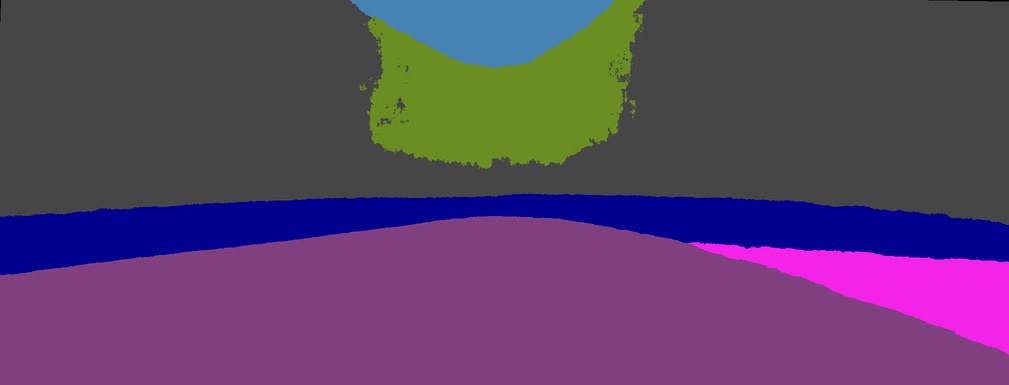}
        \put (50,4) {\makebox(0,0){\textcolor{white}{\textsf{\footnotesize static fine (SF)}}}}
        \end{overpic}
    \end{subfigure}\hfill
    \begin{subfigure}[b]{0.499\linewidth}
        \begin{overpic}[width=\textwidth]{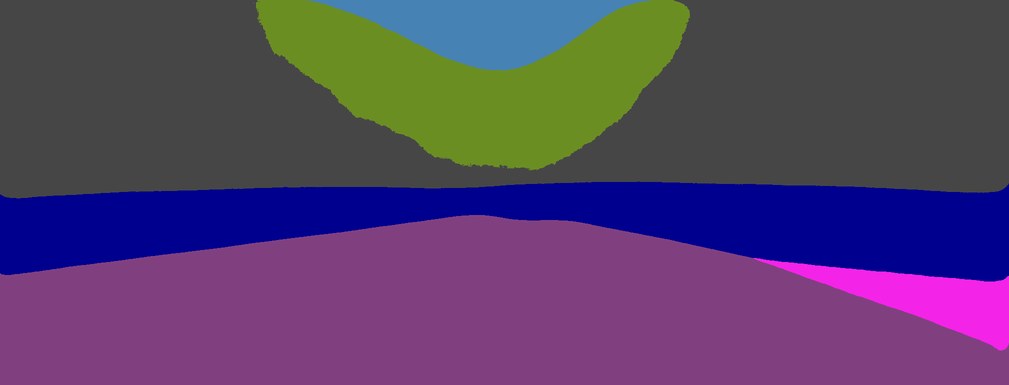}
        \put (50,4) {\makebox(0,0){\textcolor{white}{\textsf{\footnotesize static coarse (SC)}}}}
        \end{overpic}
    \end{subfigure}\\
    \begin{subfigure}[b]{0.499\linewidth}
        \begin{overpic}[width=\textwidth]{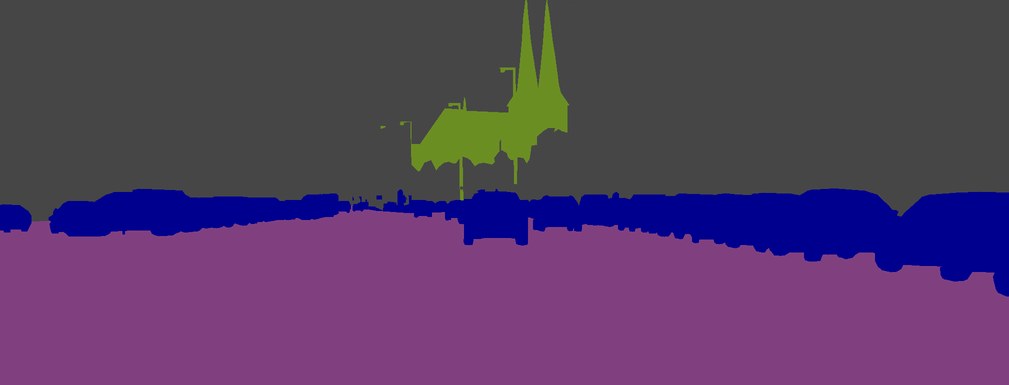}
        \put (50,4) {\makebox(0,0){\textcolor{white}{\textsf{\footnotesize GT segmentation w/ SF}}}}
        \end{overpic}
    \end{subfigure}\hfill
    \begin{subfigure}[b]{0.499\linewidth}
        \begin{overpic}[width=\textwidth]{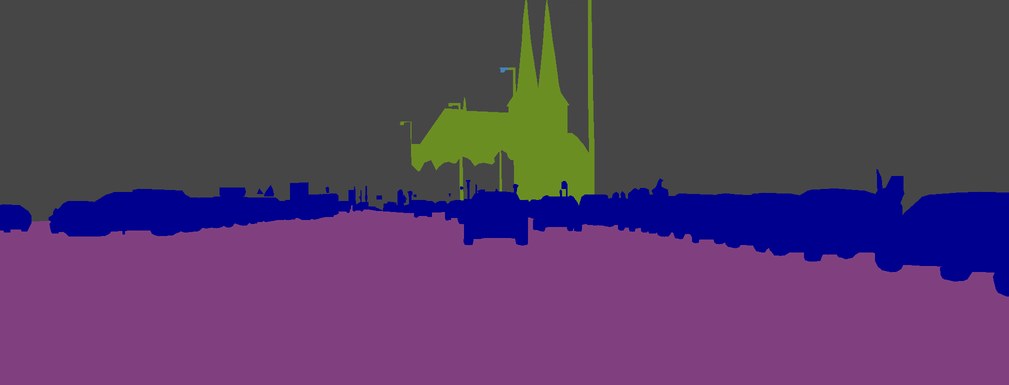}
        \put (50,4) {\makebox(0,0){\textcolor{white}{\textsf{\footnotesize GT segmentation w/ SC}}}}
        \end{overpic}
    \end{subfigure}\\
    \begin{subfigure}[b]{0.499\linewidth}
        \begin{overpic}[width=\textwidth]{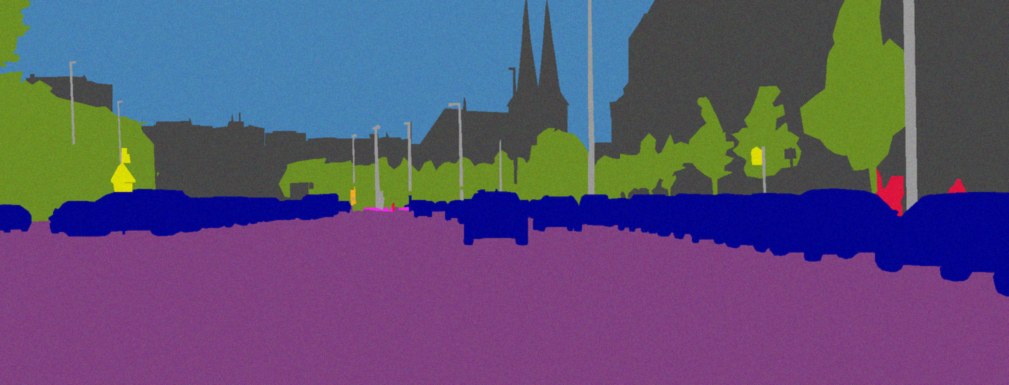}
        \put (50,4) {\makebox(0,0){\textcolor{white}{\textsf{\footnotesize GT segmentation w/ \cite{Long2015}}}}}
        \end{overpic}
    \end{subfigure}\hfill
    \begin{subfigure}[b]{0.499\linewidth}
        \begin{overpic}[width=\textwidth]{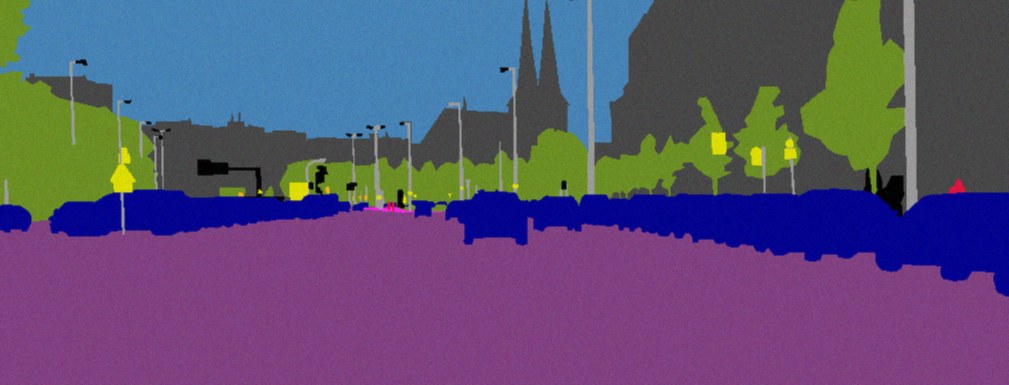}
        \put (50,4) {\makebox(0,0){\textcolor{white}{\textsf{\footnotesize GT subsampled by 2}}}}
        \end{overpic}
    \end{subfigure}\\
    \begin{subfigure}[b]{0.499\linewidth}
        \begin{overpic}[width=\textwidth]{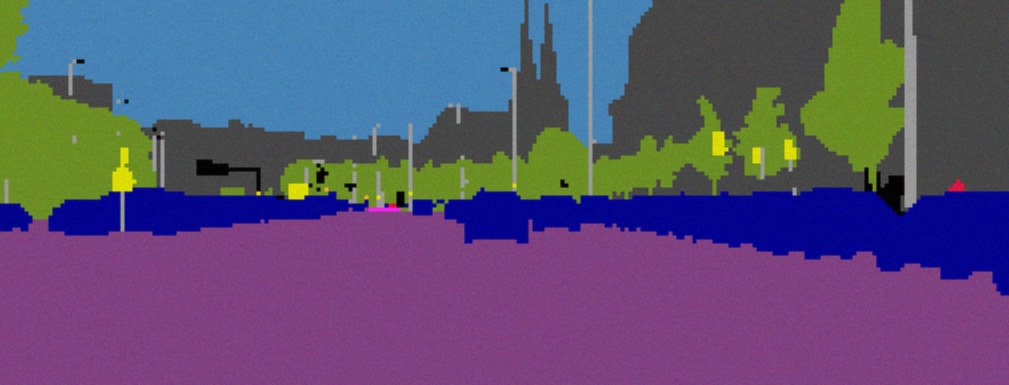}
        \put (50,4) {\makebox(0,0){\textcolor{white}{\textsf{\footnotesize GT subsampled by 8}}}}
        \end{overpic}
    \end{subfigure}\hfill
    \begin{subfigure}[b]{0.499\linewidth}
        \begin{overpic}[width=\textwidth]{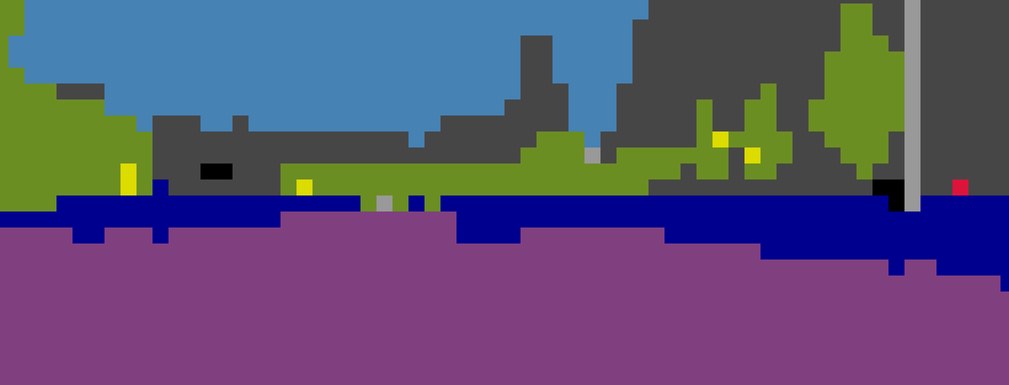}
        \put (50,4) {\makebox(0,0){\textcolor{white}{\textsf{\footnotesize GT subsampled by 32}}}}
        \end{overpic}
    \end{subfigure}\\
    \begin{subfigure}[b]{0.499\linewidth}
        \begin{overpic}[width=\textwidth]{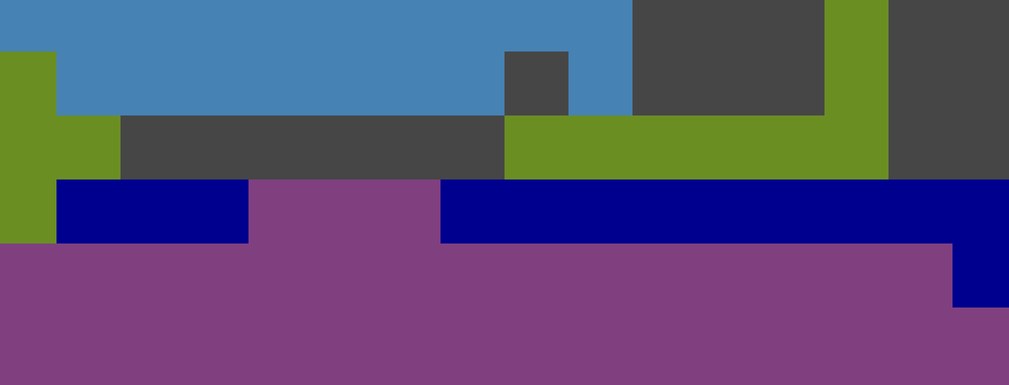}
        \put (50,4) {\makebox(0,0){\textcolor{white}{\textsf{\footnotesize GT subsampled by 128}}}}
        \end{overpic}
    \end{subfigure}\hfill
    \begin{subfigure}[b]{0.499\linewidth}
        \begin{overpic}[width=\textwidth]{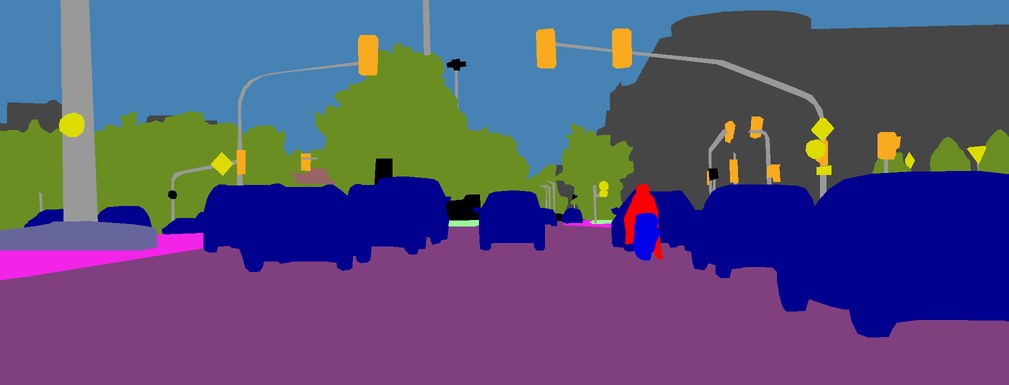}
        \put (50,4) {\makebox(0,0){\textcolor{white}{\textsf{\footnotesize nearest training neighbor}}}}
        \end{overpic}
    \end{subfigure}\\
    \vspace{2mm}
    \caption{Exemplary output of our control experiments for the pixel-level semantic labeling task, see the main paper for details. The image is part of our \textit{test} set and has the largest number of car instances.}
    \label{fig:mostcarcontrol}
\end{figure*}

\begin{figure*}[p]
    \captionsetup[subfigure]{aboveskip=2pt,belowskip=2pt}
    \centering%
    \begin{subfigure}[b]{0.499\linewidth}
        \begin{overpic}[width=\textwidth]{figures/examplesAutoInclBaselineSmallRes/berlin00182_000019_most_car_img}
        \put (50,4) {\makebox(0,0){\textcolor{white}{\textsf{\footnotesize Image}}}}
        \end{overpic}
    \end{subfigure}\hfill
    \begin{subfigure}[b]{0.499\linewidth}
        \begin{overpic}[width=\textwidth]{figures/examplesAutoInclBaselineSmallRes/berlin00182_000019_most_car_annotation}
        \put (50,4) {\makebox(0,0){\textcolor{white}{\textsf{\footnotesize Annotation}}}}
        \end{overpic}
    \end{subfigure}\\
    \begin{subfigure}[b]{0.499\linewidth}
        \begin{overpic}[width=\textwidth]{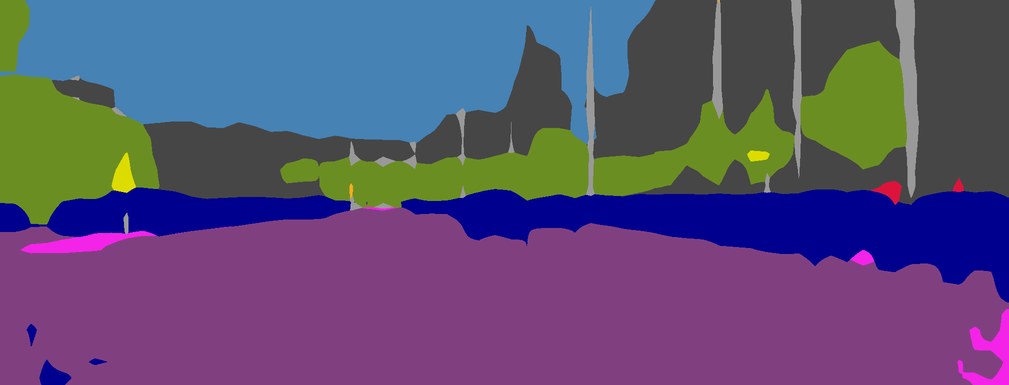}
        \put (50,4) {\makebox(0,0){\textcolor{white}{\textsf{\footnotesize FCN-32s}}}}
        \end{overpic}
    \end{subfigure}\hfill
    \begin{subfigure}[b]{0.499\linewidth}
        \begin{overpic}[width=\textwidth]{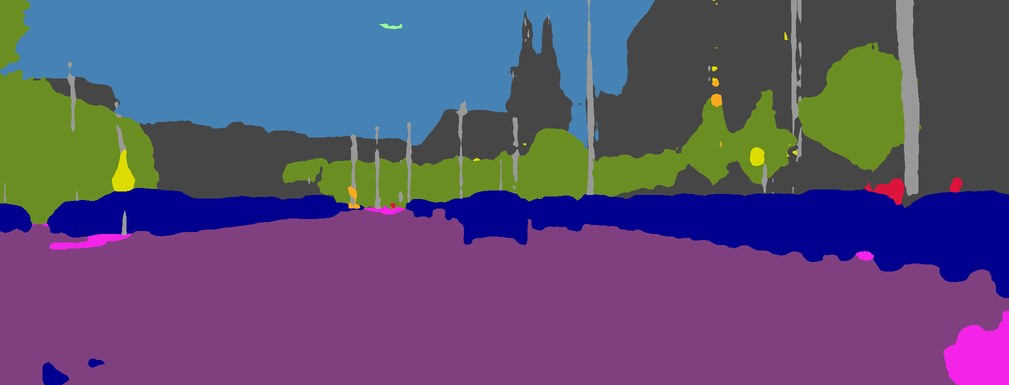}
        \put (50,4) {\makebox(0,0){\textcolor{white}{\textsf{\footnotesize FCN-8s}}}}
        \end{overpic}
    \end{subfigure}\\
    \begin{subfigure}[b]{0.499\linewidth}
        \begin{overpic}[width=\textwidth]{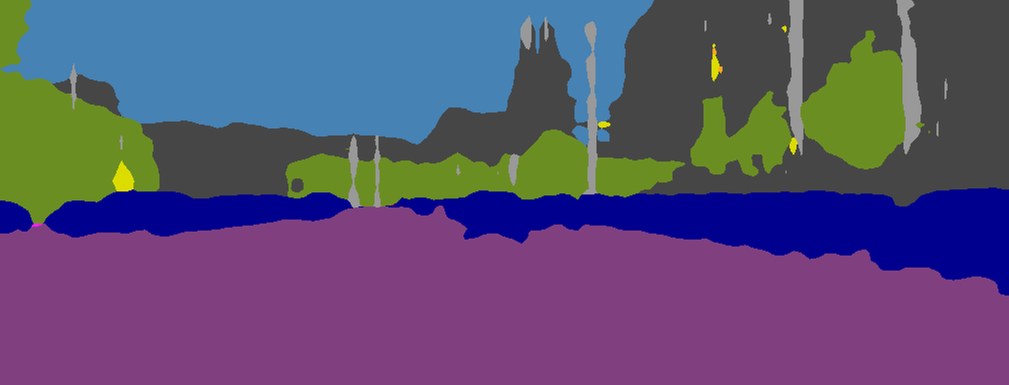}
        \put (50,4) {\makebox(0,0){\textcolor{white}{\textsf{\footnotesize FCN-8s half resolution}}}}
        \end{overpic}
    \end{subfigure}\hfill
    \begin{subfigure}[b]{0.499\linewidth}
        \begin{overpic}[width=\textwidth]{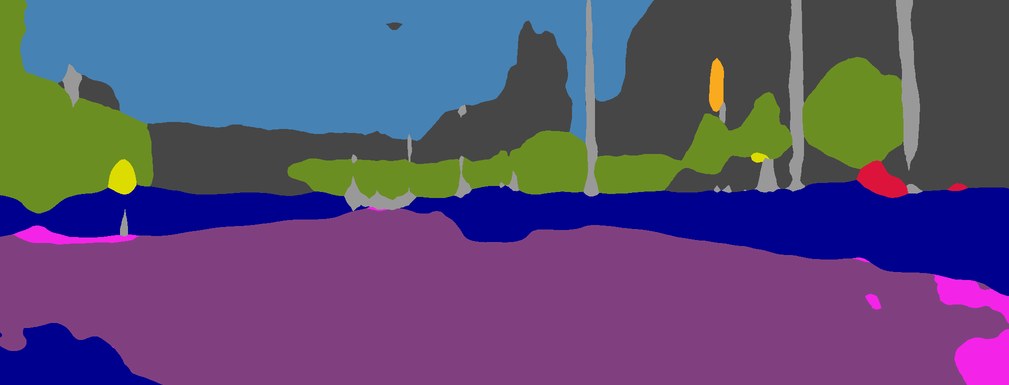}
        \put (50,4) {\makebox(0,0){\textcolor{white}{\textsf{\footnotesize FCN-8s trained on coarse}}}}
        \end{overpic}
    \end{subfigure}\\
    \begin{subfigure}[b]{0.499\linewidth}
        \begin{overpic}[width=\textwidth]{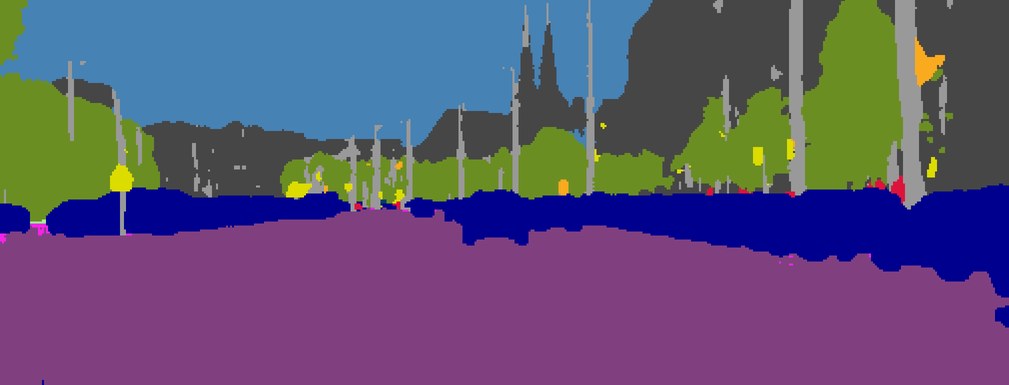}
        \put (50,4) {\makebox(0,0){\textcolor{white}{\textsf{\footnotesize SegNet basic \cite{Badrinarayanan2015}}}}}
        \end{overpic}
    \end{subfigure}\hfill
    \begin{subfigure}[b]{0.499\linewidth}
        \begin{overpic}[width=\textwidth]{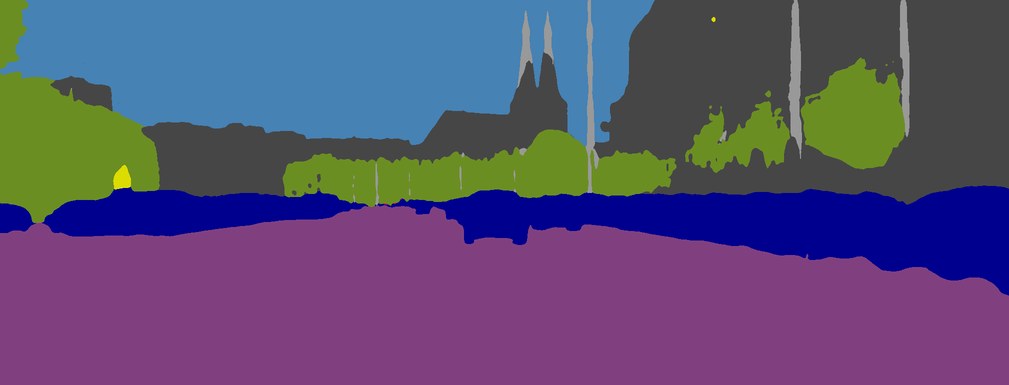}
        \put (50,4) {\makebox(0,0){\textcolor{white}{\textsf{\footnotesize DPN \cite{Liu2015}}}}}
        \end{overpic}
    \end{subfigure}\\
    \begin{subfigure}[b]{0.499\linewidth}
        \begin{overpic}[width=\textwidth]{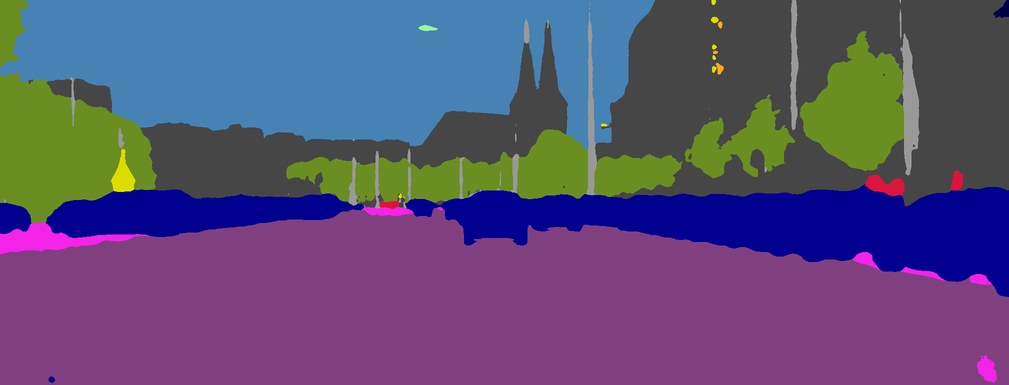}
        \put (50,4) {\makebox(0,0){\textcolor{white}{\textsf{\footnotesize CRF as RNN \cite{Zheng2015}}}}}
        \end{overpic}
    \end{subfigure}\hfill
    \begin{subfigure}[b]{0.499\linewidth}
        \begin{overpic}[width=\textwidth]{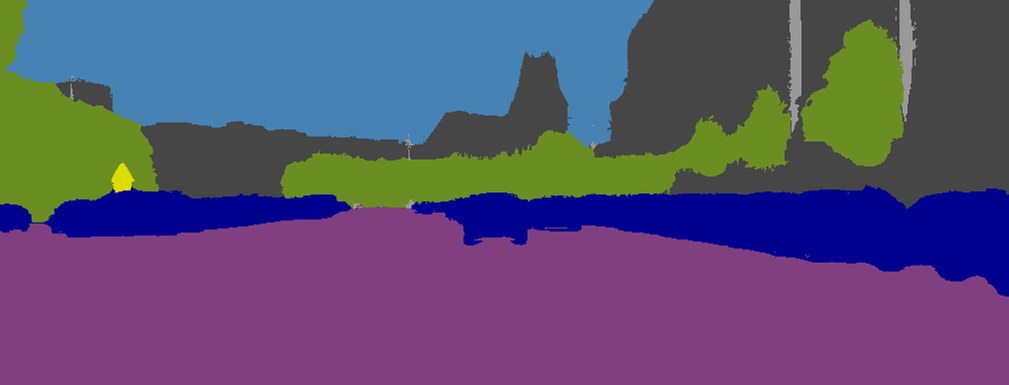}
        \put (50,4) {\makebox(0,0){\textcolor{white}{\textsf{\footnotesize DeepLab LargeFOV StrongWeak \cite{Papandreou2015}}}}}
        \end{overpic}
    \end{subfigure}\\
    \begin{subfigure}[b]{0.499\linewidth}
        \begin{overpic}[width=\textwidth]{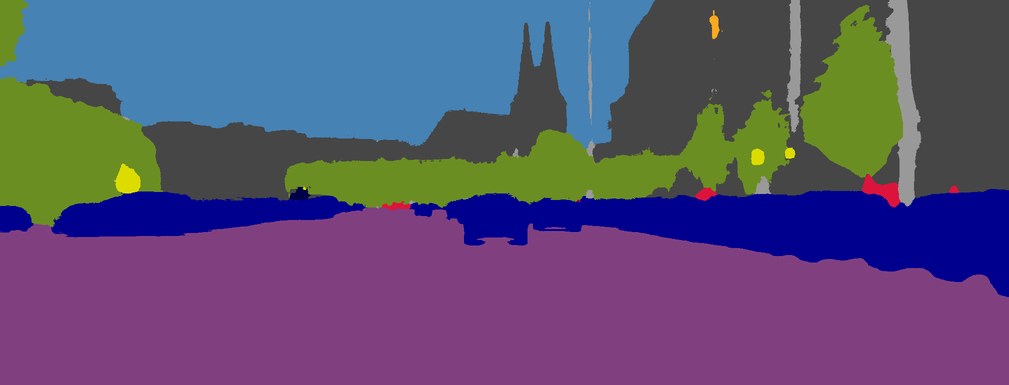}
        \put (50,4) {\makebox(0,0){\textcolor{white}{\textsf{\footnotesize Adelaide \cite{Lin2015}}}}}
        \end{overpic}
    \end{subfigure}\hfill
    \begin{subfigure}[b]{0.499\linewidth}
        \begin{overpic}[width=\textwidth]{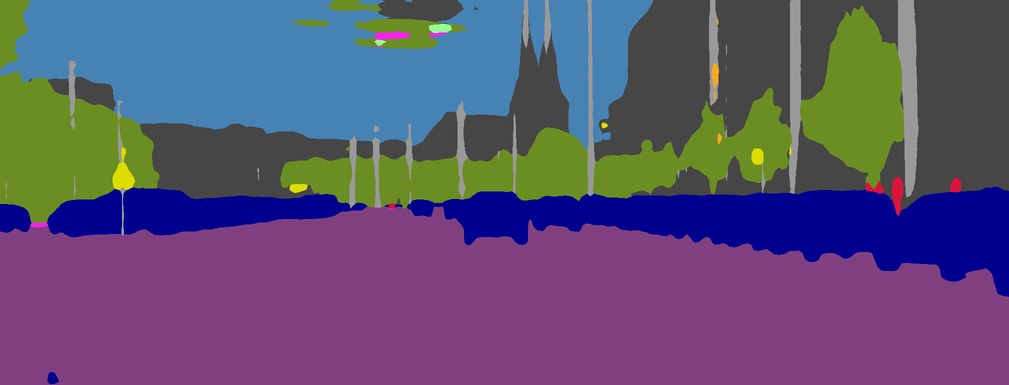}
        \put (50,4) {\makebox(0,0){\textcolor{white}{\textsf{\footnotesize Dilated10 \cite{Yu2016}}}}}
        \end{overpic}
    \end{subfigure}\\
    \vspace{2mm}
    \caption{Exemplary output of our baseline experiments for the pixel-level semantic labeling task, see the main paper for details. The image is part of our \textit{test} set and has the largest number of car instances.}
    \label{fig:mostcarbaselines}
\end{figure*}

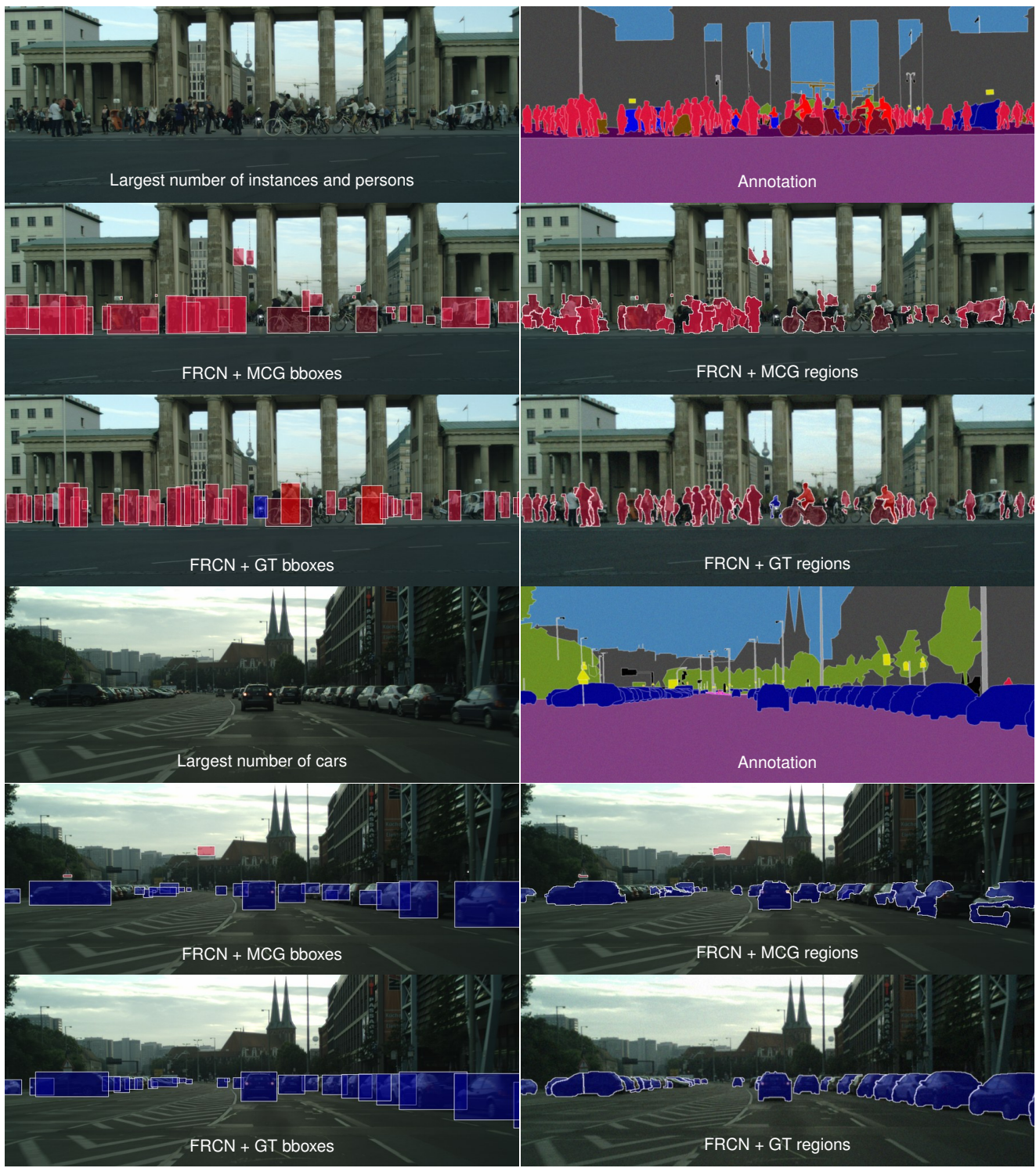
\begin{figure*}[p]
    \captionsetup[subfigure]{aboveskip=2pt,belowskip=2pt}
    \centering%
    \begin{subfigure}[b]{0.499\linewidth}
        \begin{overpic}[width=\textwidth]{figures/examplesAutoInclBaselineSmallRes/berlin00057_000019_most_all_img}
        \put (50,4) {\makebox(0,0){\textcolor{white}{\textsf{\footnotesize Largest number of instances and persons}}}}
        \end{overpic}
    \end{subfigure}\hfill
    \begin{subfigure}[b]{0.499\linewidth}
        \begin{overpic}[width=\textwidth]{figures/examplesAutoInclBaselineSmallRes/berlin00057_000019_most_all_annotation}
        \put (50,4) {\makebox(0,0){\textcolor{white}{\textsf{\footnotesize Annotation}}}}
        \end{overpic}
    \end{subfigure}\\
    \begin{subfigure}[b]{0.499\linewidth}
        \begin{overpic}[width=\textwidth]{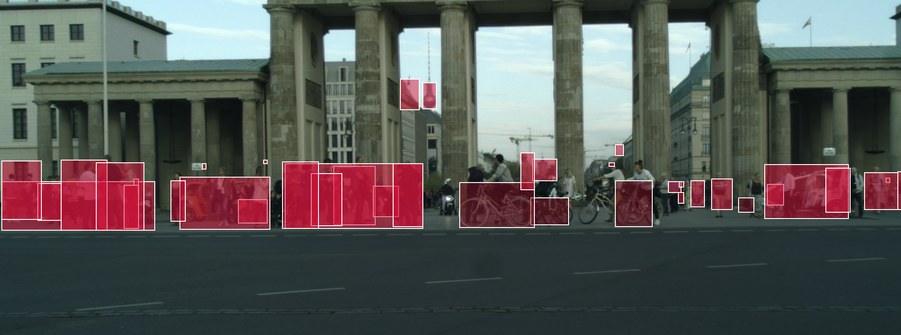}
        \put (50,4) {\makebox(0,0){\textcolor{white}{\textsf{\footnotesize FRCN + MCG bboxes}}}}
        \end{overpic}
    \end{subfigure}\hfill
    \begin{subfigure}[b]{0.499\linewidth}
        \begin{overpic}[width=\textwidth]{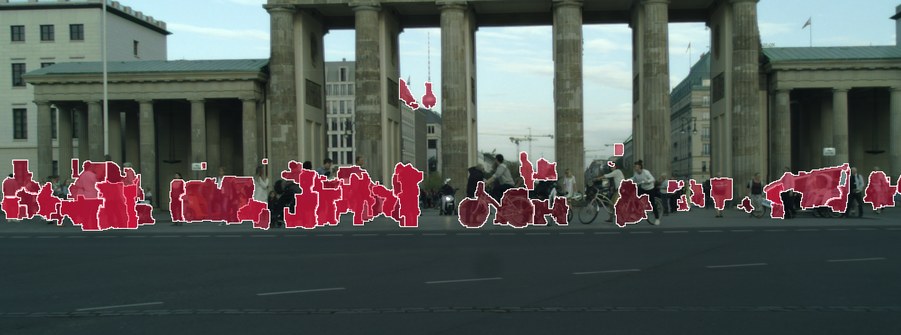}
        \put (50,4) {\makebox(0,0){\textcolor{white}{\textsf{\footnotesize FRCN + MCG regions}}}}
        \end{overpic}
    \end{subfigure}\\
    \begin{subfigure}[b]{0.499\linewidth}
        \begin{overpic}[width=\textwidth]{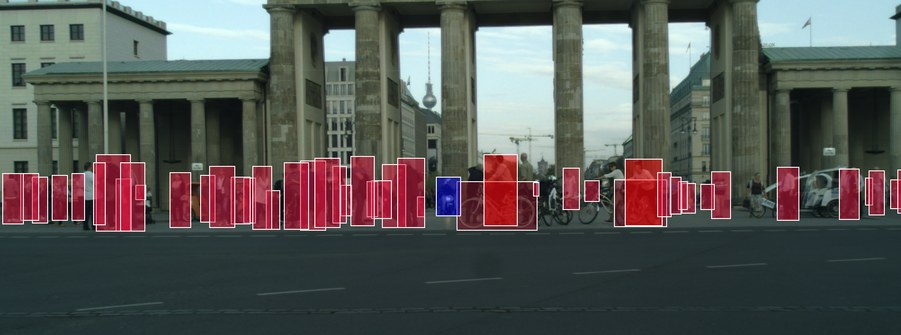}
        \put (50,4) {\makebox(0,0){\textcolor{white}{\textsf{\footnotesize FRCN + GT bboxes}}}}
        \end{overpic}
    \end{subfigure}\hfill
    \begin{subfigure}[b]{0.499\linewidth}
        \begin{overpic}[width=\textwidth]{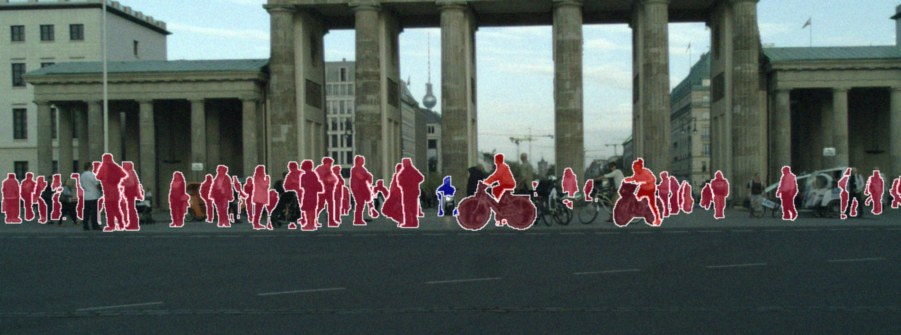}
        \put (50,4) {\makebox(0,0){\textcolor{white}{\textsf{\footnotesize FRCN + GT regions}}}}
        \end{overpic}
    \end{subfigure}\\
    \begin{subfigure}[b]{0.499\linewidth}
        \begin{overpic}[width=\textwidth]{figures/examplesAutoInclBaselineSmallRes/berlin00182_000019_most_car_img}
        \put (50,4) {\makebox(0,0){\textcolor{white}{\textsf{\footnotesize Largest number of cars}}}}
        \end{overpic}
    \end{subfigure}\hfill
    \begin{subfigure}[b]{0.499\linewidth}
        \begin{overpic}[width=\textwidth]{figures/examplesAutoInclBaselineSmallRes/berlin00182_000019_most_car_annotation}
        \put (50,4) {\makebox(0,0){\textcolor{white}{\textsf{\footnotesize Annotation}}}}
        \end{overpic}
    \end{subfigure}\\
    \begin{subfigure}[b]{0.499\linewidth}
        \begin{overpic}[width=\textwidth]{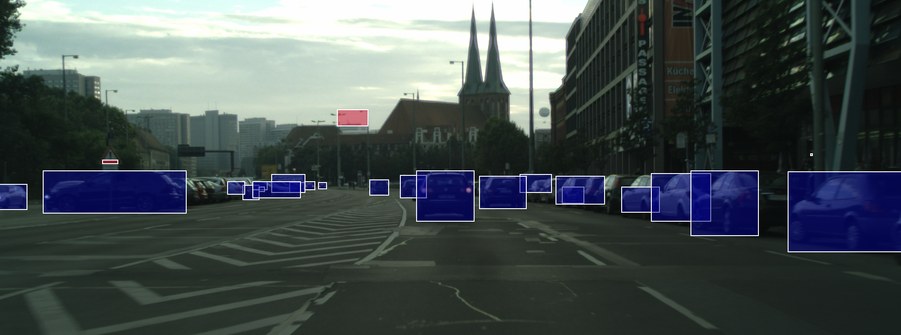}
        \put (50,4) {\makebox(0,0){\textcolor{white}{\textsf{\footnotesize FRCN + MCG bboxes}}}}
        \end{overpic}
    \end{subfigure}\hfill
    \begin{subfigure}[b]{0.499\linewidth}
        \begin{overpic}[width=\textwidth]{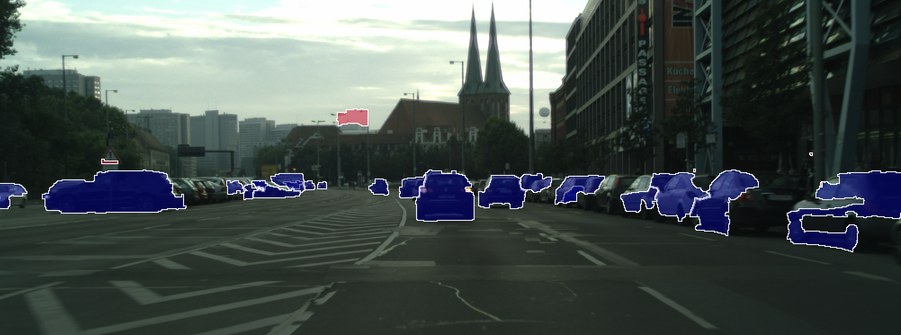}
        \put (50,4) {\makebox(0,0){\textcolor{white}{\textsf{\footnotesize FRCN + MCG regions}}}}
        \end{overpic}
    \end{subfigure}\\
    \begin{subfigure}[b]{0.499\linewidth}
        \begin{overpic}[width=\textwidth]{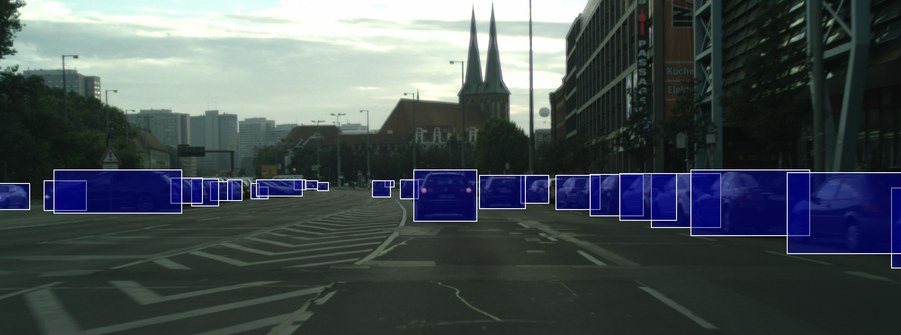}
        \put (50,4) {\makebox(0,0){\textcolor{white}{\textsf{\footnotesize FRCN + GT bboxes}}}}
        \end{overpic}
    \end{subfigure}\hfill
    \begin{subfigure}[b]{0.499\linewidth}
        \begin{overpic}[width=\textwidth]{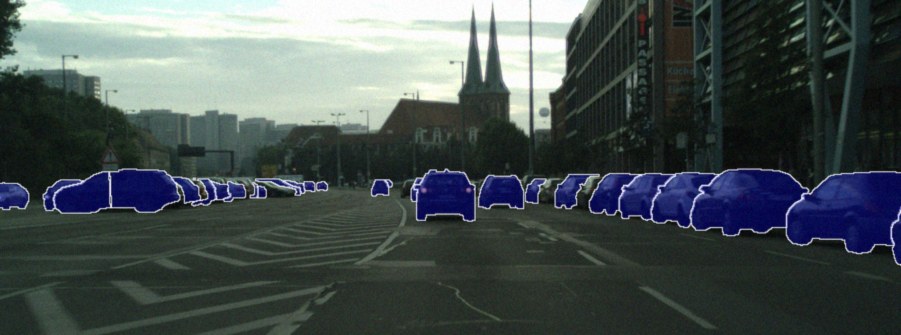}
        \put (50,4) {\makebox(0,0){\textcolor{white}{\textsf{\footnotesize FRCN + GT regions}}}}
        \end{overpic}
    \end{subfigure}\\
    \vspace{2mm}
    \caption{Exemplary output of our control experiments and baselines for the instance-level semantic labeling task, see the main paper for details.}
    \label{fig:instanceresults}
\end{figure*}

\end{document}